%% file: main.tex
\pdfoutput=1
\documentclass[10pt,twocolumn,letterpaper]{article}

\usepackage{preprint}
\input{preamble}

\definecolor{linkblue}{rgb}{0.21,0.49,0.74}
\usepackage[breaklinks,colorlinks,allcolors=linkblue]{hyperref}

\title{Edit the Shoot, Not the Shot:\\Aesthetic Portrait Reshooting with Video Action Simulators}

\author{
Shen Ou-Yang$^{1}$ \qquad
Dong Chen$^{3}$ \qquad
Xianhui Lin$^{2}$ \qquad
Xiaoming Li$^{3,\text{*}}$ \\[4pt]
$^{1}$Tsinghua University \qquad
$^{2}$Vivo \qquad
$^{3}$Nanjing University \\[5pt]
{\normalsize Project Page: \href{https://shenouy03.github.io/simshoot}{\texttt{https://shenouy03.github.io/simshoot}}}
}

\begin{document}
\twocolumn[{%
\renewcommand\twocolumn[1][]{#1}%
\maketitle
\vspace{-1.25em}
\begin{center}
    \captionsetup{type=figure}
    \includegraphics[width=\textwidth]{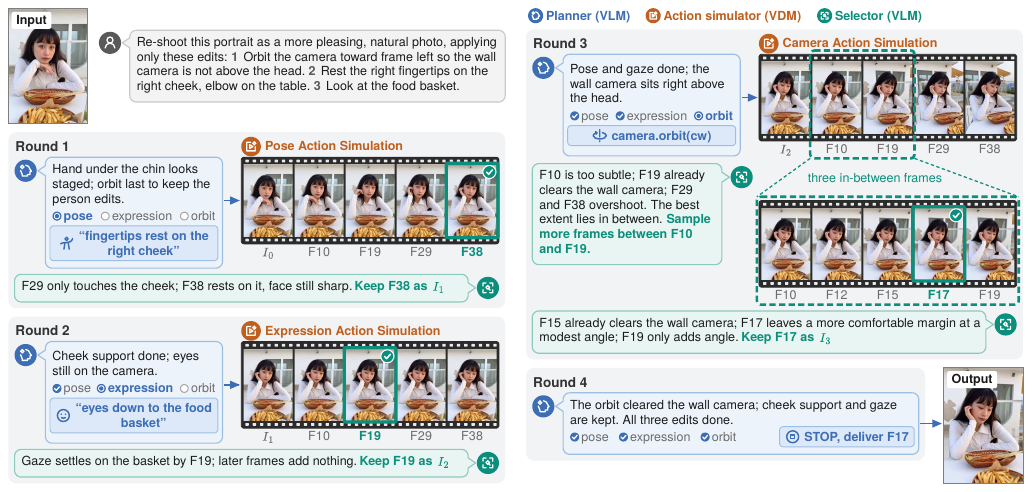}
    \captionof{figure}{\textbf{SimShoot reshoots a portrait by simulating the shoot.}
    A targeted AesReShot-Eval run illustrates partial execution in Round~2 (F19) and interval refinement in Round~3 (dashed), where three in-between frames are inspected before selecting F17.
    Green marks the selected frames.
    In Round~4, the Planner stops and returns the last kept portrait.
    Dialogue is condensed from model outputs.}
    \label{fig:teaser}
\end{center}%
\vspace{0.15em}
}]%
{\let\thefootnote\relax\begin{NoHyper}\footnotetext{\textsuperscript{*}Corresponding author.}\end{NoHyper}}%
\input{sec/0_abstract}

\input{sec/1_intro}

\input{sec/2_related}
\input{sec/3_task}

\begin{figure*}[t]
    \centering
    \includegraphics[width=\textwidth]{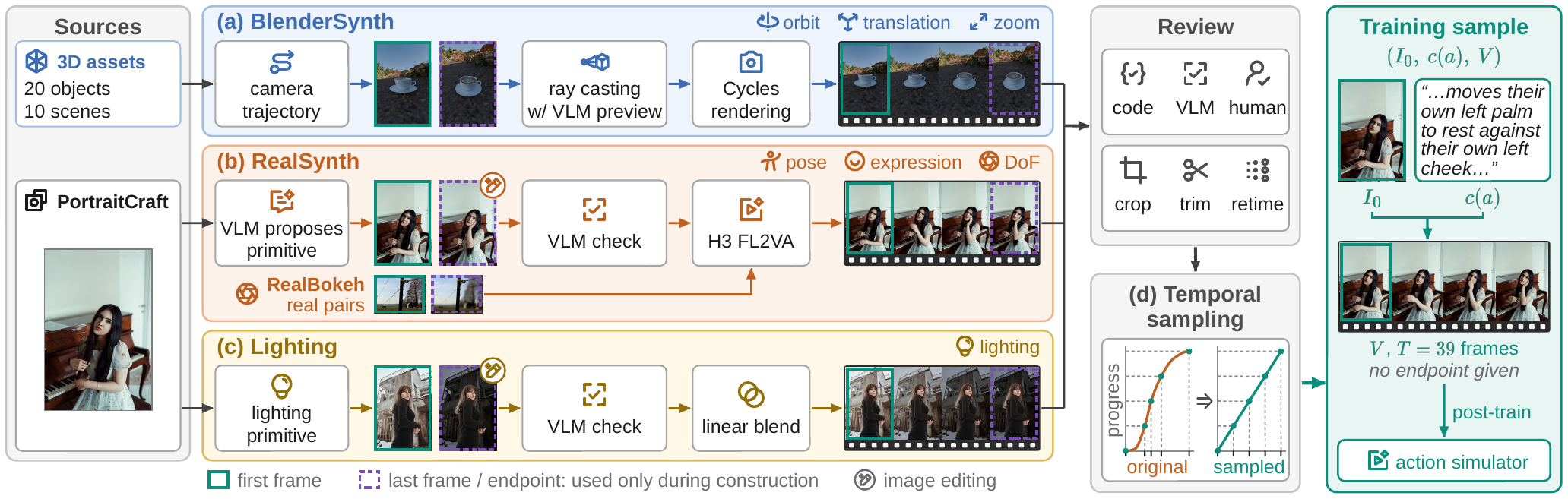}
    \caption{\textbf{Dataset construction overview.}
    Three complementary routes produce single-action videos: (a) BlenderSynth renders 3D camera trajectories, (b) RealSynth generates transitions from verified endpoint pairs, and (c) Lighting blends source portraits with verified relit endpoints smoothly in linear-light RGB.
    Route-specific screening and temporal preparation yield 39-frame training samples $(I_0,c(a),V)$; panel (d) illustrates progress-aware sampling (Sec.~\ref{sec:data-preparation}).
    Teal and dashed purple borders mark initial and endpoint images, respectively.
    Endpoint conditioning is used only offline; the simulator is trained and run from the starting portrait and action text alone.}
    \label{fig:dataset-construction}
\end{figure*}

\input{sec/4_dataset}

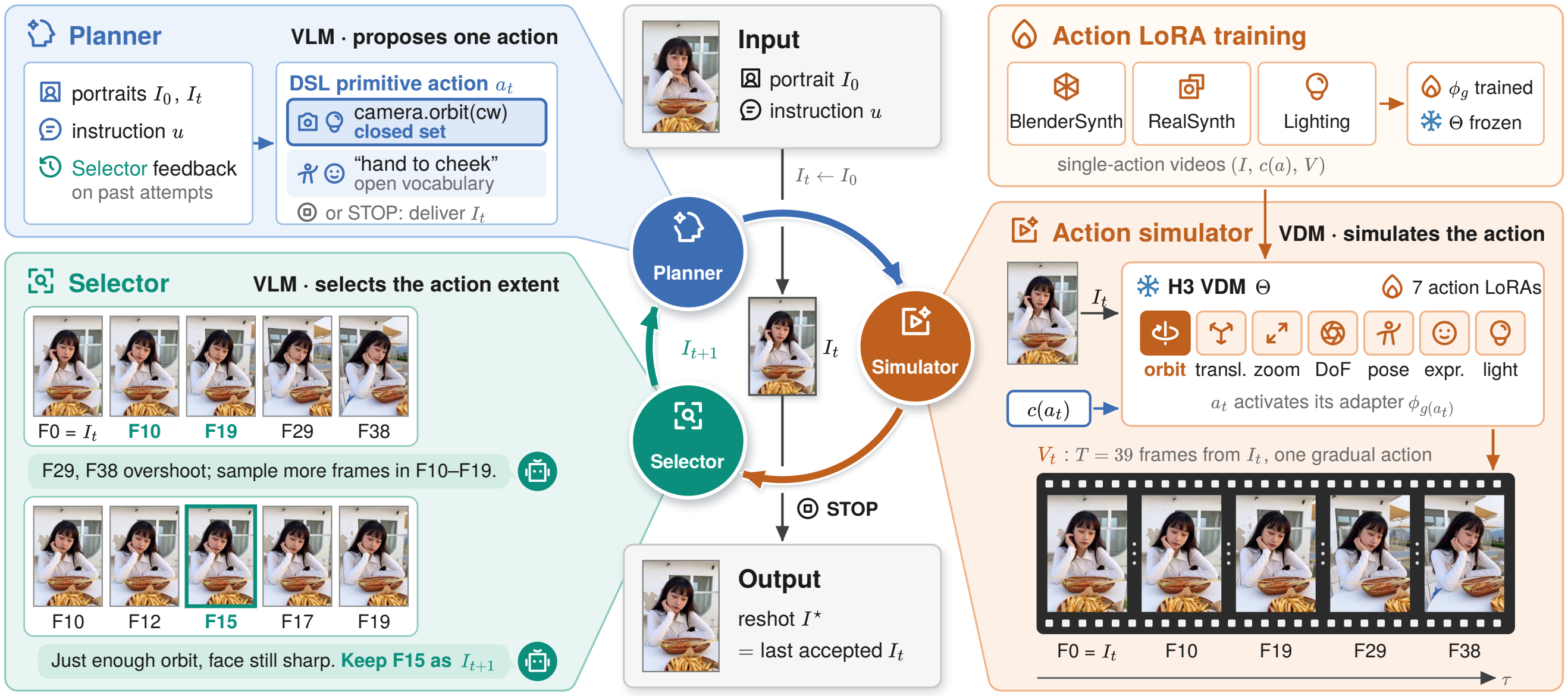
\begin{figure*}[t]
    \centering
    \begingroup
    \definecolor{wfW}{HTML}{FFFFFF}
    \definecolor{wfInk}{HTML}{1A1A1A}
    \definecolor{wfTeal}{HTML}{0B8F7C}
    \definecolor{wfBar}{HTML}{D3ECE6}
    \definecolor{wfSim}{HTML}{FDF1E7}
    \resizebox{\textwidth}{!}{%
    \begin{tikzpicture}[x=1bp,y=-1bp]
      \node[anchor=north west,inner sep=0pt,outer sep=0pt] at (0,0) {\includegraphics[width=495bp,height=221bp]{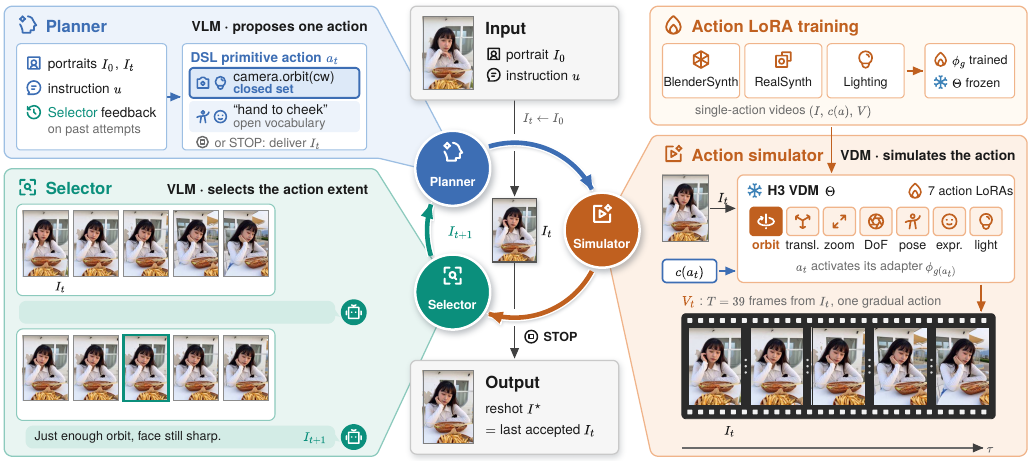}};
      \path[fill=wfW] (11.7,134.0) rectangle (25.5,140.8);
      \node[text=wfInk,anchor=base east,inner sep=0pt,font={\sffamily\fontsize{6.0bp}{6.8bp}\selectfont}] at (24.66,139.39) {F0 =};
      \path[fill=wfW] (39.8,134.0) rectangle (52.0,140.8);
      \node[text=wfTeal,anchor=base,inner sep=0pt,font={\sffamily\fontsize{6.0bp}{6.8bp}\selectfont\bfseries}] at (45.96,139.39) {F10};
      \path[fill=wfW] (63.9,134.0) rectangle (76.2,140.8);
      \node[text=wfTeal,anchor=base,inner sep=0pt,font={\sffamily\fontsize{6.0bp}{6.8bp}\selectfont\bfseries}] at (70.06,139.39) {F19};
      \path[fill=wfW] (88.0,134.0) rectangle (100.3,140.8);
      \node[text=wfInk,anchor=base,inner sep=0pt,font={\sffamily\fontsize{6.0bp}{6.8bp}\selectfont}] at (94.16,139.39) {F29};
      \path[fill=wfW] (112.1,134.0) rectangle (124.4,140.8);
      \node[text=wfInk,anchor=base,inner sep=0pt,font={\sffamily\fontsize{6.0bp}{6.8bp}\selectfont}] at (118.26,139.39) {F38};
      \path[fill=wfBar] (12.8,146.3) rectangle (158.8,153.6);
      \node[text=wfInk,anchor=base west,inner sep=0pt,font={\sffamily\fontsize{6.0bp}{6.8bp}\selectfont}] at (13.38,151.69) {F29, F38 overshoot; sample more frames in F10\textendash F19.};
      \path[fill=wfW] (15.7,193.9) rectangle (28.0,200.7);
      \node[text=wfInk,anchor=base,inner sep=0pt,font={\sffamily\fontsize{6.0bp}{6.8bp}\selectfont}] at (21.86,199.29) {F10};
      \path[fill=wfW] (39.8,193.9) rectangle (52.1,200.7);
      \node[text=wfInk,anchor=base,inner sep=0pt,font={\sffamily\fontsize{6.0bp}{6.8bp}\selectfont}] at (45.96,199.29) {F12};
      \path[fill=wfW] (63.9,193.9) rectangle (76.2,200.7);
      \node[text=wfTeal,anchor=base,inner sep=0pt,font={\sffamily\fontsize{6.0bp}{6.8bp}\selectfont\bfseries}] at (70.06,199.29) {F15};
      \path[fill=wfW] (88.0,193.9) rectangle (100.3,200.7);
      \node[text=wfInk,anchor=base,inner sep=0pt,font={\sffamily\fontsize{6.0bp}{6.8bp}\selectfont}] at (94.16,199.29) {F17};
      \path[fill=wfW] (112.1,193.9) rectangle (124.4,200.7);
      \node[text=wfInk,anchor=base,inner sep=0pt,font={\sffamily\fontsize{6.0bp}{6.8bp}\selectfont}] at (118.26,199.29) {F19};
      \path[fill=wfBar] (107.0,206.3) rectangle (144.0,213.4);
      \node[text=wfTeal,anchor=base east,inner sep=0pt,font={\sffamily\fontsize{6.0bp}{6.8bp}\selectfont\bfseries}] at (142.98,211.59) {Keep F15 as};
      \path[fill=wfSim] (332.9,203.2) rectangle (346.8,210.1);
      \node[text=wfInk,anchor=base east,inner sep=0pt,font={\sffamily\fontsize{6.0bp}{6.8bp}\selectfont}] at (345.92,208.59) {F0 =};
      \path[fill=wfSim] (366.7,203.2) rectangle (379.0,210.1);
      \node[text=wfInk,anchor=base,inner sep=0pt,font={\sffamily\fontsize{6.0bp}{6.8bp}\selectfont}] at (372.85,208.59) {F10};
      \path[fill=wfSim] (396.4,203.2) rectangle (408.7,210.1);
      \node[text=wfInk,anchor=base,inner sep=0pt,font={\sffamily\fontsize{6.0bp}{6.8bp}\selectfont}] at (402.57,208.59) {F19};
      \path[fill=wfSim] (426.1,203.2) rectangle (438.4,210.1);
      \node[text=wfInk,anchor=base,inner sep=0pt,font={\sffamily\fontsize{6.0bp}{6.8bp}\selectfont}] at (432.29,208.59) {F29};
      \path[fill=wfSim] (455.8,203.2) rectangle (468.1,210.1);
      \node[text=wfInk,anchor=base,inner sep=0pt,font={\sffamily\fontsize{6.0bp}{6.8bp}\selectfont}] at (462.01,208.59) {F38};
    \end{tikzpicture}%
    }
    \endgroup
    \caption{\textbf{SimShoot workflow.}
    Given portrait $I_0$ and optional instruction $u$, the Planner proposes one structured DSL action $a_t$.
    The VDM action simulator activates the corresponding LoRA on frozen MiniMax H3 to generate a 39-frame transition from the current portrait $I_t$.
    The Selector compares intermediate frames (\texttt{F0}--\texttt{F38}) to choose $I_{t+1}$ or retain $I_t$, with visual feedback guiding subsequent planning.
    The inset summarizes single-action video training across the seven adapters.}
    \label{fig:simshoot-workflow}
\end{figure*}

\input{sec/5_method}
\input{sec/6_experiments}
\input{sec/7_conclusion}

{
    \small
    \setlength{\bibsep}{0.7pt plus 0.5pt minus 0.5pt}
    \bibliographystyle{ieeenat_fullname}
    \bibliography{zotero,supplement_refs}
}

\clearpage
\onecolumn
\raggedbottom
\appendix
\setcounter{figure}{0}
\setcounter{table}{0}
\renewcommand{\thefigure}{S\arabic{figure}}
\renewcommand{\thetable}{S\arabic{table}}
\renewcommand{\theHfigure}{S\arabic{figure}}
\renewcommand{\theHtable}{S\arabic{table}}

\renewcommand{\topfraction}{0.92}
\renewcommand{\bottomfraction}{0.88}
\renewcommand{\textfraction}{0.06}
\renewcommand{\floatpagefraction}{0.85}
\setcounter{topnumber}{4}
\setcounter{bottomnumber}{4}
\setcounter{totalnumber}{6}
\setlength{\textfloatsep}{10pt plus 2pt minus 2pt}
\setlength{\floatsep}{9pt plus 2pt minus 2pt}
\setlength{\intextsep}{10pt plus 2pt minus 2pt}
\makeatletter
\setlength{\@fptop}{0pt}
\setlength{\@fpsep}{12pt}
\makeatother
\widowpenalty=3000
\clubpenalty=3000

\begin{center}
    {\Large\bfseries \thetitle\par}
    \vspace{0.4em}
    {\Large Appendix\par}
    \vspace{0.8em}
\end{center}

\input{sec/X_suppl}

\end{document}

%% file: preamble.tex
\usepackage[T1]{fontenc}
\usepackage{microtype}
\usepackage{colortbl}
\usepackage{placeins}
\usepackage{multirow}
\usepackage{tabularx}
\newcolumntype{Y}{>{\centering\arraybackslash}X}
\definecolor{ResultBest}{RGB}{212,227,211}
\definecolor{ResultSecond}{RGB}{237,244,236}
\newcommand{\resultshade}[2]{%
  {\setlength{\fboxsep}{0pt}\colorbox{#1}{%
    \makebox[\dimexpr\linewidth-4pt\relax]{\rule[-2pt]{0pt}{10pt}#2}}}}
\usepackage{pdfrender}
\newcommand{\resultemphasis}[1]{%
  \textpdfrender{TextRenderingMode=FillStroke,LineWidth=0.14pt}{#1}}

\usepackage{tikz}

%% file: sec/0_abstract.tex
\begin{abstract}
\textbf{Aesthetic portrait reshooting} aims to improve portrait aesthetics through adjustments available during a photoshoot, including camera control, pose, expression, and lighting.
Existing image editors can produce appealing outputs without realizing the requested photographic changes, and their single-image outputs make the extent of each adjustment hard to control.
We propose \textbf{SimShoot}, a framework that edits the shoot rather than the shot by decoupling which adjustment to make from how far to execute it.
A vision-language planner proposes one photographic action at a time; a video diffusion model, adapted as an action simulator, generates it as a gradual transition whose frames expose increasing extents; and a vision-language selector keeps the most suitable frame or rejects the action, informing the next round of planning in a closed loop.
To ensure faithful and isolated action execution, we train the simulator on single-action videos built from rendered 3D camera trajectories, endpoint-conditioned portrait transitions, and linear-light relighting.
On our 240-portrait \textbf{AesReShot-Eval} benchmark with open-ended and targeted tracks, crowd raters prefer SimShoot over all four open-weight and reshooting baselines (54.9--60.8\% overall pairwise win rates) and rate it comparable to the proprietary Nano Banana~2.1 (49.7\%), yet SimShoot outperforms it on all six automatic aesthetic metrics in targeted reshooting.
SimShoot achieves the highest instruction completion under multi-VLM judges, while human raters judge its completion comparable to the strongest editors.
The code will be released.
\end{abstract}

%% file: sec/1_intro.tex
\section{Introduction}
\label{sec:intro}

The aesthetic appeal of a portrait is largely shaped by the subject's pose and expression, camera viewpoint, lighting, and composition.
Conventional photo retouching~\cite{gharbiDeepBilateralLearning2017,huExposureWhiteBoxPhoto2018,chenDeepPhotoEnhancer2018,liangPPR10KLargeScalePortrait2021} primarily adjusts color, tone, and local surface appearance, offering little control over the geometric, postural, and optical choices established at capture time.
Recent generative image editors~\cite{openaiChatGPTImages252026,googleNanoBanana212026,qwenQwenImage212026,bfl2026flux3} make these capture-time decisions editable from a single photograph.
Building on \emph{Aesthetic Photo Reconstruction}~\cite{duAesFormerTransformEveryday2026}, we study capture-time aesthetic enhancement for single-person portraits at medium or close range, a task we term \emph{Aesthetic Portrait Reshooting}.
This setting restricts edits to adjustments available during an actual photoshoot, namely camera control, body pose, facial expression, and lighting, while preserving subject identity, wardrobe, and scene content.

Improving a single-image aesthetic score does not by itself solve portrait reshooting: post-capture stylistic grading, contrast boosting, or skin retouching can raise generic appeal scores without realizing a genuine photographic adjustment.
Recent aesthetic agents and guidance models help identify promising compositional or postural edits~\cite{duAesFormerTransformEveryday2026,yaoPhotoAgentExploratoryVisual2026,youPhotoFramerMultimodalImage2026,zengSmartPhotoCrafterUnifiedReasoning2026}, yet two execution challenges remain.
First, requested photographic changes, especially compound spatial, pose, or optical instructions in targeted reshooting, must be executed faithfully without altering unrelated attributes.
Second, single-pass image editors collapse each instruction into a single discrete output, leaving execution extent hard to calibrate.
An instruction such as ``zoom in slightly'' or ``widen the aperture'' admits a continuum of plausible strengths: too little leaves the composition weak, whereas too much crops essential context or blurs meaningful surroundings.
Prompting an image editor with numeric percentages or vague modifiers rarely yields a calibrated progression, forcing repeated trial-and-error regeneration.

A real portrait shoot resolves both challenges through continuous observation and interaction.
Rather than specifying a single static endpoint upfront, a photographer suggests one adjustment at a time, watches the subject or camera move continuously, says ``hold there'' when the framing, posture, or light falls into place, and decides the next adjustment from the resulting shot.
Inspired by this, we reformulate portrait reshooting from single-pass image editing into \emph{interactive photographic action planning with a video action simulator}.

Advances in open-weight video diffusion models (VDMs) such as MiniMax H3~\cite{minimaxMiniMaxH3Open2026,chenH3WorldTurningLanguage2026} make this paradigm practical.
We introduce \textbf{SimShoot}, a closed-loop reshooting framework that coordinates a vision-language model (VLM) planner and selector with a VDM action simulator post-trained on single-action photographic transitions.
At each round, the planner proposes one structured photographic action from the current portrait, optional user instructions, and cross-round feedback.
Conditioned on the current portrait, the simulator rolls out a continuous transition video depicting the isolated progression of that adjustment.
Drawing on video-based image manipulation~\cite{rotsteinPathwaysImageManifold2025,wuChronoEditTemporalReasoning2025}, joint spatiotemporal modeling encourages coherent intermediate states along the trajectory.
A two-stage coarse-to-fine selector then inspects candidate frames across execution extents, optionally zooming into promising local intervals at zero additional video generation cost, to pick a well-balanced frame or retain the current portrait and turn continuous extent tuning into grounded visual comparison.

Our main contributions are threefold:
\begin{itemize}
    \item We formulate Aesthetic Portrait Reshooting as closed-loop photographic action planning with a video action simulator, decoupling \emph{which} shooting adjustment to make from \emph{how far} to execute it via visual frame selection.
    \item We construct single-action video training data across 65 photographic primitives and 98 variants through three complementary routes (\textbf{BlenderSynth} for rendered 3D camera trajectories, \textbf{RealSynth} for endpoint-conditioned transitions, and linear-light synthesis for portrait relighting), combined with progress-aware temporal sampling.
    \item We present \textbf{SimShoot} and evaluate it on our 240-portrait \textbf{AesReShot-Eval} benchmark across automatic aesthetic and background metrics, three VLM instruction judges, a 30,480-response crowd study, and full-benchmark closed-loop ablations, demonstrating consistent gains in portrait aesthetics, instruction adherence, and extent control.
\end{itemize}

%% file: sec/2_related.tex
\section{Related Work}
\label{sec:related}

\noindent\textbf{Aesthetic Photo Reconstruction and Reshooting.}
Aesthetic photo reconstruction revisits capture-time choices such as composition, viewpoint, and subject posture to improve an existing photograph.
AesFormer~\cite{duAesFormerTransformEveryday2026} formalizes this problem with a two-stage pipeline in which AesThinker diagnoses aesthetic weaknesses and plans photographic adjustments that AesEditor synthesizes into a reconstructed image.
Related efforts explore capture-time guidance and controllable re-rendering from complementary angles: Venus~\cite{duVenusBenchmarkingEmpowering2026} studies aesthetic critique and cropping, ShutterMuse~\cite{liShutterMuseCaptureTimePhotography2026} suggests framing and pose adjustments during live capture, Before the Shutter~\cite{jiangShutterAestheticActionable2026} plans and renders portrait shots in prepared 3D environments, and ReX-Shot~\cite{zhangReXShotSingleImageRephotography2026} enables viewpoint, focal-length, and optical-effect adjustments from a single photograph.
Our work focuses on single-person portrait reshooting within a principled photoshoot action space, addressing both faithful multi-round execution and fine-grained control over edit extent.

\noindent\textbf{Instruction-Guided and Agentic Image Editing.}
Instruction-based image editors support diverse visual modifications from natural-language prompts~\cite{brooksInstructPix2PixLearning2023,sheyninEmuEditPrecise2024,liuStep1XEditPracticalFramework2025}, advancing rapidly across proprietary~\cite{openaiChatGPTImages252026,googleNanoBanana212026} and open-weight models~\cite{qwenQwenImage212026,labsFLUX1KontextFlow2025,blackforestlabsFLUX2Klein2026,bfl2026flux3}.
To improve multi-step reasoning for aesthetic enhancement, agentic systems couple multimodal planners with image-editing backbones~\cite{zengSmartPhotoCrafterUnifiedReasoning2026,wuRetouchIQMLLMAgents2026}: PhotoAgent~\cite{yaoPhotoAgentExploratoryVisual2026} iteratively plans edits and critiques generated images, while PhotoFramer~\cite{youPhotoFramerMultimodalImage2026} unifies compositional guidance with exemplar synthesis.
However, invoking single-image editors across sequential rounds exposes two bottlenecks that we directly evaluate in Sec.~\ref{sec:ablations}: discrete text prompts lack a monotonic scale for execution extent, and repeated full-image re-synthesis accumulates texture and structural drift.
A parallel line of work leverages video priors for image editing~\cite{zhangFramePainterEndowingInteractive2025,shenCameraEditorCameraControlledImage2026,zhangAreImagetoVideoModels2025}: Frame2Frame~\cite{rotsteinPathwaysImageManifold2025} selects edited frames along generated video paths, and ChronoEdit~\cite{wuChronoEditTemporalReasoning2025} models temporal reasoning frames toward a target edit.
SimShoot bridges agentic aesthetic planning with specialized single-action video simulation, using coarse-to-fine trajectory selection and closed-loop feedback to execute compound reshooting plans cleanly.

\noindent\textbf{Controllable Video Generation as an Action Simulator.}
Controllable video diffusion models enable explicit camera trajectories and object motion control~\cite{wangMotionCtrlUnifiedFlexible2023,heCameraCtrlEnablingCamera2024,baiReCamMasterCameraControlledGenerative2025,shuaiFreeFormMotionControl2025,fu3DTrajMasterMastering3D2025}.
LAMP~\cite{kizilLAMPLanguageAssistedMotion2026} translates natural-language requests into a structured motion DSL specifying 3D camera and object paths.
Beyond geometric motion, video diffusion has also been adapted to model continuous illumination changes~\cite{zhouLightMoverGenerativeLight2026,guoLightCrafterPBRConditionedVideo2026}, such as GenLit~\cite{bharadwajGenLitReformulatingSingleImage2025} for single-image relighting.
Rather than targeting a single modality, our VDM action simulator unifies 3D camera movement, optical depth of field, human body pose, facial expression, and portrait lighting under a shared structured action interface.

%% file: sec/3_task.tex
\section{Task Formulation}
\label{sec:task}
\label{sec:task-definition}
\label{sec:action-representation}

Building on Aesthetic Photo Reconstruction~\cite{duAesFormerTransformEveryday2026}, we study aesthetic enhancement restricted to adjustments achievable during a real portrait photoshoot, focusing on single-person portraits captured at medium or close range.
The permitted photographic action space $\mathcal{A}$ spans four categories (\emph{Camera}, \emph{Pose}, \emph{Expression}, and \emph{Lighting}), while explicitly excluding post-capture cropping, aspect-ratio changes, global color grading, beauty retouching, and scene object insertion or removal.
Given an input portrait $I_0$ and an optional user goal $u$, let $\mathcal{R}_{\mathcal{A}}(I_0,u)$ denote the set of candidate portraits reachable through valid photoshoot actions in $\mathcal{A}$ that satisfy $u$ when specified (while preserving subject identity, clothing, and non-target scene attributes).
We seek a reshoot
\begin{equation}
    I^\star \in
    \operatorname*{arg\,max}_{I \in \mathcal{R}_{\mathcal{A}}(I_0,u)}
    S_{\mathrm{aes}}(I),
    \label{eq:reshooting-objective}
\end{equation}
where $S_{\mathrm{aes}}$ measures portrait aesthetic quality.
Under \emph{open-ended reshooting}, $u$ specifies only a shared photographic improvement goal and the system autonomously determines which actions to take; under \emph{targeted reshooting}, $u$ prescribes a portrait-specific set of requested adjustments and preservation constraints.

\paragraph{Photographic action DSL.}
To provide an unambiguous interface between high-level VLM planning and low-level video simulation, we represent single-step adjustments in a structured \emph{domain-specific language (DSL)}.
Rather than modeling arbitrary unconstrained video motion, we organize task-relevant photoshoot operations into semantically distinct \emph{DSL primitives}.
Each structured action $a=(p,\theta)$ pairs a primitive $p$ with typed directional or spatial parameters $\theta$, such as \texttt{camera.\allowbreak zoom(direction=\allowbreak in)}, \texttt{camera.\allowbreak aperture(direction=\allowbreak wider)}, or \texttt{light.\allowbreak fill(direction=\allowbreak stronger)}.
Unlike post-capture cropping, optical camera primitives such as bidirectional \texttt{camera.zoom} synthesize full-canvas focal-length framing changes, including zooming out to reveal wider surroundings without losing spatial resolution.

To eliminate left--right ambiguity between the photographed person and the viewer, the DSL separates two coordinate conventions: body-pose and unilateral facial-expression parameters use \emph{subject-centric} anatomical sides (the person's own left and right), whereas lateral/vertical camera reframing, gaze targets, and directional key lighting use viewer \emph{image-plane} directions, complemented by subject-relative 3D orbit, elevation, and depth conventions.

\paragraph{Hybrid closed-set and open-vocabulary execution.}
Each structured action compiles into a canonical natural-language description $c(a)$ that specifies the target operation alongside explicit preservation and temporal-continuity clauses.
At inference, \emph{Camera} and \emph{Lighting} expose $25$ closed-set parameterized variants covering core optical and studio/natural illumination controls.
Because human posture and facial nuance are inherently open-ended, \emph{Pose} and \emph{Expression} expose two open-vocabulary slots (\texttt{pose.open} and \texttt{expression.open}) backed by $73$ trained variants---resolving exact matches directly to the trained primitive while wrapping free-form single-action descriptions in the same preservation structure.

%% file: sec/4_dataset.tex
\section{Dataset Crafting}
\label{sec:dataset}

\subsection{Motivation and Data Requirements}
\label{sec:data-motivation}
For a video generator to serve as a reliable photographic action simulator, each rollout must satisfy two properties: \emph{faithful action execution}, realizing the requested camera, pose, expression, or lighting adjustment in the specified direction, and \emph{action isolation}, holding identity, non-target attributes, and unrelated scene elements steady while permitting natural physical consequences of the edit.
Off-the-shelf pretrained VDMs frequently ignore specialized optical or directional commands, entangle subject motion with camera drift, or stall with minimal extent.
We therefore construct curated single-action transition videos for targeted post-training.
Crucially, this supervision teaches \emph{how} an isolated photographic action unfolds visually rather than \emph{whether} its endpoint is aesthetically superior: deciding which actions improve a given portrait and how far to carry them out is delegated to the closed-loop planner and selector at inference.

\subsection{Single-Action Video Construction}
\label{sec:data-construction}
We construct a multi-category training corpus
\begin{equation}
    \mathcal{D}=\{(I_{i,0},\,c(a_i),\,V_i)\}_{i=1}^{N},
    \label{eq:dataset-crafting-samples}
\end{equation}
where $a_i$ is a DSL primitive action (Sec.~\ref{sec:task}), $c(a_i)$ is its compiled description, and $V_i=(I_{i,0},\ldots,I_{i,T-1})$ is a $T=39$-frame transition clip starting from $I_{i,0}$.
As illustrated in Figure~\ref{fig:dataset-construction}, no single data source can supply accurate 3D camera geometry, diverse human articulation, optical defocus, and drift-free relighting simultaneously; we therefore develop three complementary synthesis routes.

\paragraph{BlenderSynth: physically based 3D camera rendering.}
To obtain exact, disentangled 3D camera motion priors free of subject movement, we place 20 rigid objects from Objaverse~\cite{deitkeObjaverseUniverseAnnotated2023} across 10 indoor and outdoor 3D scenes and render bidirectional camera orbit, lateral/vertical/depth translation, and fixed-position focal-length zoom trajectories in Blender Cycles while keeping scene geometry, materials, and lighting fixed.
Automated geometric ray-casting checks filter out camera--geometry collisions, unintended occlusions, and framing violations, followed by visual preview screening before full trajectory rendering.
Despite containing synthetic rigid scenes rather than human subjects, these trajectories isolate pure camera-to-scene projective geometry that transfers zero-shot to real portraits on a pretrained video backbone.

\paragraph{RealSynth: endpoint-conditioned portrait transitions.}
For human pose, facial expression, and optical depth of field, we synthesize continuous transitions from verified start--end image pairs.
Starting from single-person portraits in PortraitCraft~\cite{lou1stPortraitCraftChallenge2026} screened for part visibility and resolution (including close-up crops so subtle facial changes remain clear), we generate candidate edited endpoints using ChatGPT Images~2.5 (Sunburst)~\cite{openaiChatGPTImages252026} and modern portrait editors across the pose, expression, and shallow-focus pools.
Every candidate pair undergoes separate automated and VLM verification for (i)~clear completion of the target action in the intended direction and (ii)~strict preservation of identity, wardrobe, non-target body parts, and background; non-compliant pairs are repaired or discarded.
For depth of field (\texttt{camera.aperture}), we combine verified portrait aperture pairs with real multi-aperture scene pairs ($f/22 \leftrightarrow f/2$) from RealBokeh~\cite{seizingerBokehliciousPhotorealisticBokeh2025}, including temporally reversed clips to cover both aperture widening and narrowing.
Each verified image pair conditions pretrained MiniMax H3~\cite{minimaxMiniMaxH3Open2026} in its first-and-last-frame-to-video-and-audio (FL2VA) mode (with the audio branch disabled) alongside $c(a_i)$ to synthesize a continuous intermediate transition.

\paragraph{Linear-light relighting synthesis.}
Interpolating illumination changes directly with a video generator often introduces mid-clip exposure surges or subtle facial drift.
For lighting primitives spanning color temperature, fill, rim, background illumination, and directional key-light balance, we instead pair PortraitCraft sources with VLM-verified relit endpoints and blend each sRGB-decoded pair along a cubic smoothstep curve in linear light, producing clean illumination transitions free of interpolation drift between aligned endpoints.

\begin{table}[t]
\centering
\footnotesize
\captionsetup{skip=4pt}
\caption{\textbf{Dataset statistics.} P/V counts trained primitives and variants in \texttt{Train}. Split rows report curated 39-frame clips.}
\label{tab:dataset-crafting-statistics}
\setlength{\tabcolsep}{1.5pt}
\renewcommand{\arraystretch}{0.96}
\begin{tabular*}{\columnwidth}{@{\extracolsep{\fill}}l*{8}{c}@{}}
\toprule
 & \multicolumn{4}{c}{Camera} & \multirow{2}{*}{Pose} & \multirow{2}{*}{Expression} & \multirow{2}{*}{Lighting} & \multirow{2}{*}{\textbf{Total}} \\
\cmidrule(lr){2-5}
 & Orbit & Transl. & Zoom & DoF & & & & \\
\midrule
P/V & 2/4 & 3/6 & 1/2 & 1/2 & 29/46 & 23/27 & 6/11 & \textbf{65/98} \\
Train & 176 & 264 & 88 & 174 & 191 & 220 & 145 & \textbf{1,258} \\
Val. & 16 & 24 & 8 & 4 & 31 & 33 & 26 & \textbf{142} \\
Test & 16 & 24 & 8 & 4 & 30 & 26 & 25 & \textbf{133} \\
\bottomrule
\end{tabular*}
\end{table}

\subsection{Verification and Progress-Aware Sampling}
\label{sec:data-preparation}
Synthesized clips undergo automated quality checks, VLM clip review, and human auditing to remove temporal flicker, mid-trajectory artifacts, or unintended drift.
Because VDM-interpolated transitions often exhibit stationary endpoint holds and uneven motion speeds, uniform subsampling wastes frames on pauses and compresses the active transition.
We therefore apply \emph{progress-aware temporal sampling}~\cite{zhiMGSamplerExplainableSampling2021} to variable-speed clips (Figure~\ref{fig:dataset-construction}(d)): we track cumulative frame-to-frame luminance change across the active clip and select $T=39$ strictly increasing frame indices at approximately equal increments of visual progression while pinning the exact initial and terminal frames.
Analytically rendered camera trajectories and tail-trimmed portrait aperture clips are sampled uniformly to 39 frames, while temporally normalized expression clips are used directly at 39 frames.

\subsection{Dataset Statistics}
\label{sec:data-statistics}
Table~\ref{tab:dataset-crafting-statistics} summarizes the resulting dataset across seven adapter groups, totaling 65 trained DSL primitives, 98 parameterized variants, and 1,533 curated 39-frame clips (1,258 \texttt{Train}, 142 \texttt{Val.}, and 133 \texttt{Test}).
Within each release, splits are partitioned by 3D object (BlenderSynth), physical scene (RealBokeh), or canonical source portrait and subject identity group so derived crops and opposite directions remain within the same split, and hash verification confirms zero source-image overlap with our evaluation benchmark.

%% file: sec/5_method.tex
\section{Method}
\label{sec:method}

Figure~\ref{fig:simshoot-workflow} gives an overview of \textbf{SimShoot}.
Rather than predicting a final edited image in a single opaque jump, SimShoot decouples \emph{choosing a photographic action} from \emph{choosing its execution extent} inside an iterative closed loop: a VLM planner proposes one structured action at a time, a specialized VDM action simulator generates a continuous 39-frame visual rollout, and a two-stage VLM selector inspects intermediate frames to commit to an appropriate extent or retain the current portrait before returning feedback to the planner.

\subsection{VDM Action Simulator}
\label{sec:action-simulator}

Given portrait $I_t$ at round $t$ and proposed DSL action $a_t$ (Sec.~\ref{sec:task}), the simulator synthesizes a $T$-frame transition
\begin{equation}
    V_t = (\hat I_{t,0},\ldots,\hat I_{t,T-1})
    = \mathcal{S}_{\Theta,\phi_{g(a_t)}}(I_t,c(a_t)),
    \label{eq:action-simulation}
\end{equation}
where $c(a_t)$ is the compiled action prompt, $\Theta$ denotes the frozen pretrained VDM backbone parameters, and $\phi_{g(a_t)}$ is the category-specific adapter selected by $a_t$.
Successive frames of $V_t$ trace a continuous trajectory from $I_t$ across increasing execution extents, exposing intermediate outcomes for visual inspection before selection.

We instantiate the simulator on MiniMax H3~\cite{minimaxMiniMaxH3Open2026} and specialize it via seven independent low-rank adaptation (LoRA)~\cite{huLoRALowRankAdaptation2022} modules corresponding to our action groups: \emph{camera orbit}, \emph{camera translation}, \emph{camera zoom}, \emph{depth of field}, \emph{body pose}, \emph{facial expression}, and \emph{lighting}.
All seven adapters share a single frozen backbone, and only the adapter $\phi_{g(a_t)}$ matching the requested action family is activated during a rollout, preventing cross-modality interference between geometric camera motion, human articulation, and illumination changes.
Closed-set Camera and Lighting actions use their exact canonical descriptions, whereas open-vocabulary Pose and Expression actions insert the planner's single-action description into the corresponding preservation template and route it to the \texttt{pose} or \texttt{expression} adapter.

Each adapter $\phi_g$ is trained on its curated single-action subset $\mathcal{D}_g$ from Sec.~\ref{sec:dataset}.
Given a training tuple $(I, c(a), V)$, let $z=E(V)$ be the latent encoding of the target clip under the frozen video VAE.
Sampling Gaussian noise $\epsilon\sim\mathcal{N}(0,\mathbf{I})$ and flow timestep $\sigma$, we form $z_\sigma=(1-\sigma)z+\sigma\epsilon$ and minimize the conditional flow-matching loss~\cite{lipmanFlowMatchingGenerative2023}:
\begin{equation}
    \mathcal{L}_g =
    \mathbb{E}\!\left[
      \left\|v_{\Theta,\phi_g}(z_\sigma,\sigma;I,c(a))
      -(\epsilon-z)\right\|_2^2
    \right],
    \label{eq:simulator-training}
\end{equation}
where $v_{\Theta,\phi_g}$ predicts the conditional velocity field, the expectation runs over $(I,c(a),V)\sim\mathcal{D}_g$, $\sigma$, and $\epsilon$, and only $\phi_g$ is updated.
Although some training clips are synthesized offline using endpoint pairs, adapter training and online inference condition \emph{exclusively} on the single starting portrait ($I$ during training, $I_t$ at inference) and action text $c(a)$ without requiring any target endpoint image.

\subsection{Two-Stage Action-Extent Selection}
\label{sec:action-extent}

Even when an action is simulated faithfully, its optimal extent is context-dependent: zooming in or turning the head may strengthen composition midway through the clip before becoming overly tight or unnatural near the terminal frame \texttt{F38}.
Conversely, a requested posture or camera shift may reach completion only near the end of the rollout.
SimShoot casts this continuous calibration problem as two-stage visual selection over $0$-based indices \texttt{F0}--\texttt{F38}:

\begin{itemize}
    \item \textbf{Coarse trajectory inspection.}
    We replace index zero with the exact unchanged portrait $I_t$ (\texttt{F0}) so keeping \texttt{F0} preserves $I_t$ without VAE reconstruction loss.
    The VLM selector first inspects five evenly spaced frames (\texttt{\{F0, F10, F19, F29, F38\}}) paired with enlarged face crops to make subtle expression, gaze, and sharpness differences clearly legible (Figure~\ref{fig:simshoot-workflow}).
    \item \textbf{Zero-video-cost interval refinement.}
    When the coarse sheet indicates that the sweet spot lies between two adjacent coarse candidates (\eg, \texttt{F10} under-executes a subtle smile while \texttt{F19} overshoots), the selector can open that local interval to inspect evenly spaced in-between frames already present in $V_t$ (such as \texttt{F12, F15, F17}) before making its final choice, achieving fine-grained extent control with zero additional video generation calls.
\end{itemize}

Comparing candidates against both the current state $I_t$ and the original input $I_0$, the selector evaluates target action progress, aesthetic balance, identity and scene preservation, and cumulative drift across rounds, supplemented by an automated face-quality safeguard using TOPIQ~\cite{chenTOPIQTopDownApproach2024} to prevent severe facial degradation.
If no candidate improves the shot within constraints, the selector picks \texttt{F0} and retains $I_t$.

\subsection{Closed-Loop Planning and Reshooting}
\label{sec:iterative-reshooting}

Complex portrait reshoots typically coordinate multiple adjustments, such as reframing the camera, relaxing a stiff hand pose, softening expression, and balancing fill light.
Executing all changes in one prompt often causes edits to interfere or drop clauses, whereas open-loop execution of a fixed script cannot adapt as earlier edits reshape the composition.

In SimShoot, the VLM planner operates closed-loop over rounds $t=0,1,\ldots$ conditioned on $I_0$, the latest accepted portrait $I_t$, the task goal $u$, and a structured memory of prior attempts recording accepted frames, trajectory summaries, and rejection rationales.
Under open-ended reshooting, the planner critiques the current portrait across composition, pose, expression, optical focus, and lighting to propose high-leverage adjustments; under targeted reshooting, it tracks remaining user-requested edits and attached constraints.
An accepted frame advances the portrait state to $I_{t+1}$, whereas a rejected rollout keeps $I_{t+1}=I_t$ and informs the planner whether to retry with an alternative rollout, reformulate the action (\eg, adding explicit preservation phrasing in an open-vocabulary slot), or move to another adjustment.
The loop terminates voluntarily once the planner determines that no pending plan steps remain or when the edit or attempt budget is reached, returning the final accepted portrait.

%% file: sec/6_experiments.tex
\input{sec/6_experiment_tables}
\hyphenation{reshooting}

\reshootMainResults

\section{Experiments}
\label{sec:experiments}

\subsection{Experimental Setup}
\label{sec:experimental-setup}

\paragraph{AesReShot-Eval benchmark and tasks.}
To evaluate both autonomous selection of beneficial shooting adjustments and faithful execution of user-specified ones, we construct \textbf{AesReShot-Eval} from $240$ high-quality single-person portraits selected from AesRecon~\cite{duAesFormerTransformEveryday2026} on a shared $768\times1120$ canvas, split evenly into two disjoint tracks ($120$ portraits each).
\textbf{Open-ended reshooting} supplies a shared prompt requesting a more aesthetically pleasing, natural-looking portrait within the photoshoot action space, leaving the choice and strength of adjustments entirely to each method.
\textbf{Targeted reshooting} provides each portrait with $1$--$4$ coordinated photographic requests ($308$ finalized actions in total, mean $2.57$ per portrait), partitioned into $60$ \emph{fixed-capture} portraits ($134$ pose or expression actions with camera framing, focus, depth of field, and lighting held fixed) and $60$ \emph{broader-scope} portraits ($50$ camera/depth-of-field, $99$ pose/expression, and $25$ lighting actions).
Both tracks require preserving subject identity, body shape, hairstyle, clothing, and non-target scene elements without beauty retouching or stylistic re-grading.

\paragraph{Baselines and metrics.}
We compare SimShoot against five representative baselines: the proprietary Nano Banana~2.1~\cite{googleNanoBanana212026}, the open-weight image editors Qwen-Image~2.1~\cite{qwenQwenImage212026} and FLUX~3 Klein~\cite{bfl2026flux3} (using their official prompt enhancers), the video-based image editor ChronoEdit~\cite{wuChronoEditTemporalReasoning2025}, and the two-stage reshooting model AesFormer~\cite{duAesFormerTransformEveryday2026}.
All methods receive identical $768\times1120$ inputs and are evaluated at full canvas resolution across four dimensions:
(1)~\emph{Aesthetic improvement}: relative mean-score gain (\%) from input to output on the Structure, Composition, and Emotion dimensions of PortraitAes~\cite{xie2026portraitaes} plus PhotoFramer-Assessment~\cite{youPhotoFramerMultimodalImage2026}, Q-Align~\cite{wuQAlignTeachingLMMs2023}, and ArtiMuse~\cite{caoArtiMuseFineGrainedImage2025};
(2)~\emph{Targeted instruction following}: per-portrait completion rates judged independently by GPT-6.1 Sol~\cite{openaiGPT61Sol2026}, Gemini~3.8 Flash~\cite{googleGemini38Flash2026}, and Qwen3.8 Max~\cite{qwenQwen38Max2026} over the $118/120$ portraits with valid verdicts from all three judges (two portraits encountered an upstream Gemini safety block);
(3)~\emph{Background preservation}: PSNR, SSIM~\cite{wang2004ssim}, LPIPS~\cite{zhangUnreasonableEffectivenessDeep2018}, and mean absolute CIELAB lightness shift $\lvert\Delta L^*\rvert$ on foreground-excluded backgrounds (segmented via SAM~3~\cite{carion2025sam3}) across the $60$ fixed-capture targeted portraits; and
(4)~\emph{Crowd-sourced human evaluation}: $12{,}000$ pairwise preference votes and $18{,}480$ per-action completion judgments on Rapidata.

\paragraph{Implementation summary.}
We train seven rank-$16$ ($\alpha=16$) LoRA adapters on a frozen MiniMax H3~\cite{minimaxMiniMaxH3Open2026} backbone.
At inference, each action generates a 39-frame transition at 24~fps on NVIDIA A100 GPUs, and GPT-6.1 Sol serves as both VLM Planner and two-stage Selector.

\subsection{Main Aesthetic Results}
\label{sec:main-results}
Table~\ref{tab:main-results} reports relative aesthetic gains across both tracks.
In open-ended reshooting, SimShoot achieves the strongest PortraitAes-Structure gain ($+15.47\%$) and ranks first on PhotoFramer ($+5.10\%$) and Q-Align ($+10.15\%$), outperforming AesFormer, ChronoEdit, and FLUX~3 Klein on all six metrics and Qwen-Image~2.1 on five.
Nano Banana~2.1 leads on Composition, Emotion, and ArtiMuse, whereas SimShoot leads across Structure, PhotoFramer, and Q-Align.

In targeted reshooting, where aesthetic gains must come from prescribed shooting changes rather than free-form retouching, SimShoot leads all three PortraitAes dimensions, widening its Structure lead to $+10.16\%$ vs.\ $+5.85\%$ for Nano Banana~2.1 and surpassing it on all six metrics.
On the general-purpose single-image scorers, SimShoot ranks second on PhotoFramer ($+2.78\%$, behind AesFormer) and third on Q-Align ($+3.51\%$, behind AesFormer and ChronoEdit) and ArtiMuse ($+1.77\%$, behind Qwen-Image~2.1 and AesFormer); as Sec.~\ref{sec:human-evaluation} confirms, general-purpose scorers often reward contrast grading or background smoothing even when requested edits fail, whereas PortraitAes closely tracks human preference.

\setcounter{topnumber}{2}\setcounter{totalnumber}{3}
\reshootAdherenceTable
\reshootBackgroundTable
\reshootHumanPreferenceTable
\subsection{Instruction and Scene Preservation}
\label{sec:instruction-following}
\label{sec:preservation}
Table~\ref{tab:adherence} evaluates whether targeted edits actually realize the requested adjustments.
SimShoot ranks first under every judge individually, reaching a mean completion rate of $83.71\%$ and outperforming the strongest baseline, FLUX~3 Klein ($75.40\%$), by $+8.31$\,pp.
Crucially, this advantage does not stem from GPT-6.1 Sol serving as both workflow agent and judge: averaging exclusively over the two external judges (Gemini~3.8 Flash and Qwen3.8 Max), SimShoot still ranks first under each model individually and leads runner-up FLUX~3 Klein by $+5.19$\,pp across all verdicts ($80.37\%$ vs.\ $75.18\%$, or $+6.85$\,pp excluding ambiguous verdicts).

On the $60$ fixed-capture portraits ($134$ pose/expression actions; Table~\ref{tab:background}), SimShoot ranks second to FLUX~3 Klein on all four background metrics ($27.30$\,dB PSNR, $0.7916$ SSIM, $0.1083$ LPIPS, $3.13$ $\lvert\Delta L^*\rvert$), outperforming all other baselines while beating FLUX~3 Klein in completion ($+6.08$\,pp on the $58$ three-judge subset; $+8.31$\,pp overall).

\subsection{Crowd-Sourced Human Evaluation}
\label{sec:human-evaluation}
To complement automatic scorers with direct human judgment, we conducted two blinded Rapidata studies on the exact final images evaluated above (task interfaces in Appendix~\ref{sec:supp-human-visuals}), with portrait-clustered bootstrap intervals and five-baseline Holm correction~\cite{holm1979sequential}.

\paragraph{Photographic preference ($12{,}000$ votes).}
For each of the $240$ portraits and five baselines ($1{,}200$ pairs, $10$ raters per pair), raters viewed the source portrait alongside the unlabeled SimShoot and baseline edits and chose the better photograph.
As shown in Table~\ref{tab:human-preference}, raters significantly prefer SimShoot over ChronoEdit ($60.79\%$), AesFormer ($56.33\%$), FLUX~3 Klein ($55.00\%$), and Qwen-Image~2.1 ($54.92\%$) overall (all Holm $p<0.001$), over three baselines on open-ended reshooting (ChronoEdit, Qwen-Image~2.1, and AesFormer; Holm $p<0.05$, with $53.42\%$ vs.\ FLUX~3 Klein at $p=0.137$), and over all four open-weight and reshooting baselines on targeted reshooting (up to $64.83\%$, Holm $p\le 0.032$), while showing no statistically significant difference from Nano Banana~2.1 ($49.67\%$ overall, Holm $p=0.789$; $51.00\%$ targeted, $p=0.560$).
Comparing Table~\ref{tab:main-results} against human votes explains the metric split: PortraitAes tracks preference closely ($\text{Spearman }\rho=+0.83$--$+1.00$ overall, $+0.49$--$+0.83$ open-ended, $+0.66$--$+0.94$ targeted; $+0.94$ Structure), whereas general scorers show moderate open-ended alignment ($\rho=+0.09$--$+0.66$) yet drop to $\rho\le +0.03$ on targeted ($+0.03$ PhotoFramer/ArtiMuse, $-0.60$ Q-Align).
For example, on the $72.5\%$ of portraits where ArtiMuse scores Qwen-Image~2.1 higher than SimShoot, raters still prefer SimShoot in $56.2\%$ of votes.

\paragraph{Targeted action completion ($18{,}480$ responses).}
Across all $308$ targeted actions and six methods ($10$ responses per item over \emph{Completed}, \emph{Not completed}, \emph{Can't tell}), SimShoot achieves the highest numerical completion ($57.28\%$ portrait-weighted vs.\ $57.10\%$ for FLUX~3 Klein, $56.97\%$ for Nano Banana~2.1, $54.52\%$ for Qwen-Image~2.1, $54.24\%$ for ChronoEdit, and $50.30\%$ for AesFormer; $71.33\%$ on definite responses), remaining statistically comparable to FLUX~3 Klein and Nano Banana~2.1 while significantly outperforming AesFormer (Holm $p<0.001$, plus ChronoEdit on definite responses).
Crowd ($120$ portraits) and VLM ($118$ portraits) completion show exact $6$-of-$6$ rank agreement ($\rho=1.00$, $r=0.93$; $r=0.98$ on definite responses over the $118$ common portraits), with narrower crowd spread from $\approx 20\%$ \emph{Can't tell} clicks on subtle optical shifts and leniency on partial poses.

\setcounter{topnumber}{1}\setcounter{totalnumber}{2}
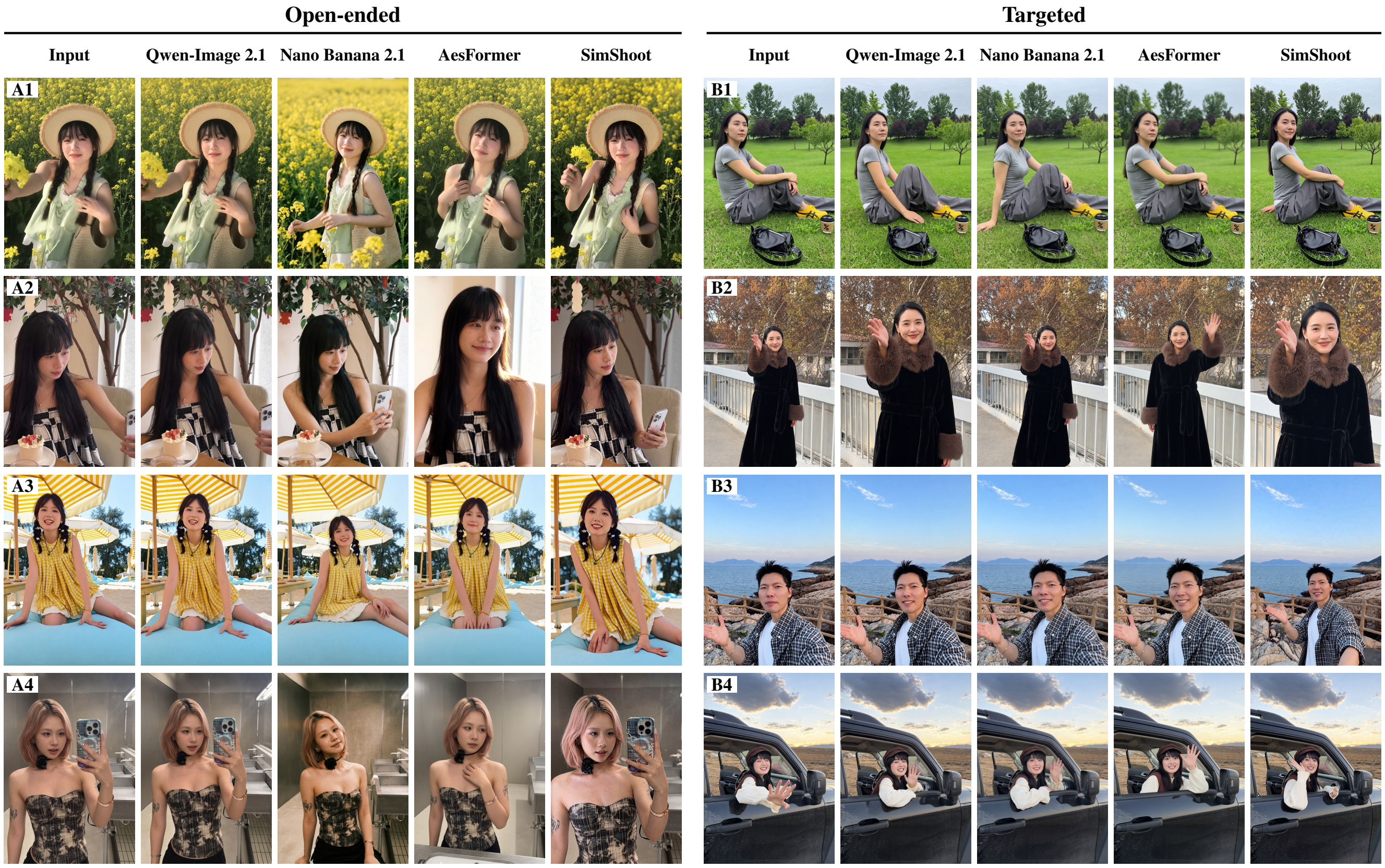
\begin{figure*}[t]
  \centering
  \setlength{\tabcolsep}{1.1pt}
  \captionsetup{skip=2pt}
  \newcommand{\qualgroupgap}{8pt}
  \newcommand{\qualwidth}{\dimexpr(\textwidth-\qualgroupgap-16\tabcolsep)/10\relax}
  \newcommand{\qualhead}[1]{\makebox[\qualwidth][c]{\fontsize{6.2}{7.2}\selectfont\bfseries #1}}
  \newcommand{\qualheaders}{
    \qualhead{Input} &
    \qualhead{Qwen-Image 2.1} &
    \qualhead{Nano Banana 2.1} &
    \qualhead{AesFormer} &
    \qualhead{SimShoot}}
  \newcommand{\quallabeledimg}[2]{
    \begin{tikzpicture}[inner sep=0pt,outer sep=0pt,baseline=(portrait.south)]
      \node[anchor=south west] (portrait) {#1};
      \node[anchor=north west,fill=white,inner xsep=1.6pt,inner ysep=.7pt,
        xshift=1.1pt,yshift=-1.1pt,font=\fontsize{6.5}{7}\selectfont\bfseries]
        at (portrait.north west) {#2};
    \end{tikzpicture}}
  {\footnotesize
  \begin{tabular}{@{}*{5}{c}@{\hspace{\qualgroupgap}}*{5}{c}@{}}
    \multicolumn{5}{@{}c@{\hspace{\qualgroupgap}}}{\textbf{Open-ended}} &
    \multicolumn{5}{c@{}}{\textbf{Targeted}} \\[-1pt]
    \cmidrule[\heavyrulewidth](r{\qualgroupgap}){1-5}\cmidrule[\heavyrulewidth](l{1pt}){6-10}
    \qualheaders & \qualheaders \\[3.5pt]
    \quallabeledimg{\includegraphics[width=\qualwidth]{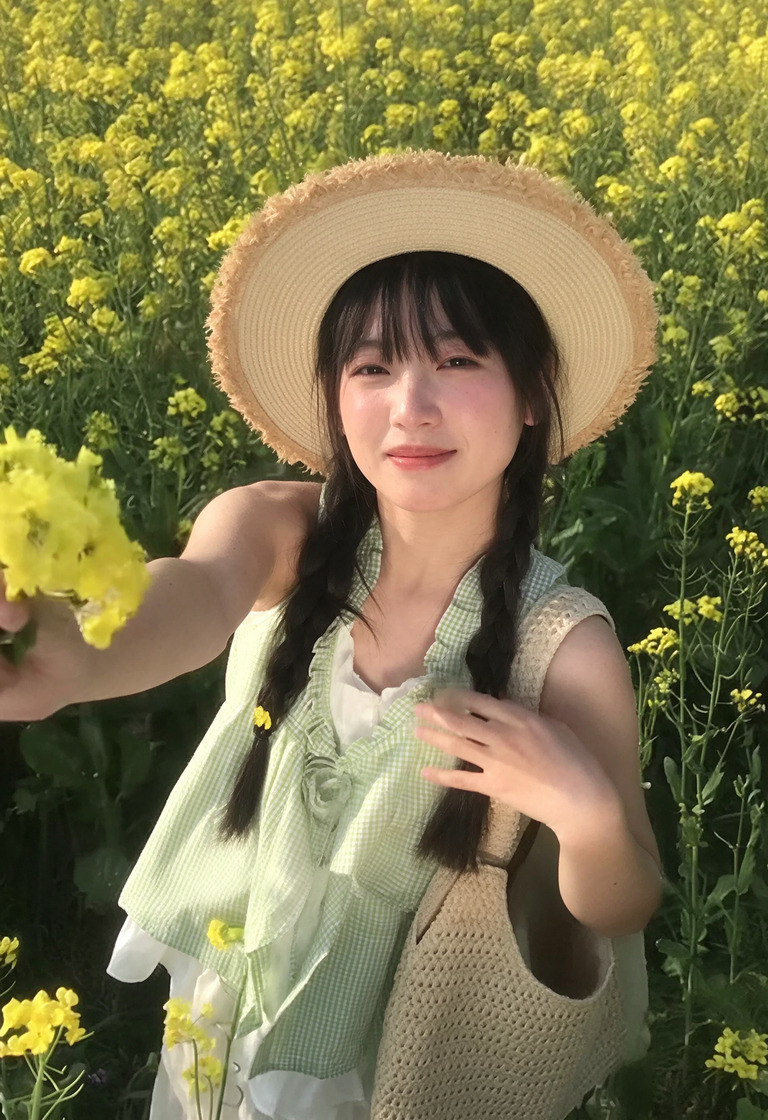}}{A1} & \includegraphics[width=\qualwidth]{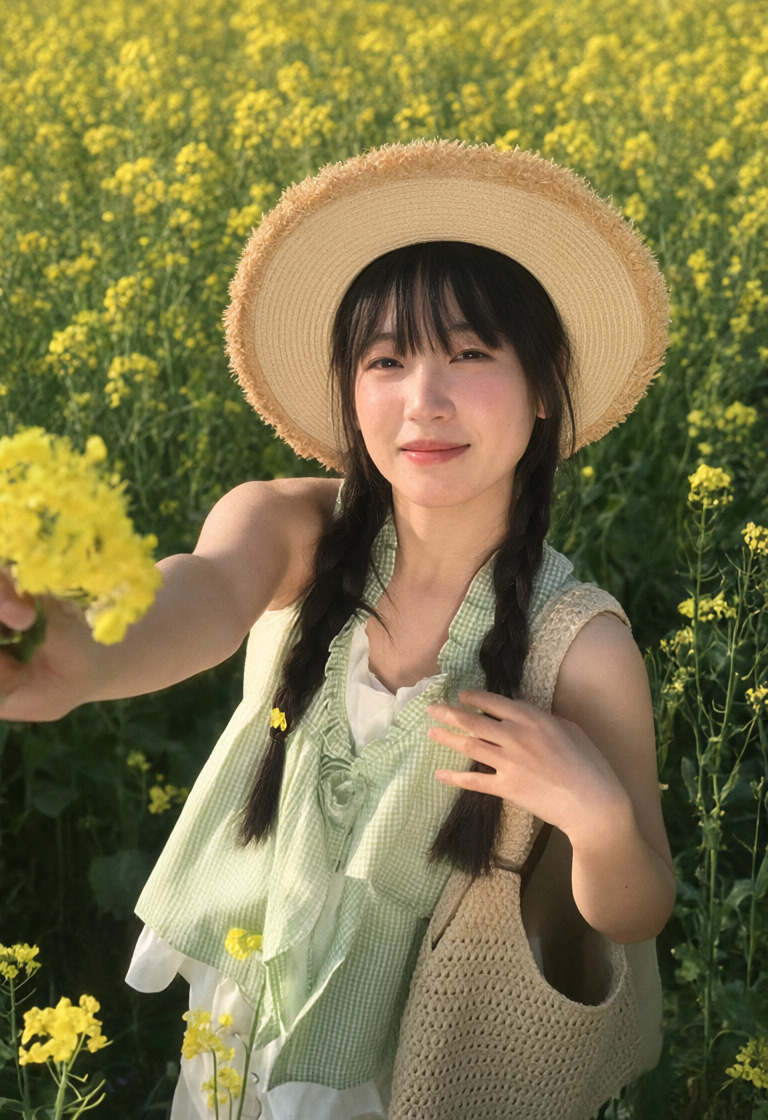} & \includegraphics[width=\qualwidth]{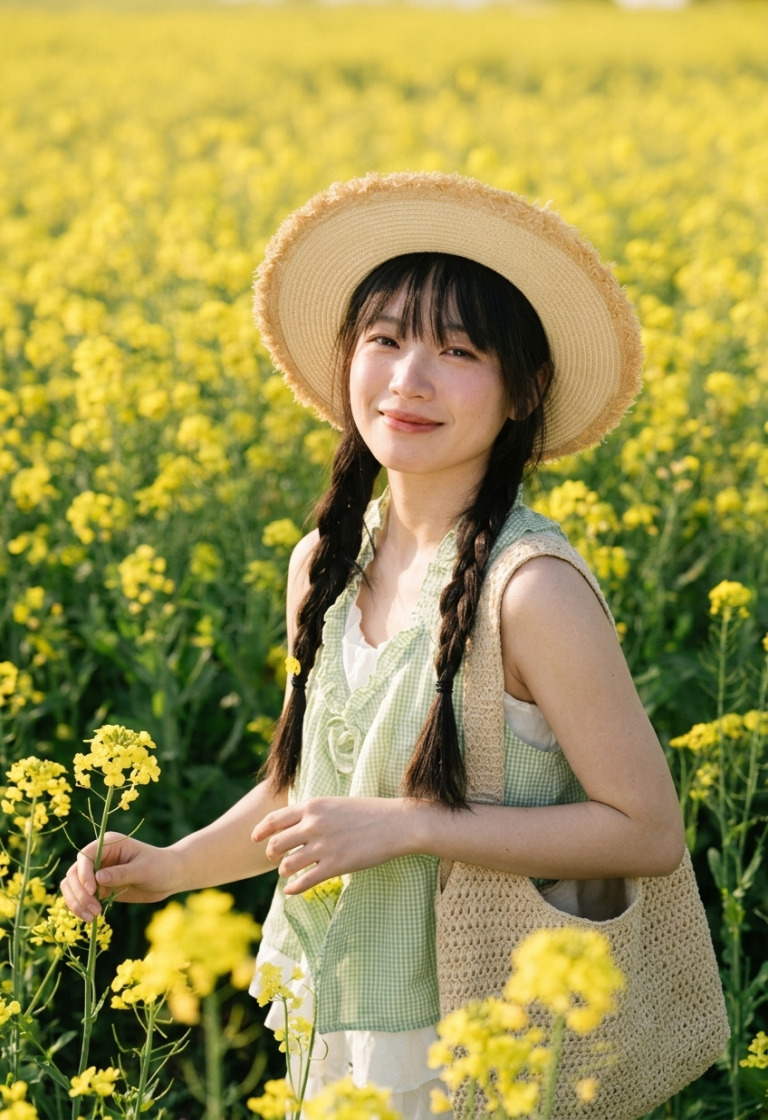} & \includegraphics[width=\qualwidth]{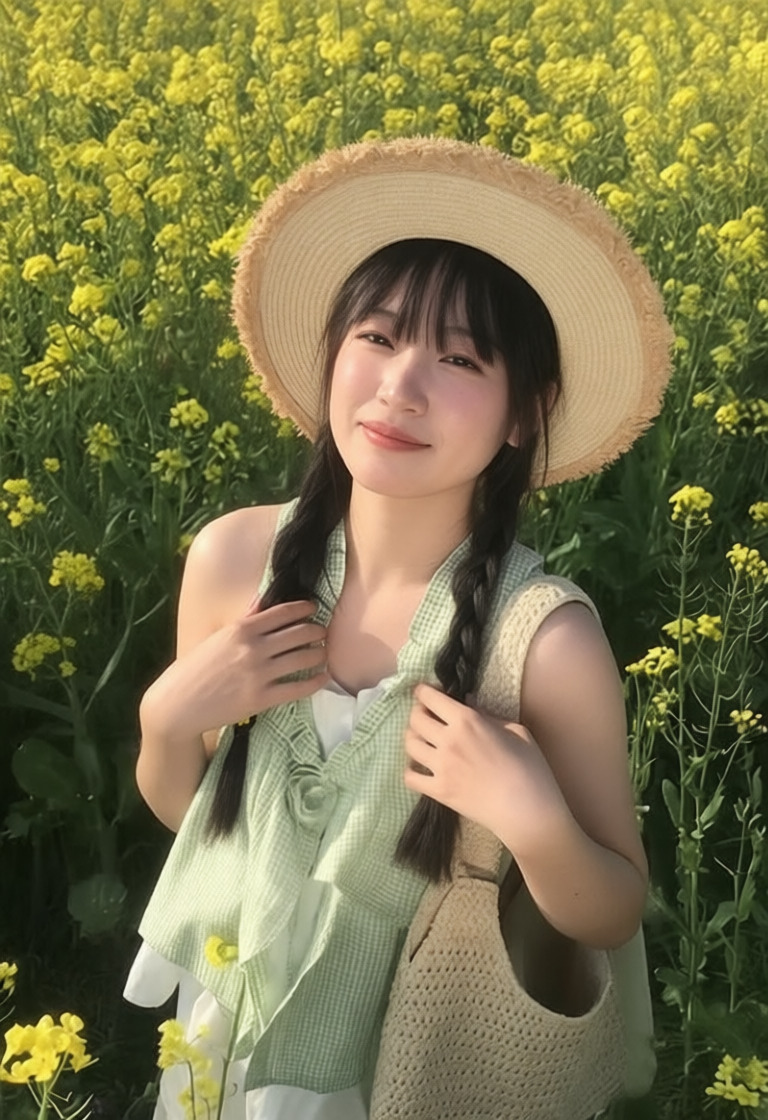} & \includegraphics[width=\qualwidth]{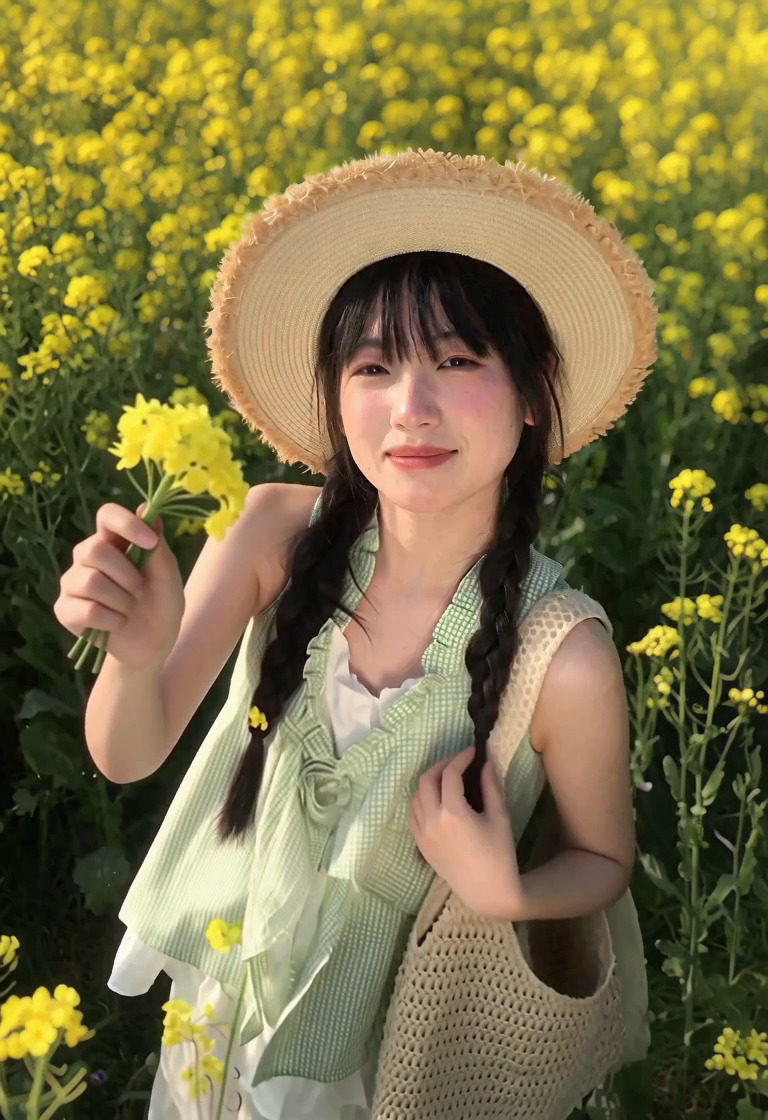} & \quallabeledimg{\includegraphics[width=\qualwidth]{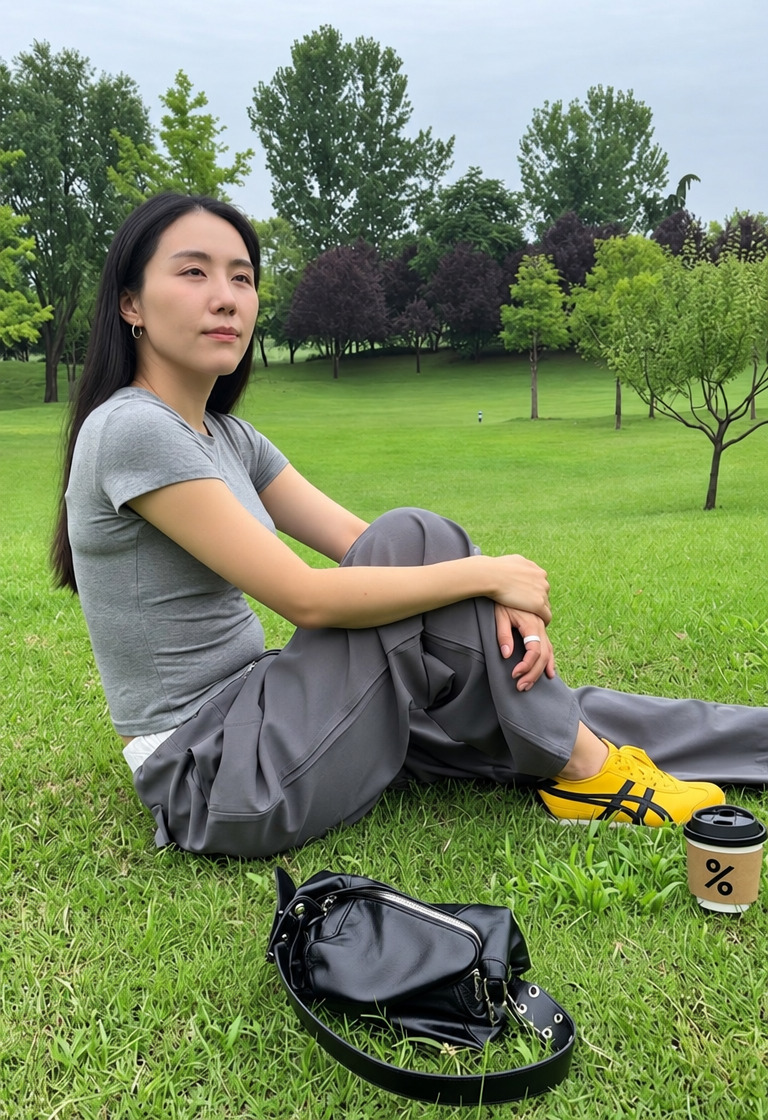}}{B1} & \includegraphics[width=\qualwidth]{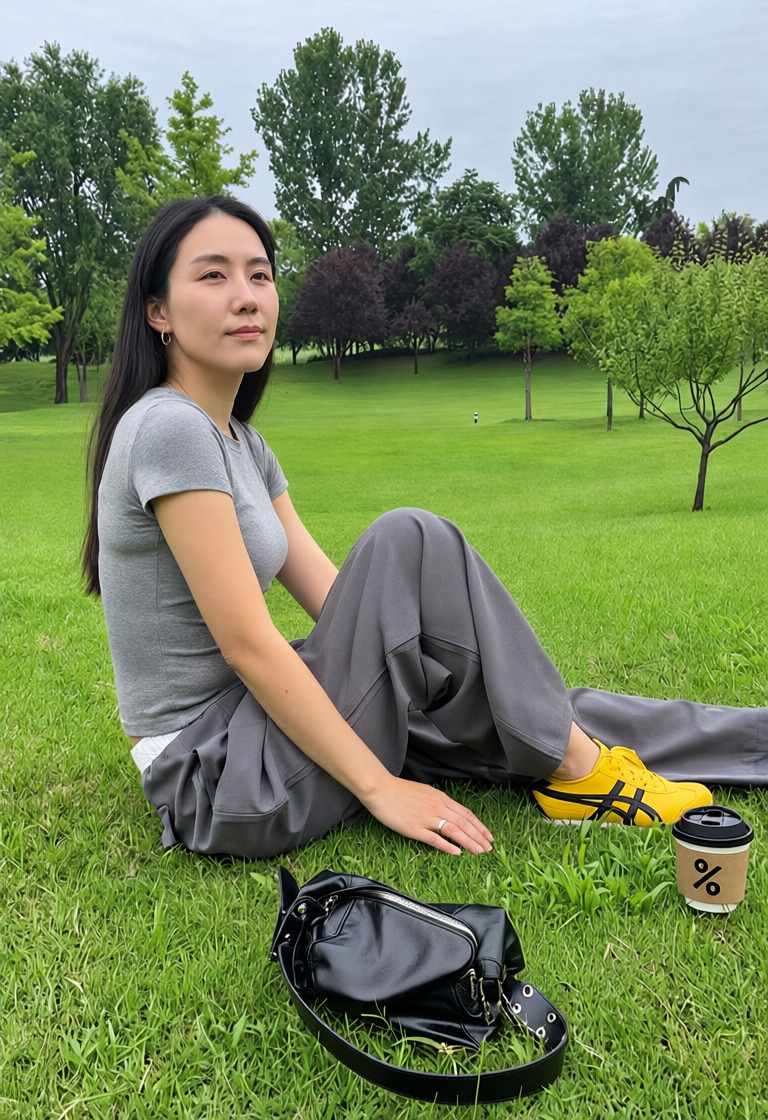} & \includegraphics[width=\qualwidth]{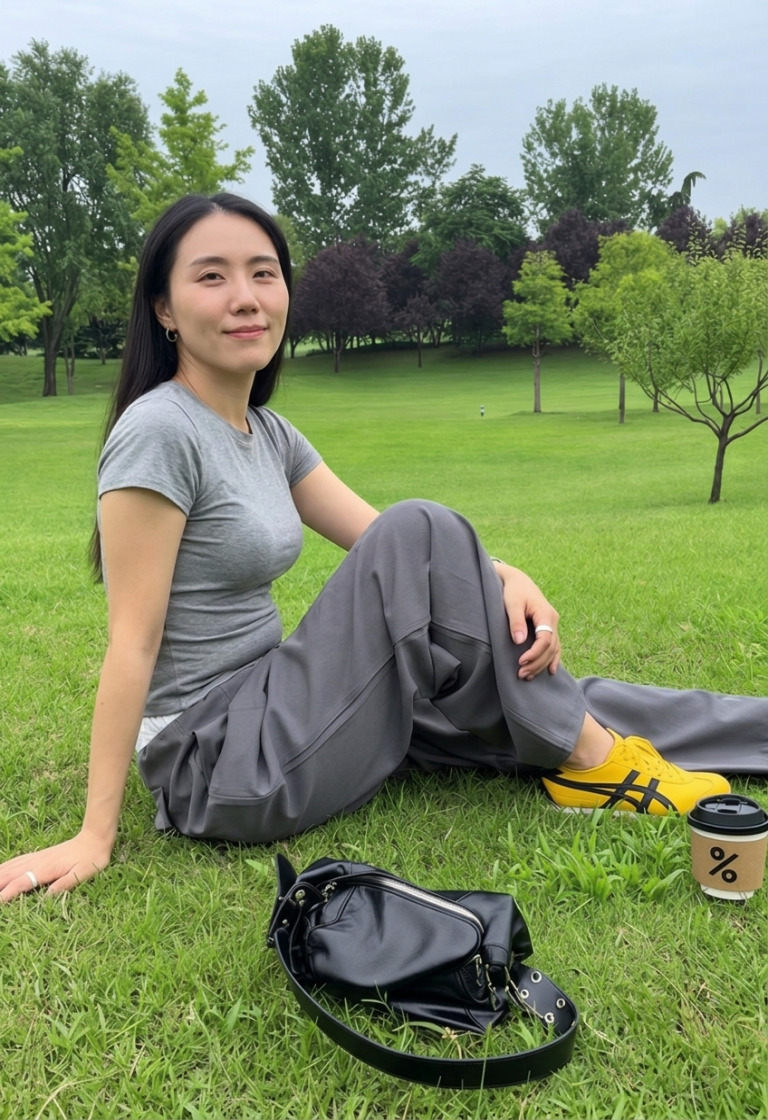} & \includegraphics[width=\qualwidth]{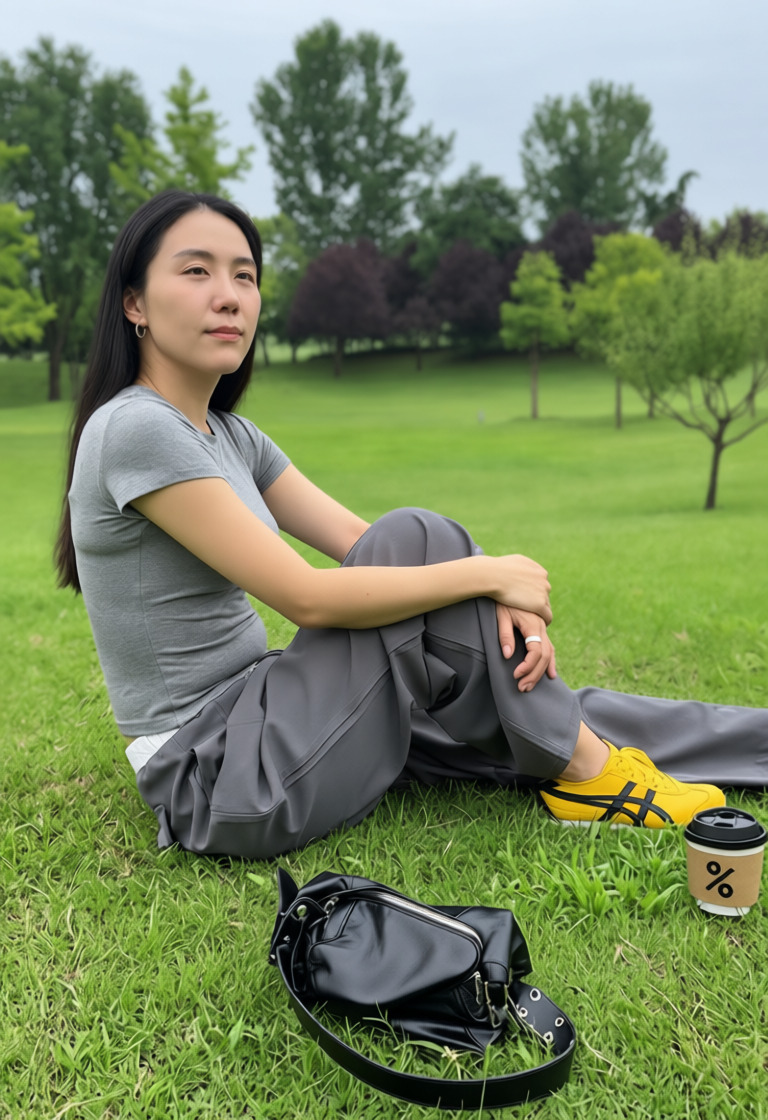} & \includegraphics[width=\qualwidth]{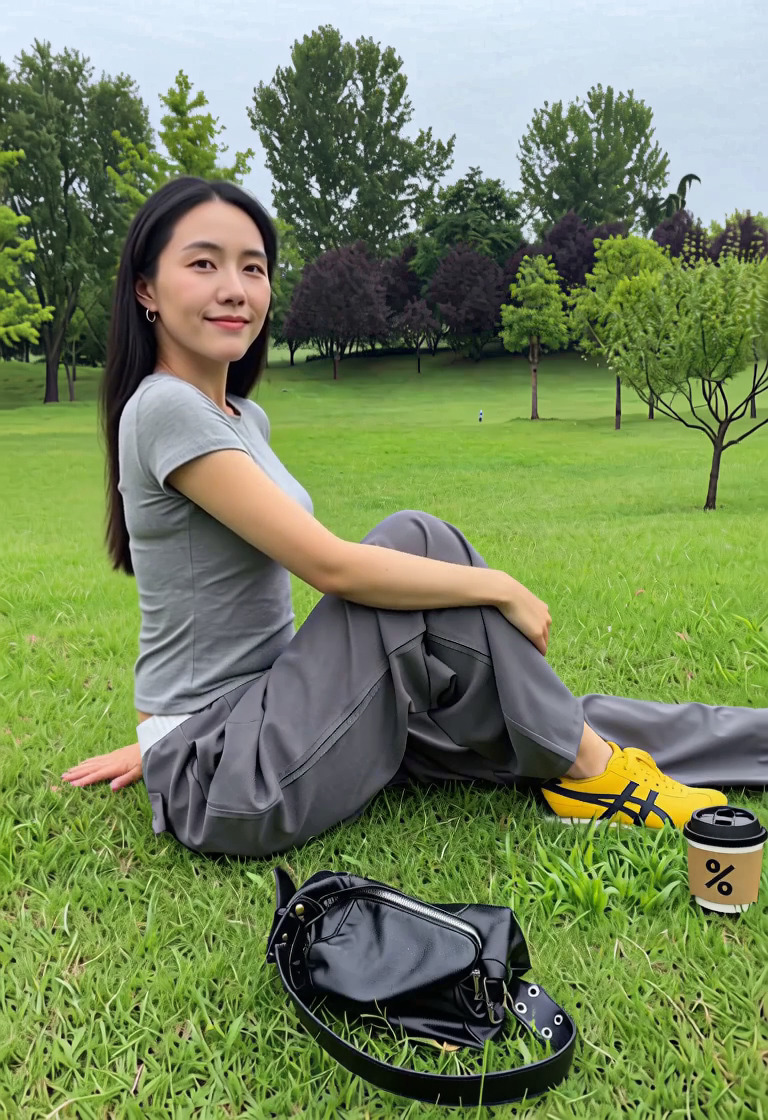} \\[0pt]
    \quallabeledimg{\includegraphics[width=\qualwidth]{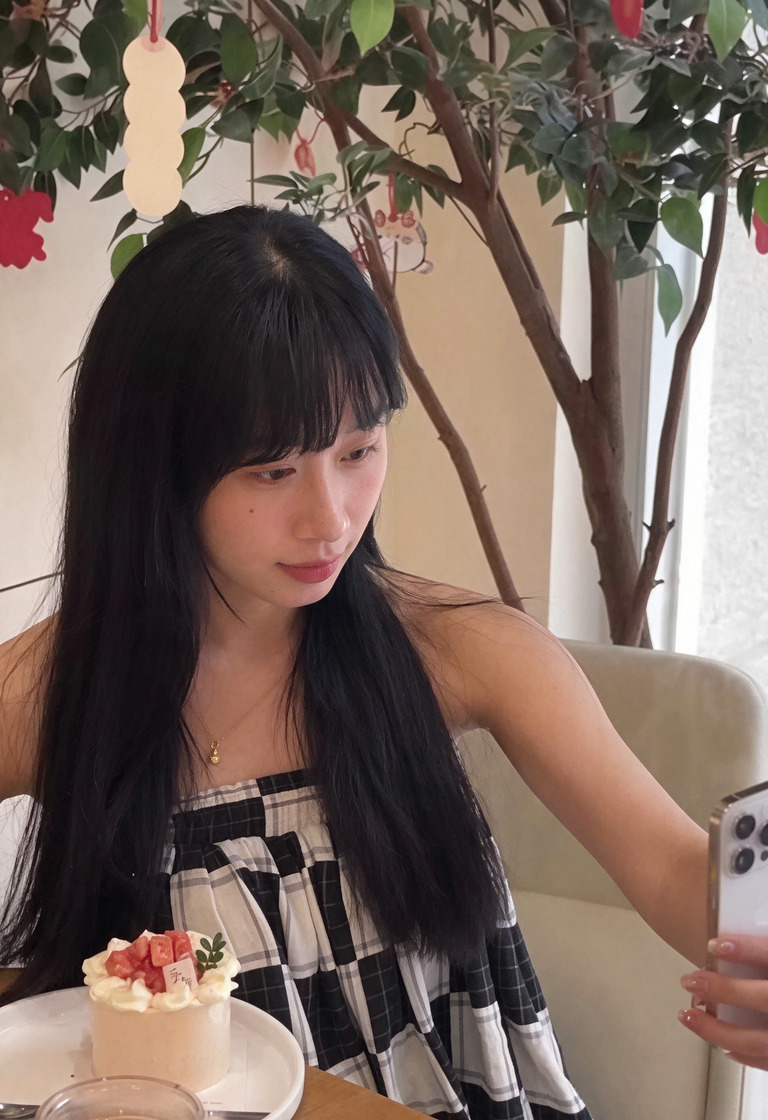}}{A2} & \includegraphics[width=\qualwidth]{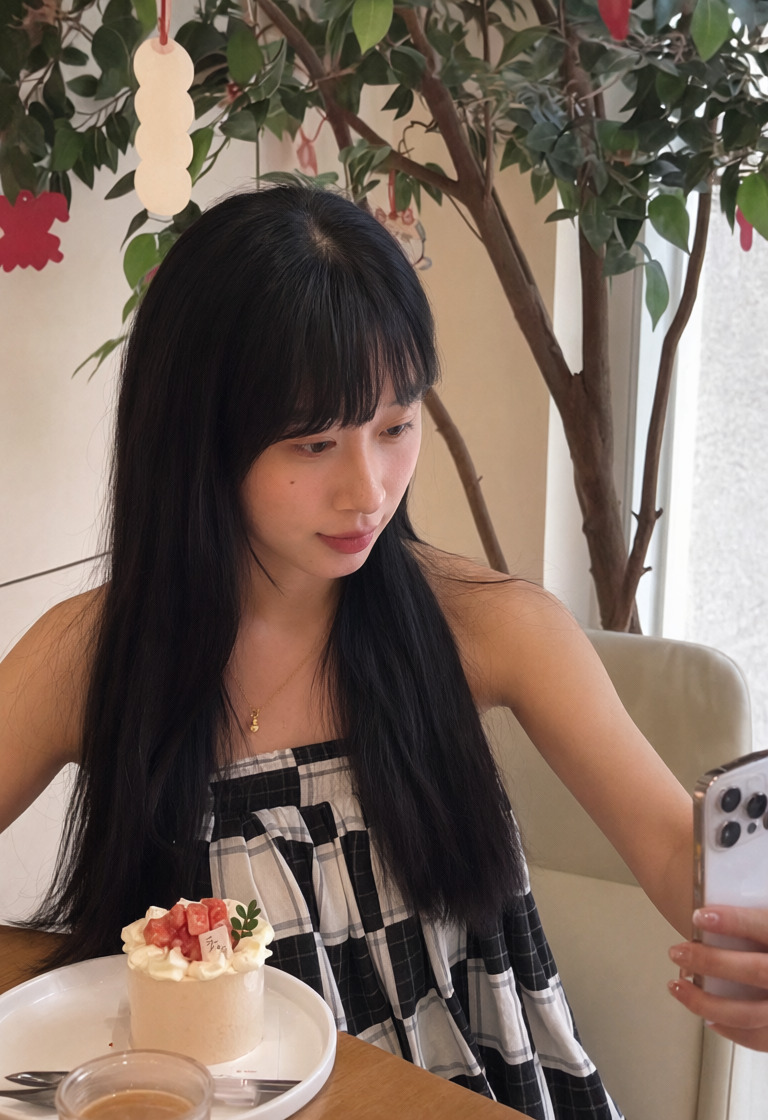} & \includegraphics[width=\qualwidth]{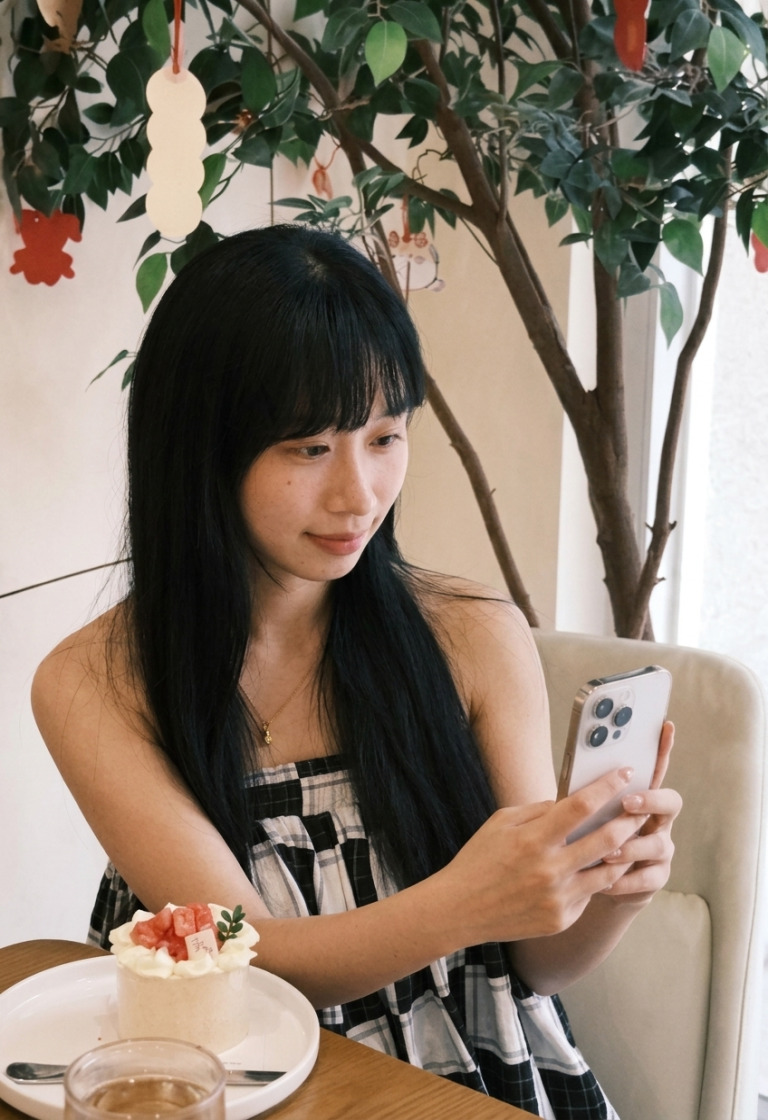} & \includegraphics[width=\qualwidth]{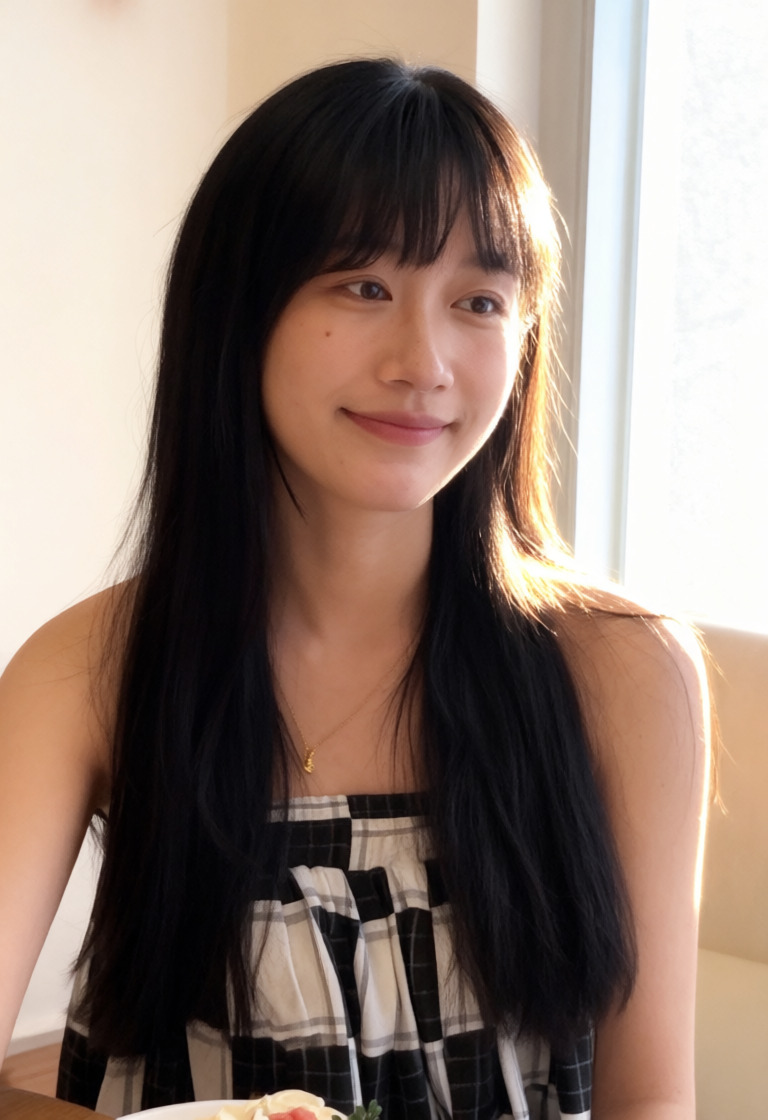} & \includegraphics[width=\qualwidth]{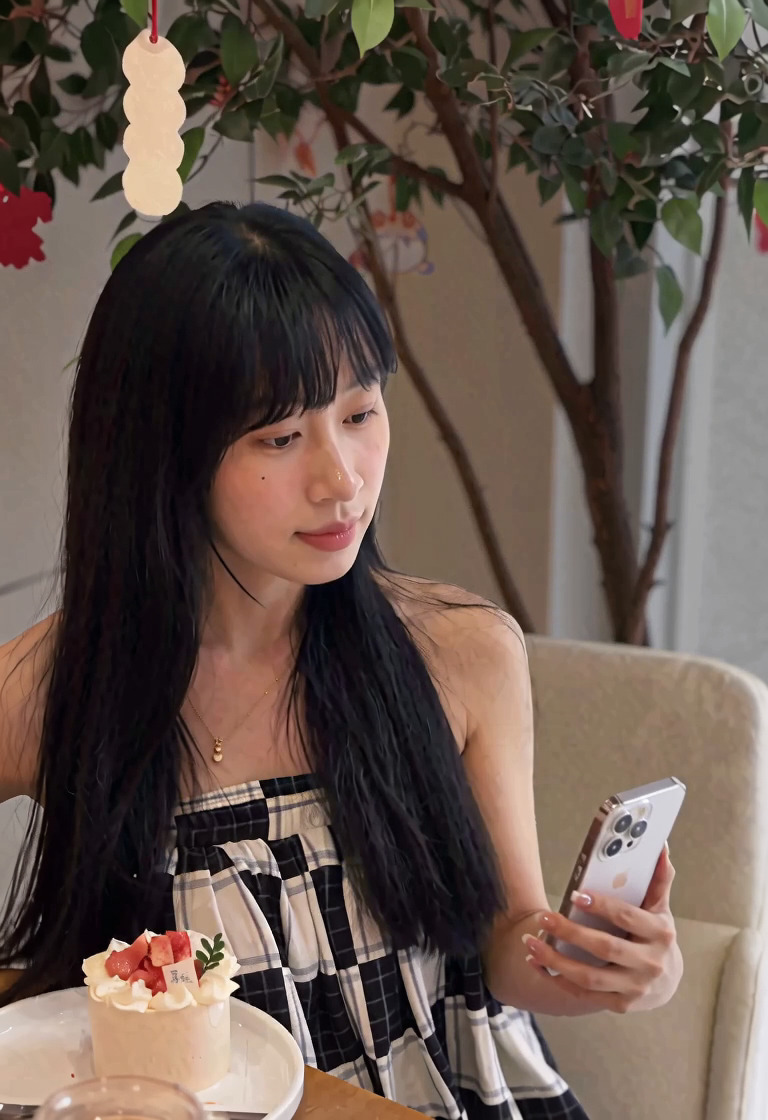} & \quallabeledimg{\includegraphics[width=\qualwidth]{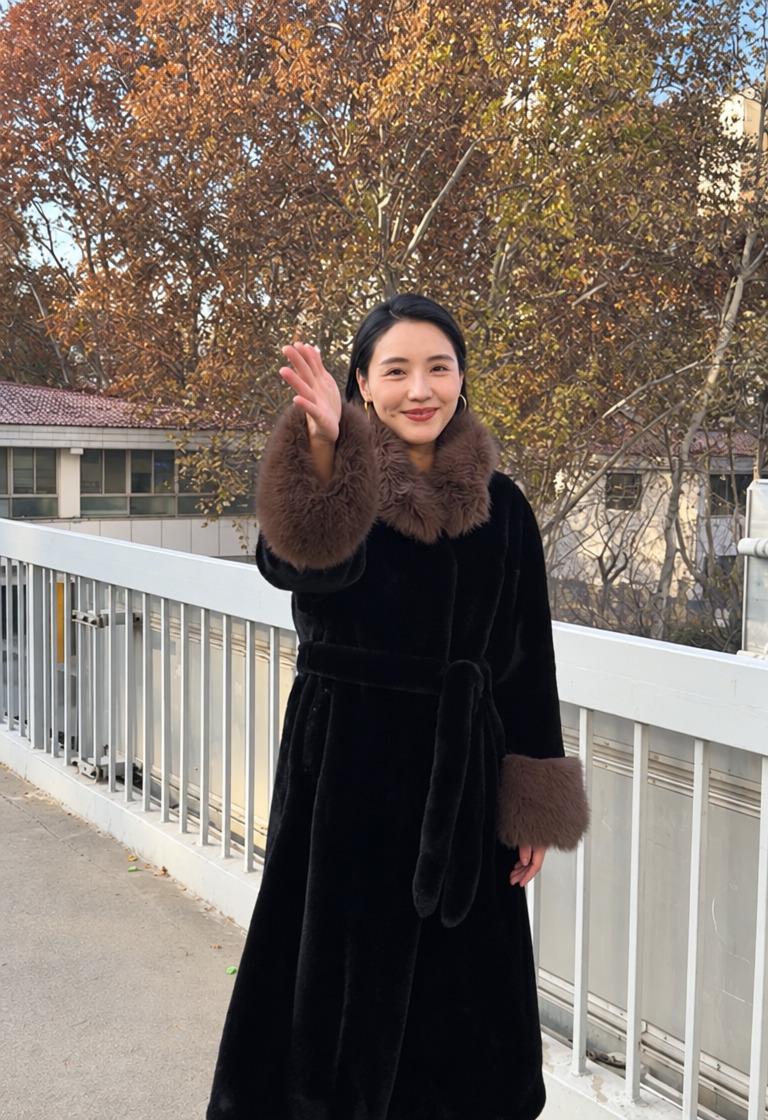}}{B2} & \includegraphics[width=\qualwidth]{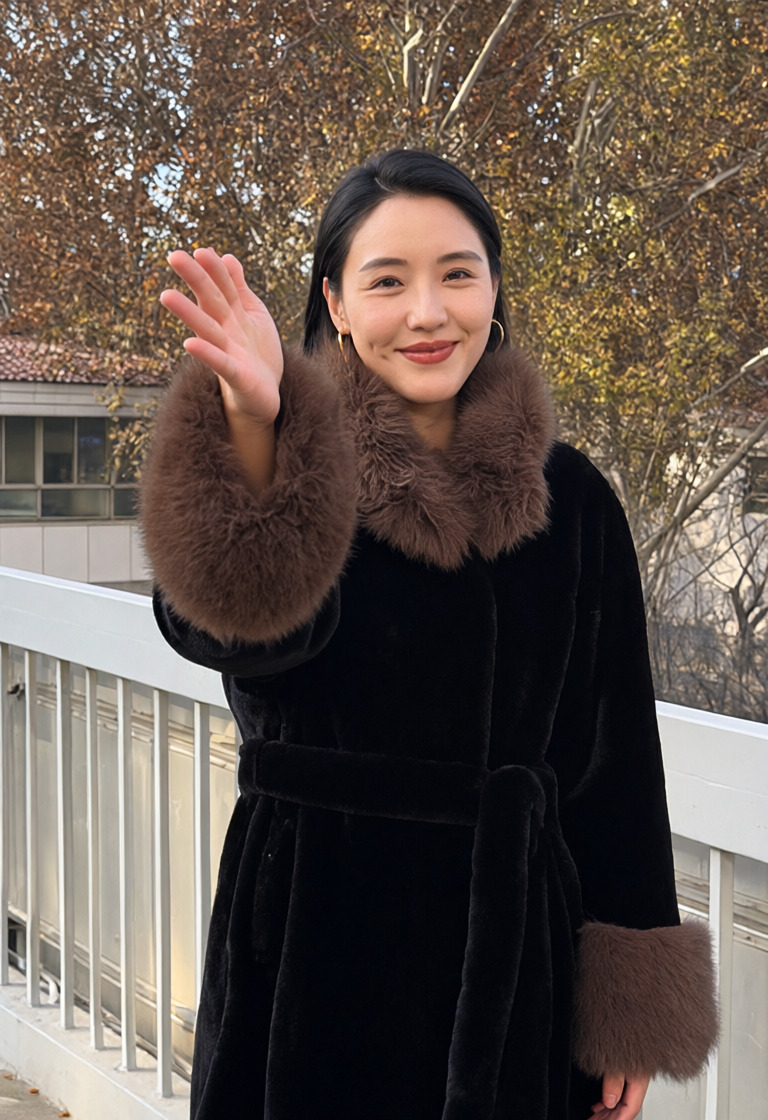} & \includegraphics[width=\qualwidth]{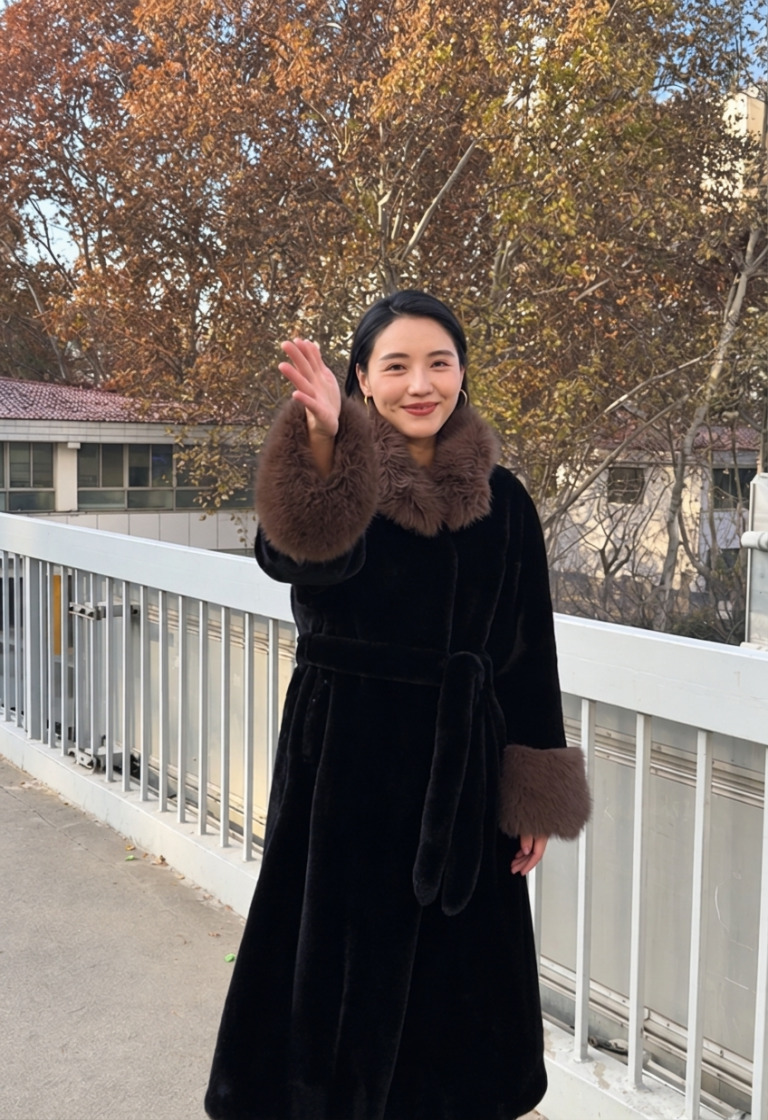} & \includegraphics[width=\qualwidth]{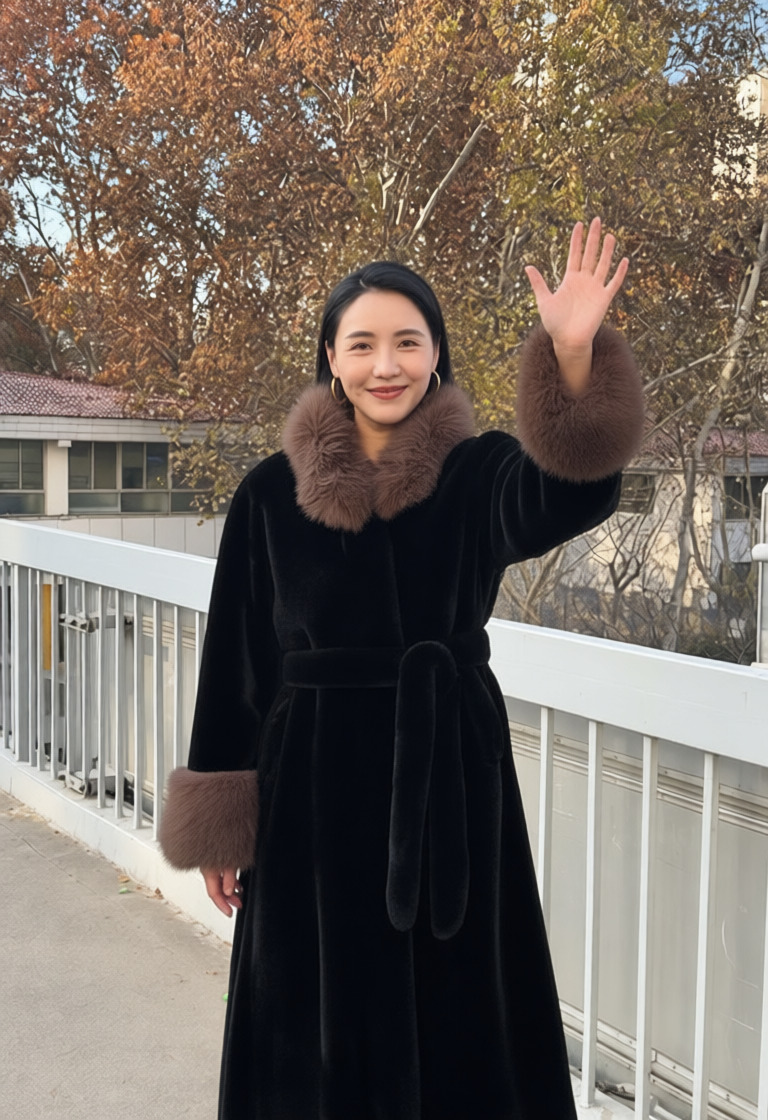} & \includegraphics[width=\qualwidth]{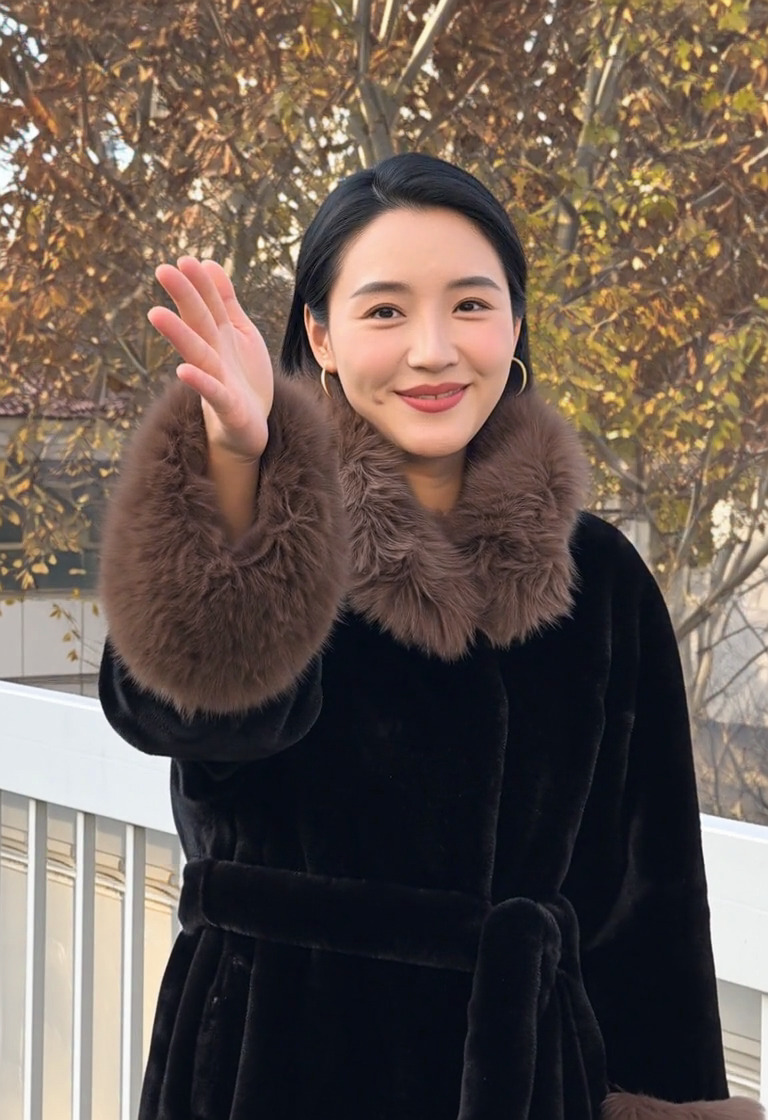} \\[0pt]
    \quallabeledimg{\includegraphics[width=\qualwidth]{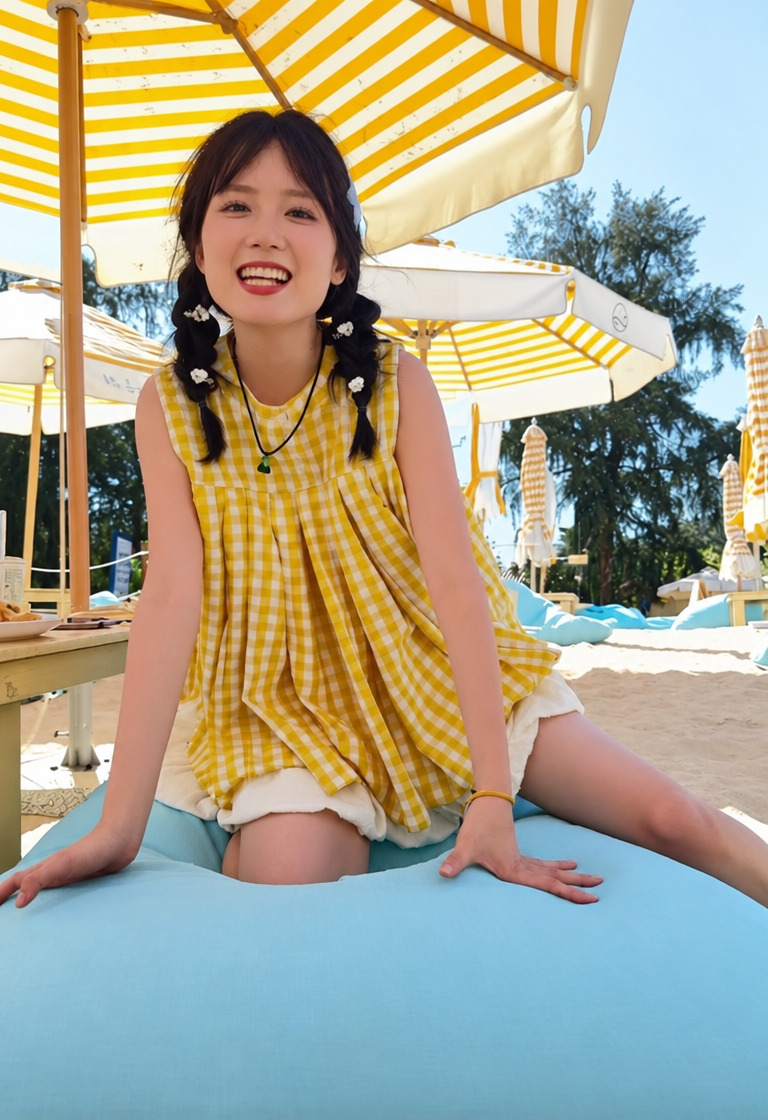}}{A3} & \includegraphics[width=\qualwidth]{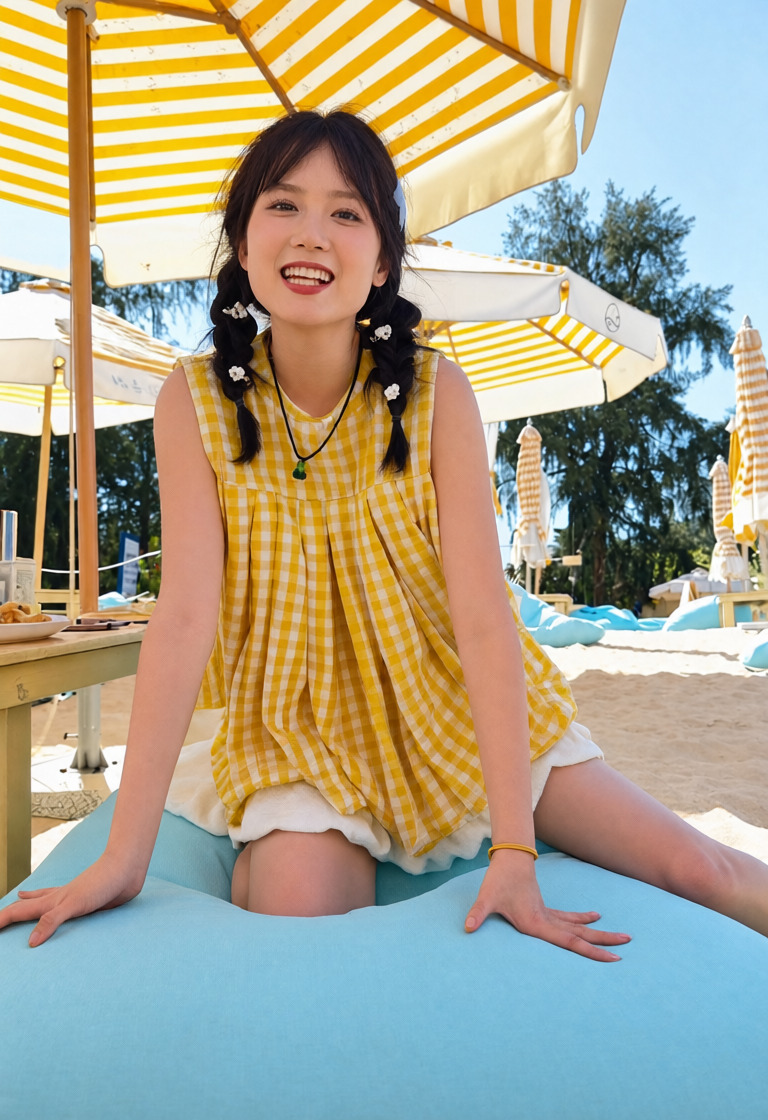} & \includegraphics[width=\qualwidth]{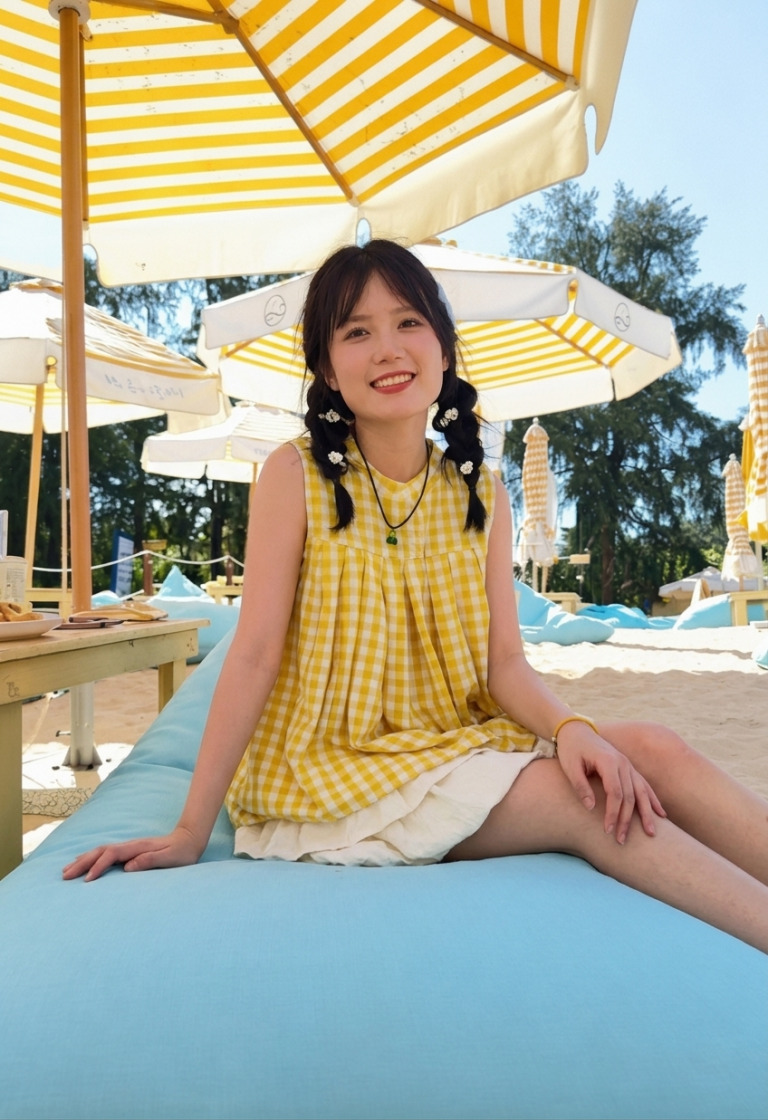} & \includegraphics[width=\qualwidth]{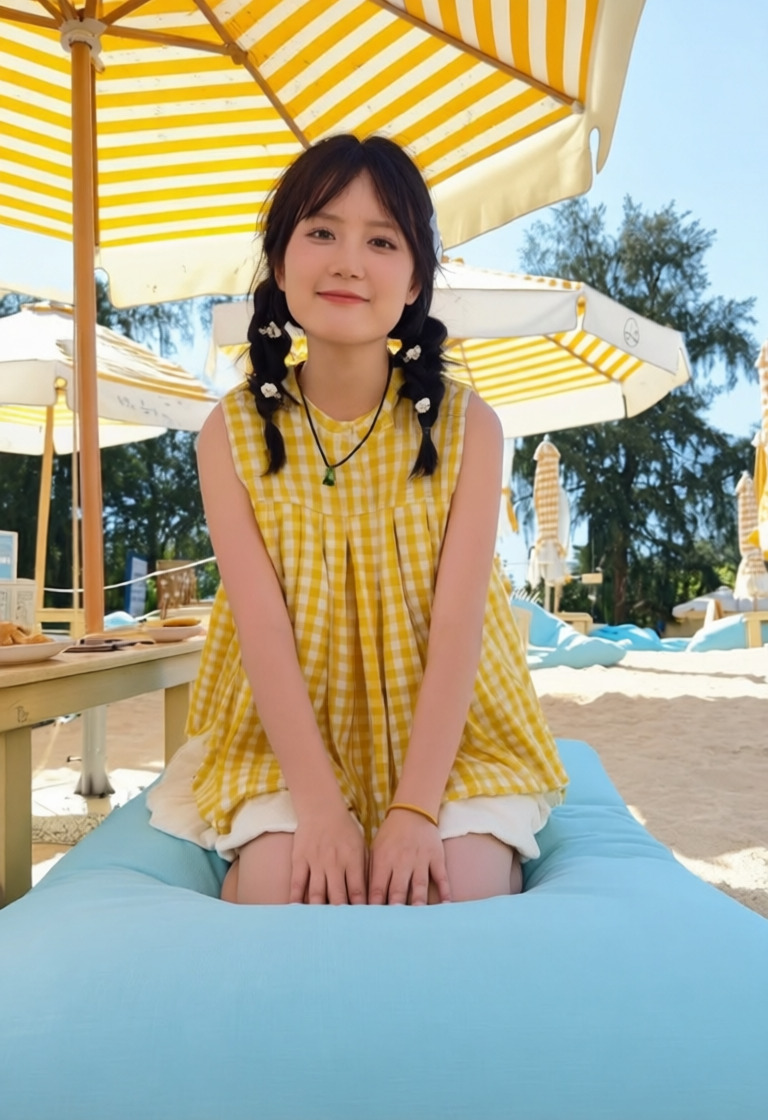} & \includegraphics[width=\qualwidth]{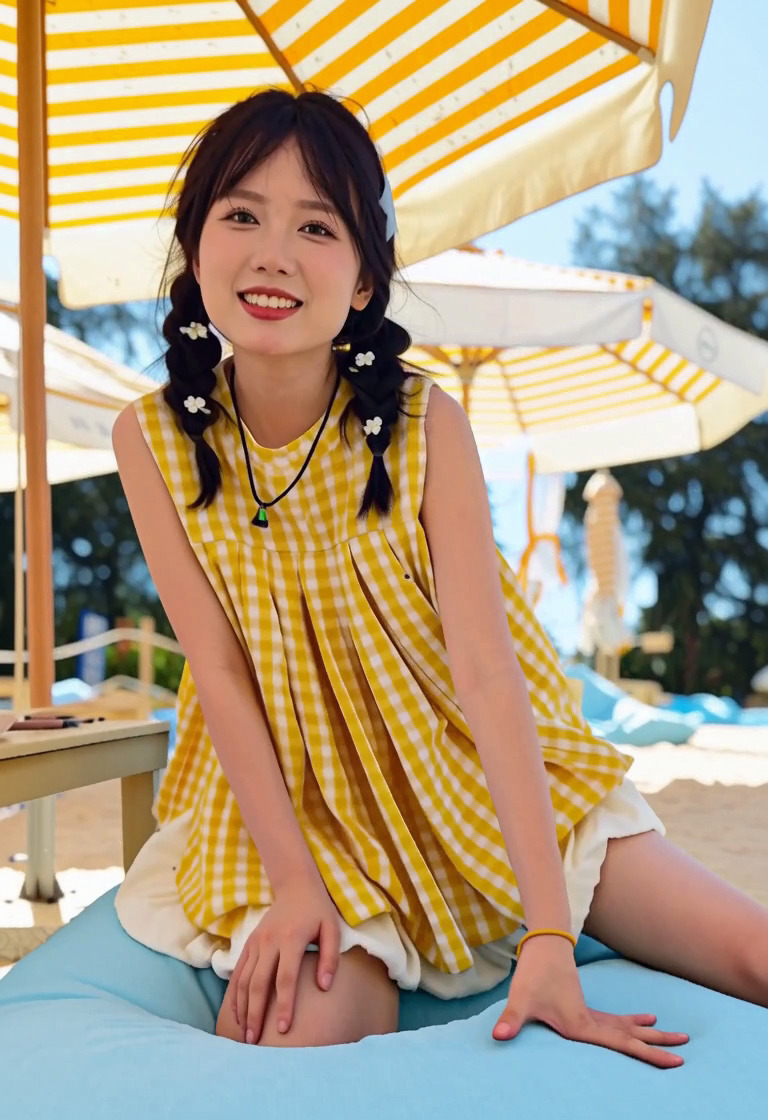} & \quallabeledimg{\includegraphics[width=\qualwidth]{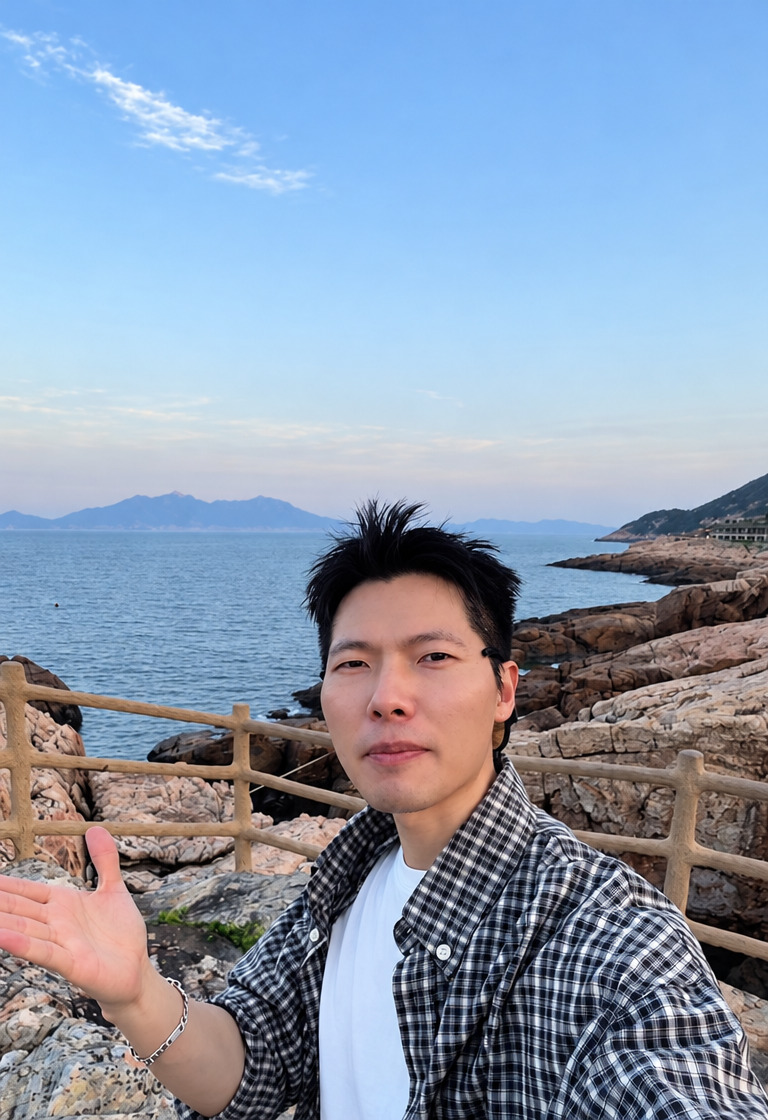}}{B3} & \includegraphics[width=\qualwidth]{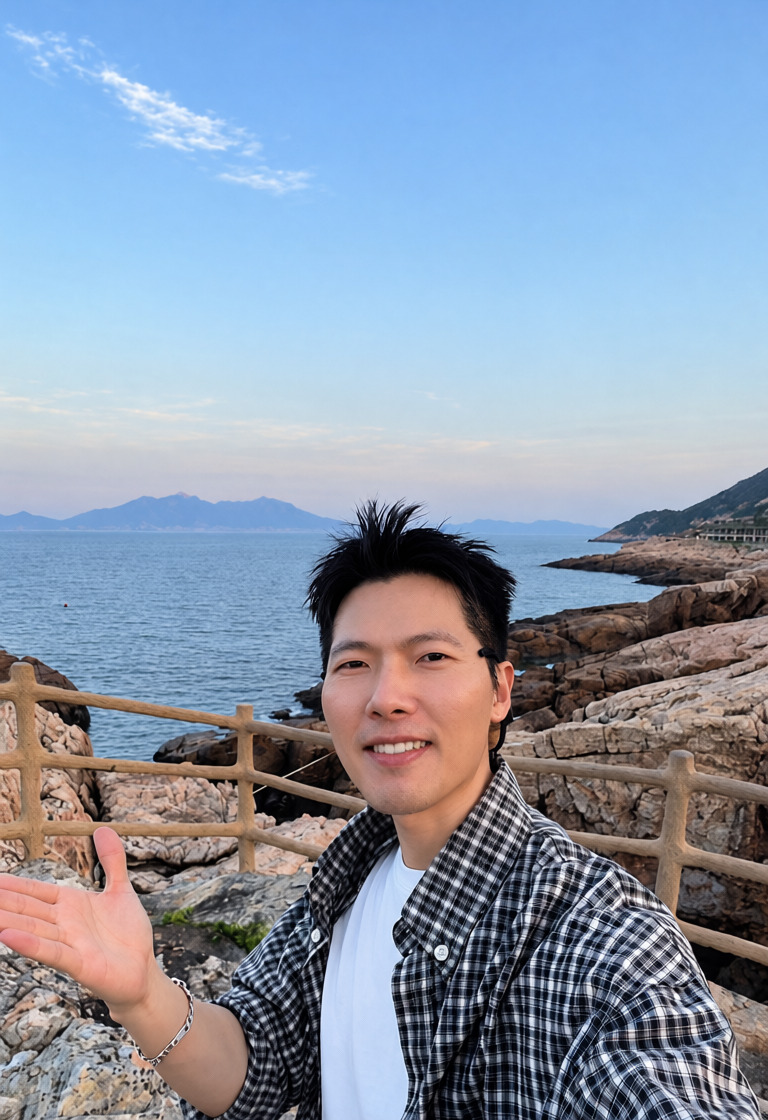} & \includegraphics[width=\qualwidth]{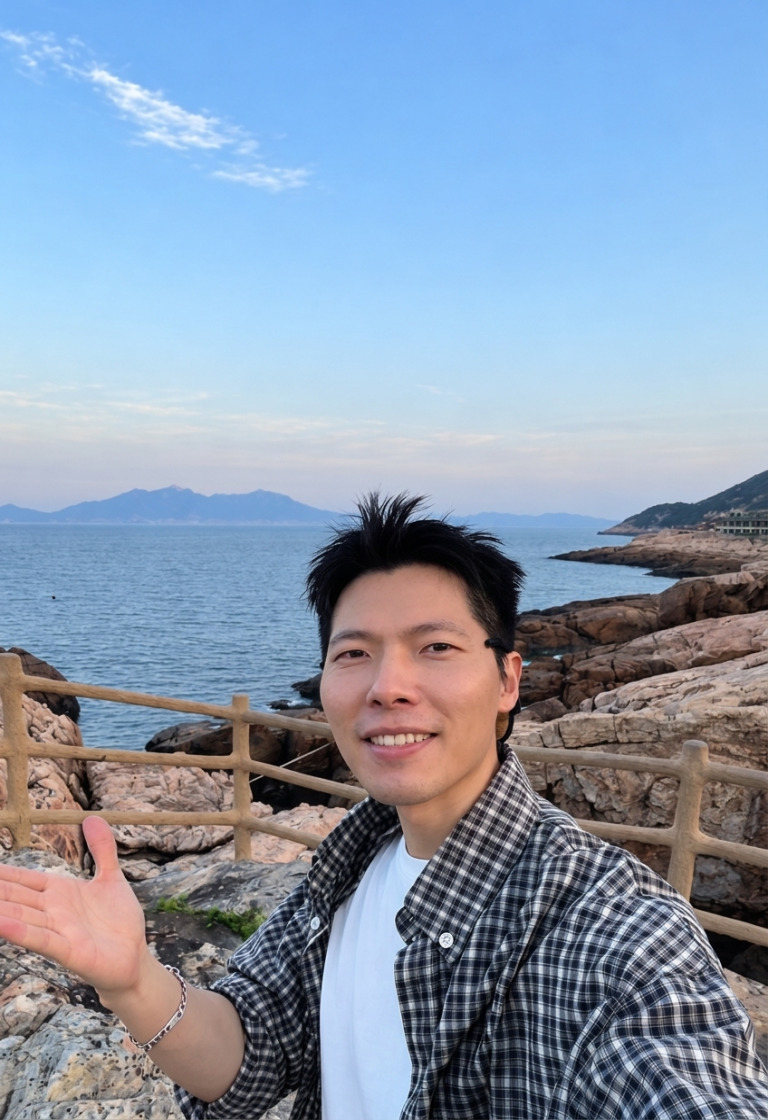} & \includegraphics[width=\qualwidth]{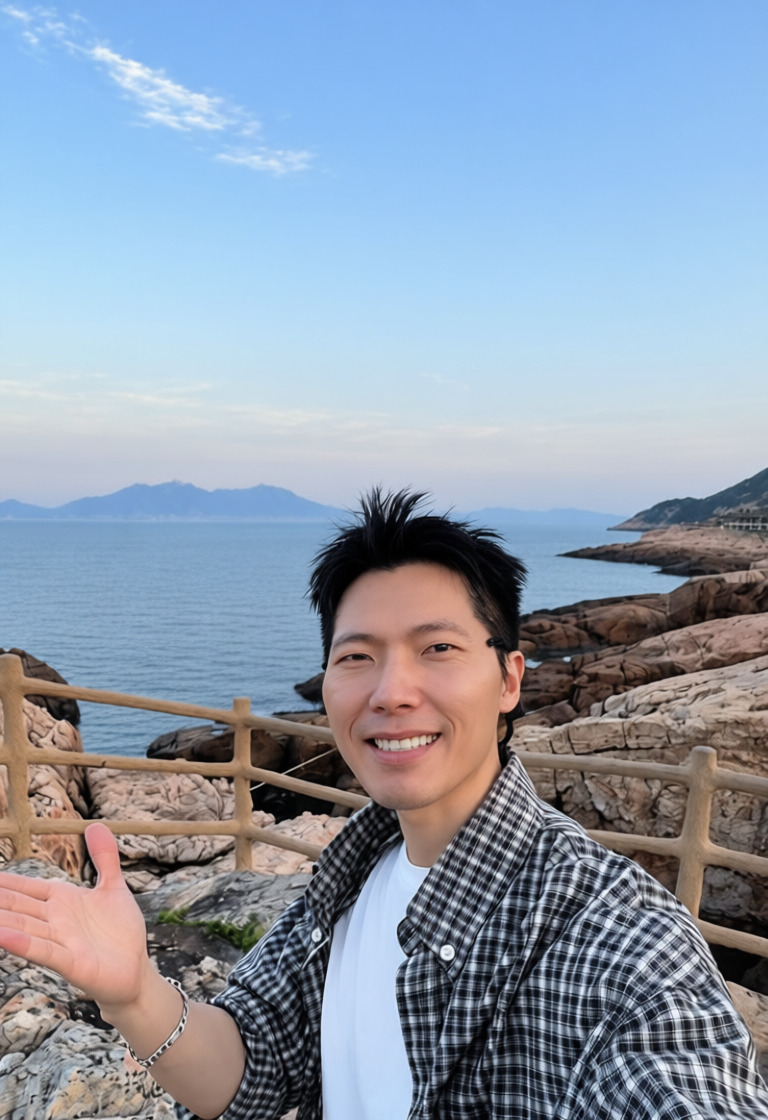} & \includegraphics[width=\qualwidth]{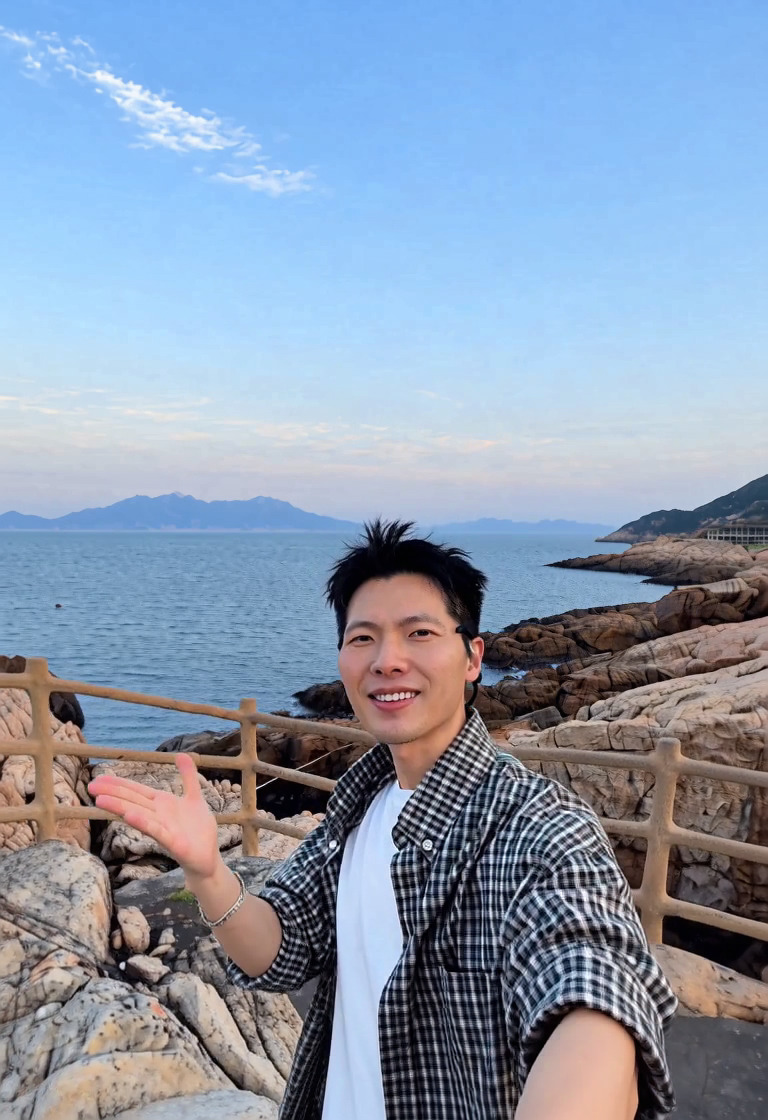} \\[0pt]
    \quallabeledimg{\includegraphics[width=\qualwidth]{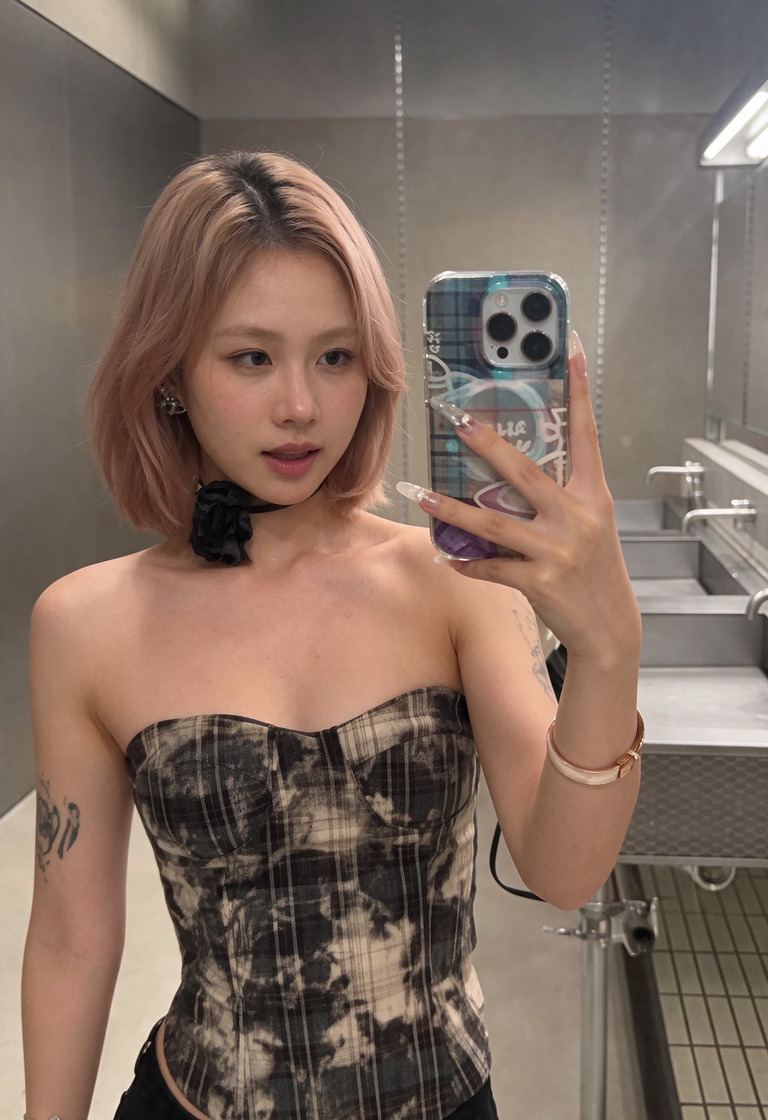}}{A4} & \includegraphics[width=\qualwidth]{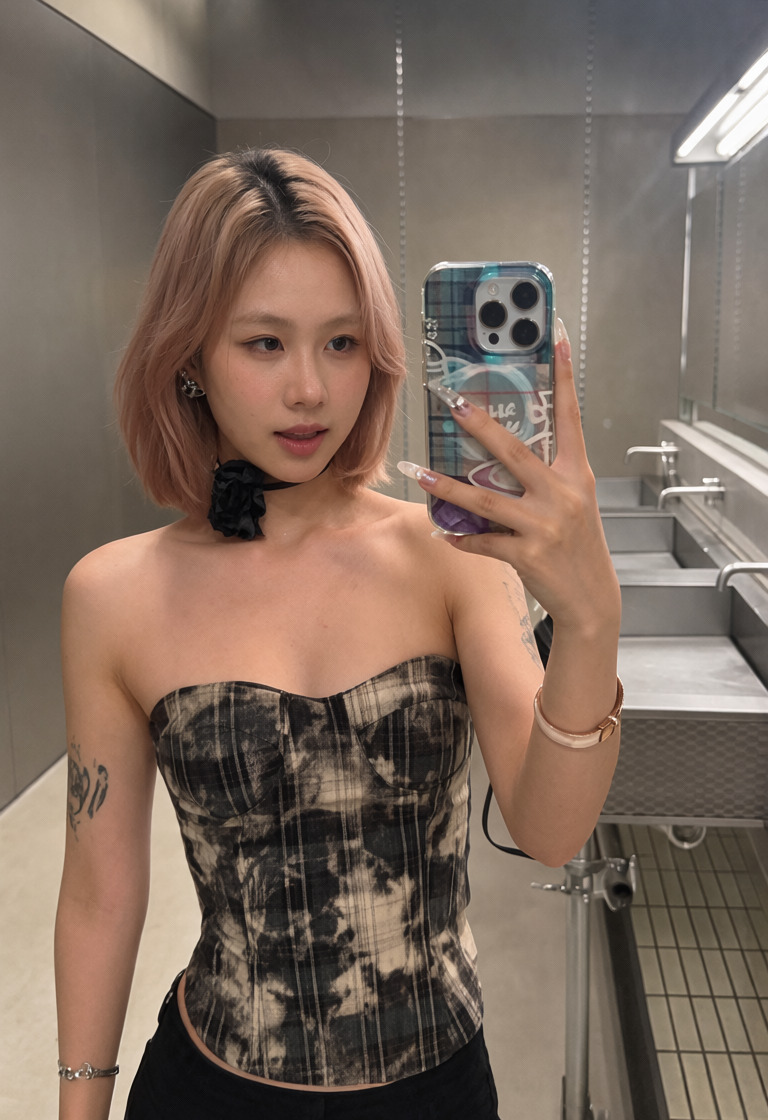} & \includegraphics[width=\qualwidth]{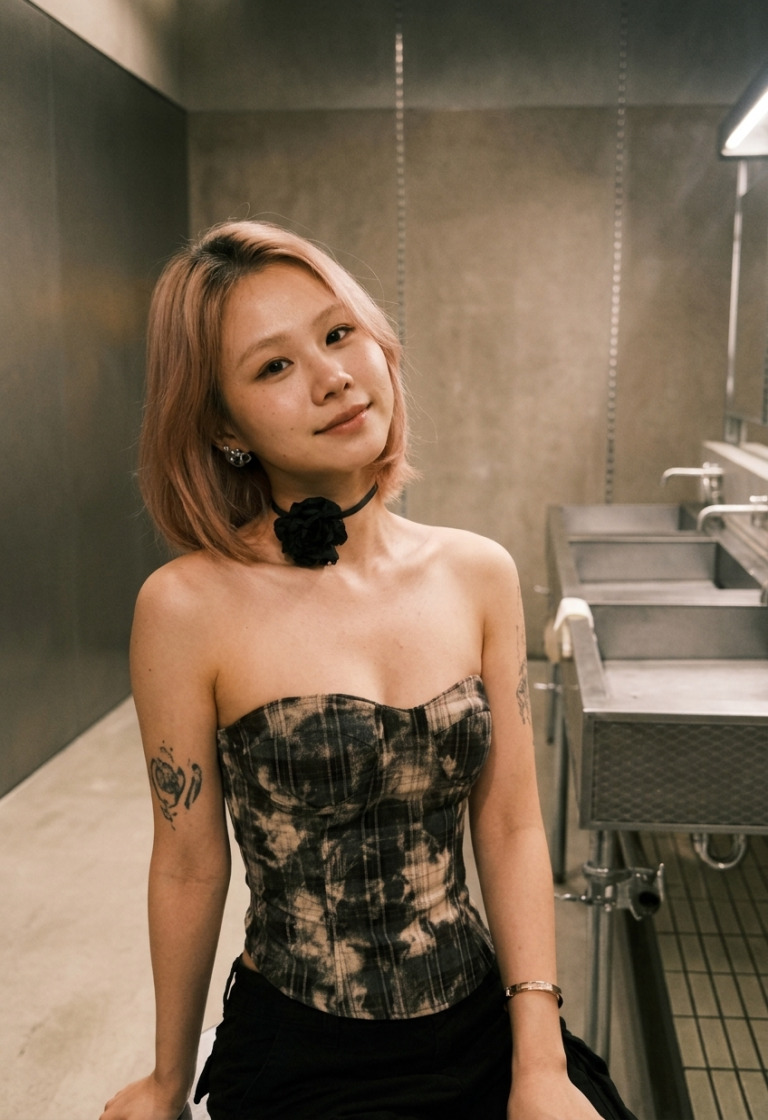} & \includegraphics[width=\qualwidth]{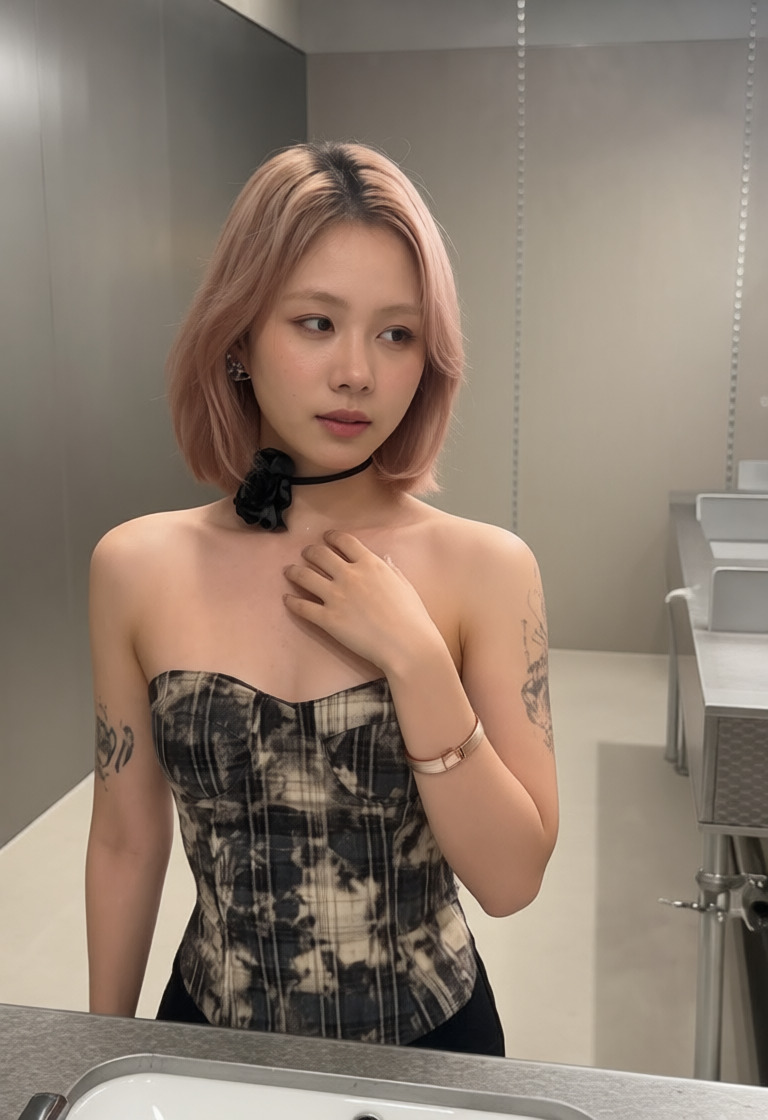} & \includegraphics[width=\qualwidth]{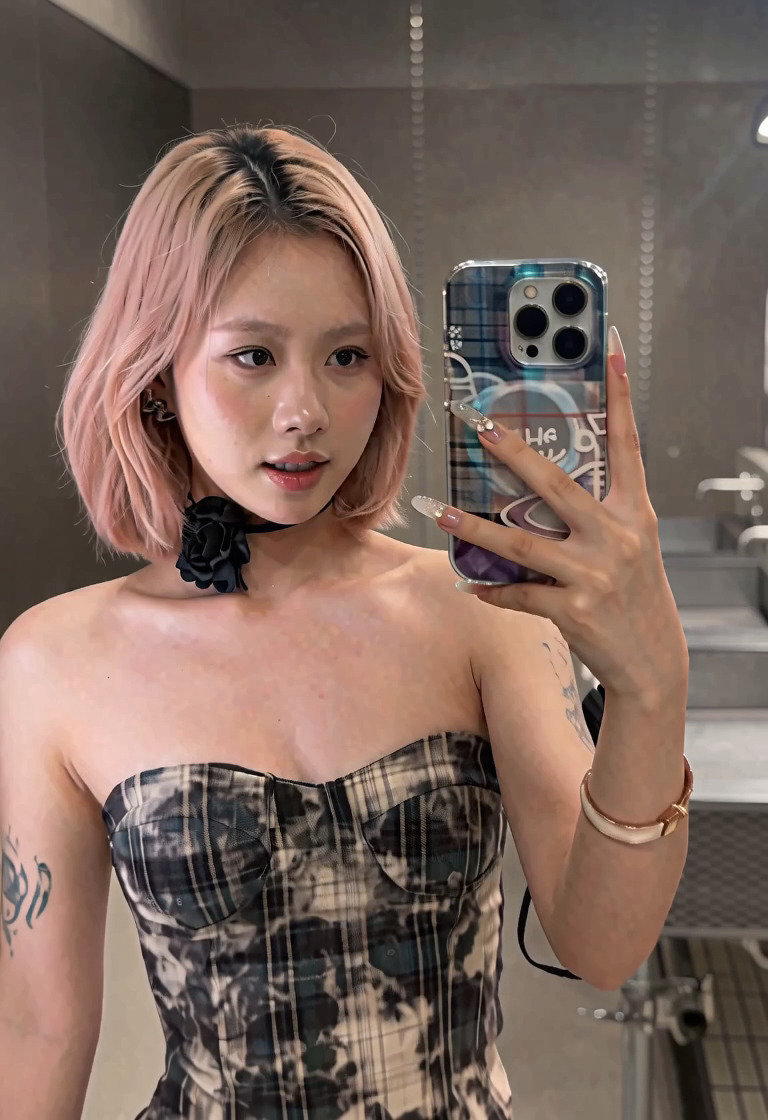} & \quallabeledimg{\includegraphics[width=\qualwidth]{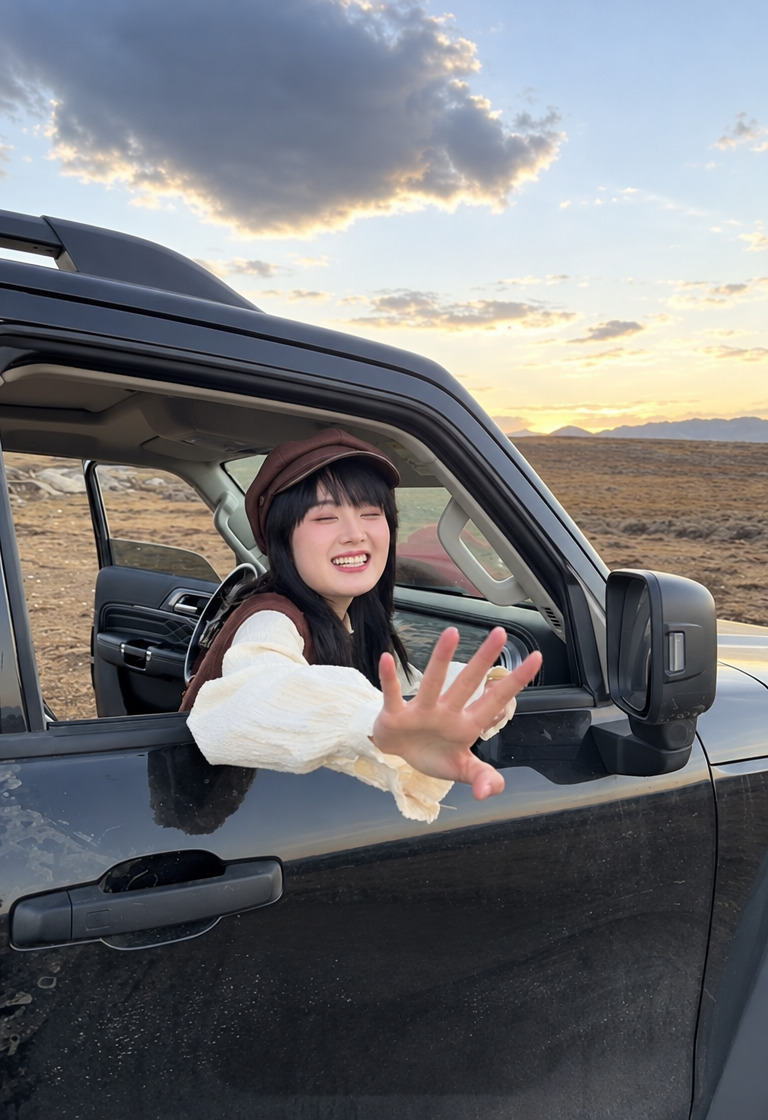}}{B4} & \includegraphics[width=\qualwidth]{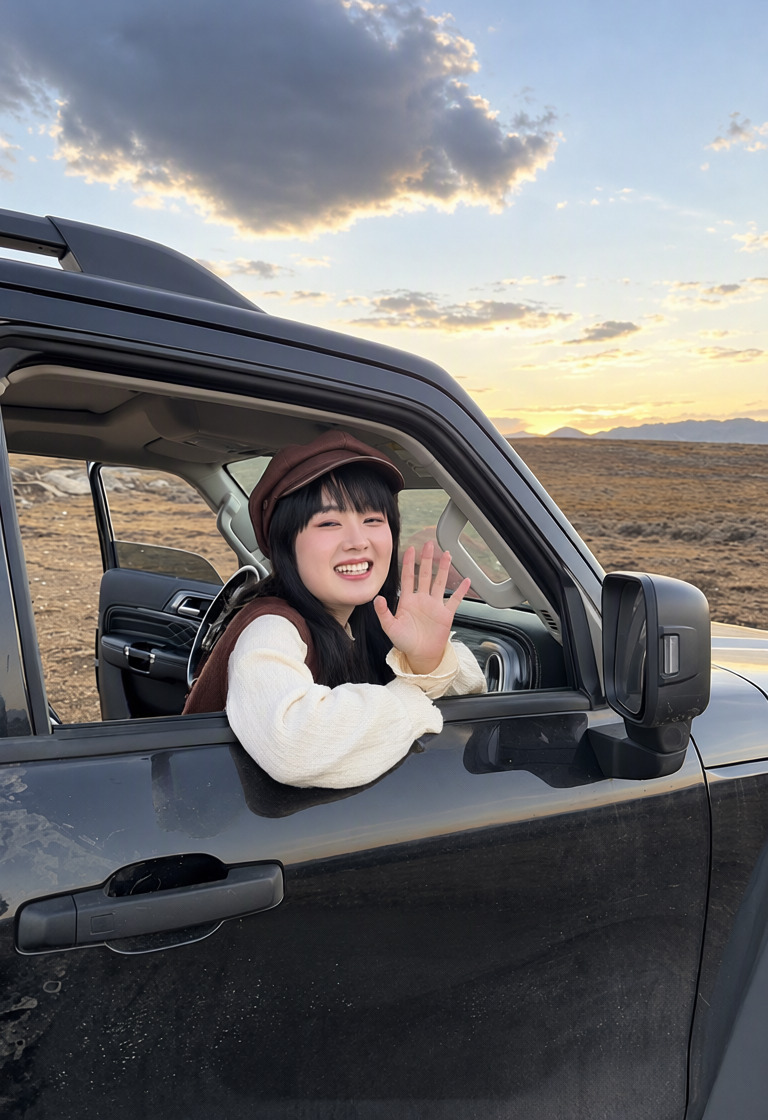} & \includegraphics[width=\qualwidth]{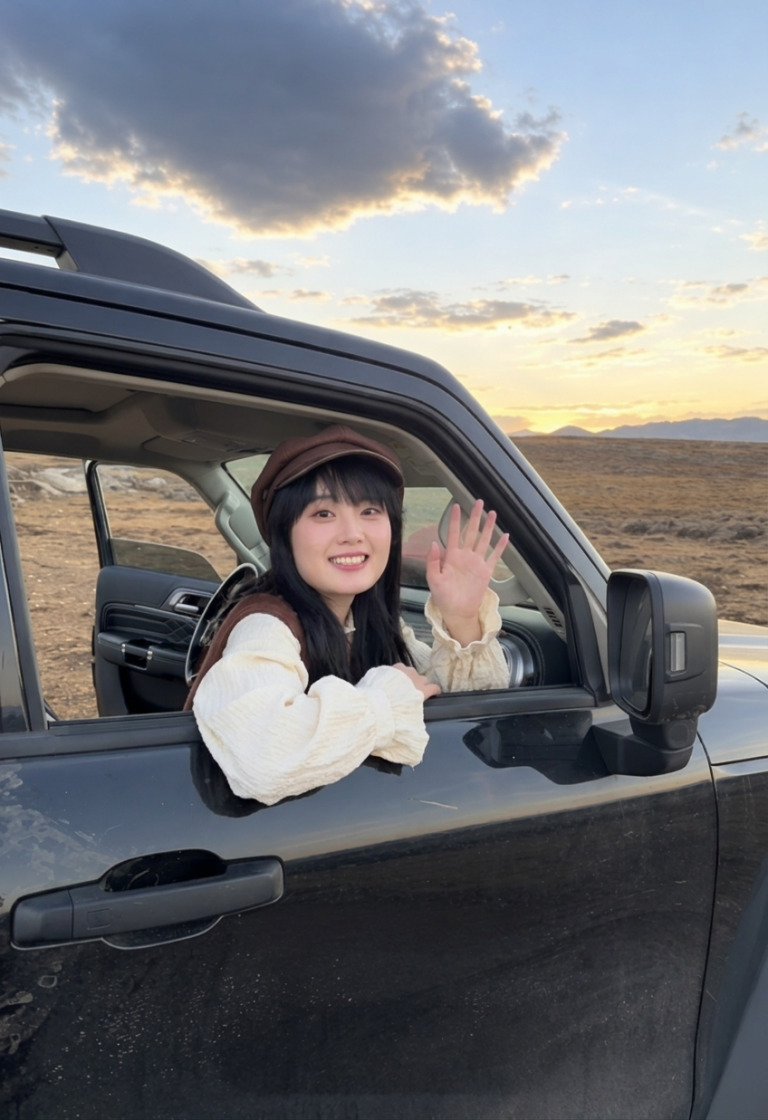} & \includegraphics[width=\qualwidth]{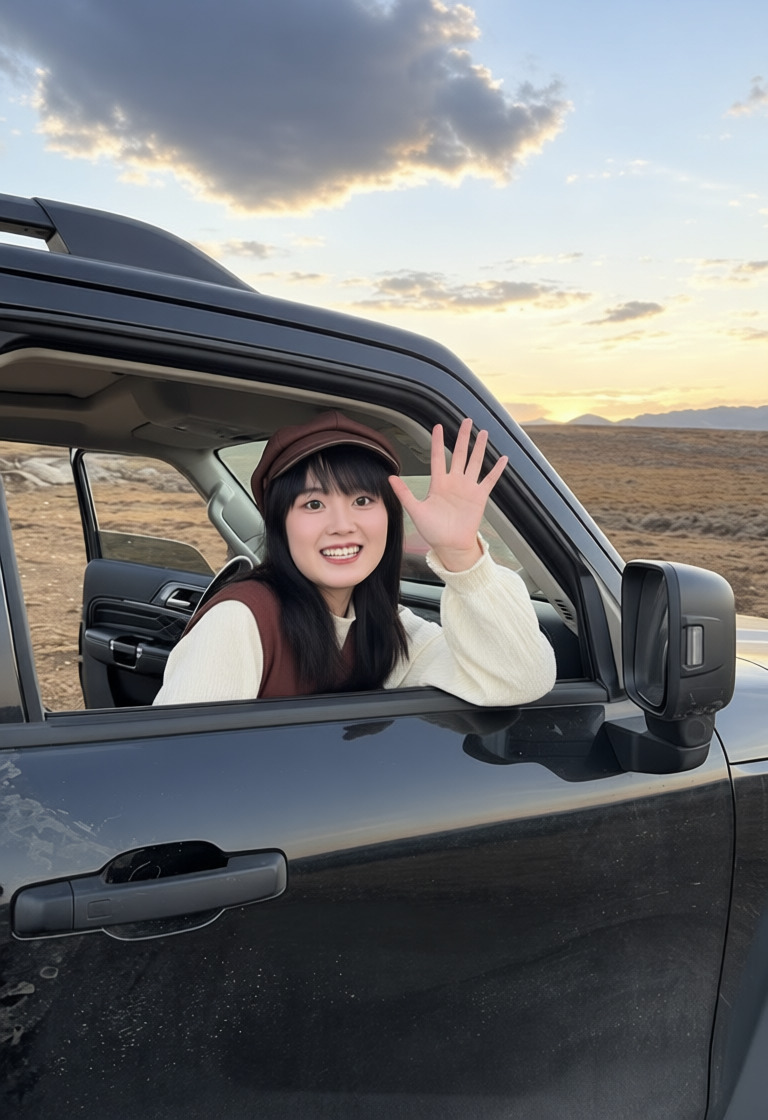} & \includegraphics[width=\qualwidth]{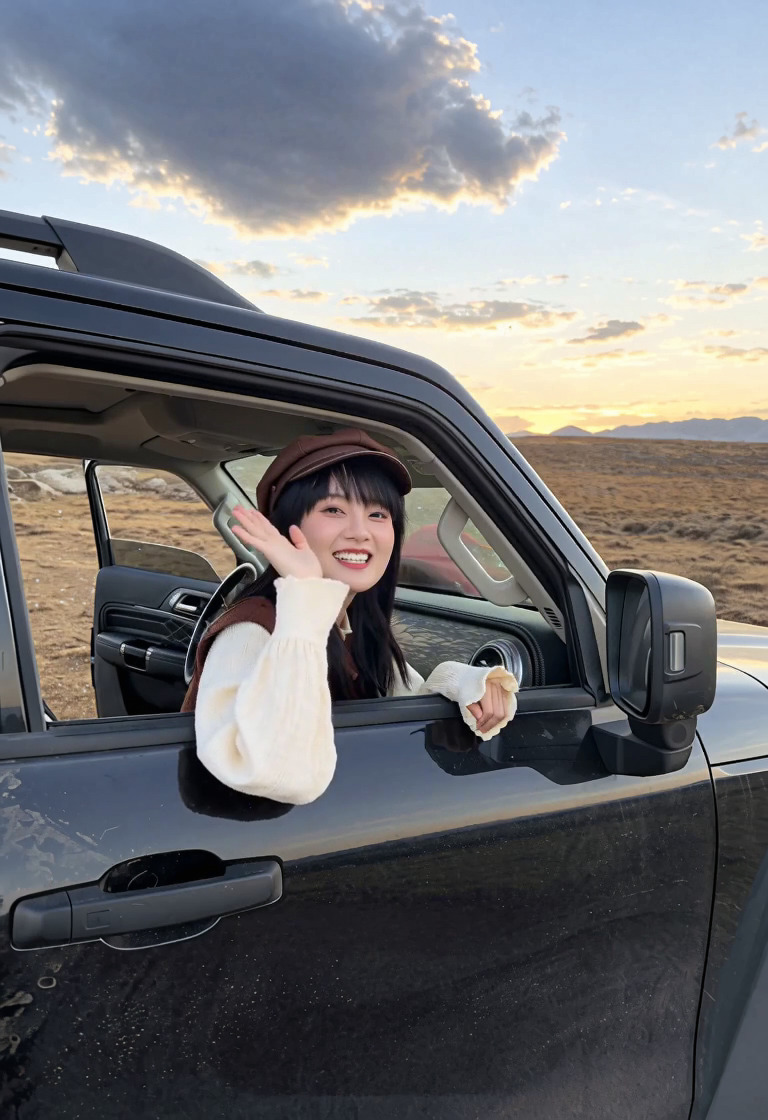}
  \end{tabular}}
  \caption{\textbf{Qualitative reshooting comparison.}
    A1--A4 share the open-ended reshooting prompt; targeted instructions for B1--B4 are summarized below.
    \enspace\textbf{B1:} Sit upright, keeping one leg bent and the other extended; brace with the left arm beside the left hip; look at the camera and smile slightly.
    \enspace\textbf{B2:} Zoom in to waist-up framing and keep the raised right-hand pose.
    \enspace\textbf{B3:} Pull back to include the right hand, shoulders, and far bank; keep direct gaze, show upper teeth, and relax the brow and eyes.
    \enspace\textbf{B4:} Draw the outstretched hand back to wave beside the shoulder, keeping the elbow on the car door's top edge; open the eyes and look at the camera with the same toothy smile.}
  \label{fig:targeted-qualitative}
\end{figure*}

\subsection{Qualitative Results}
\label{sec:qualitative}
Figure~\ref{fig:targeted-qualitative} compares SimShoot against three representative baselines across both tracks.
In open-ended reshooting (A1--A4), SimShoot improves posture, gaze, and framing while preserving scene objects and original photographic style: in A1 it draws the cut-off outstretched right arm inward to hold the yellow flowers cleanly within the frame with direct eye contact; in A2 it brings the cut-off phone cleanly into frame while keeping the decorated tree (which AesFormer omits); in A3 it levels the upward chin tilt into direct eye contact; and in A4 it tightens framing into a balanced chest-up close-up without applying Nano Banana~2.1's heavy warm color grade or losing the phone as in AesFormer.

In targeted reshooting (B1--B4), SimShoot resolves spatial and anatomical constraints while keeping unrelated scene attributes stable (two additional targeted trajectories in Appendix~\ref{sec:supp-case-studies}): sitting upright with direct gaze and the requested left-arm grass brace in B1 (where Nano Banana~2.1 braces with the right arm); zooming to waist-up framing while keeping the raised right hand in B2 (where Qwen-Image~2.1 and Nano Banana~2.1 under-zoom and AesFormer raises the opposite hand); pulling back to include the outstretched right arm and shoulders in B3; and drawing the outstretched hand back to wave beside the shoulder in B4 (where all three wave the wrong hand).

\reshootClosedLoopAblationTable

\subsection{Ablation Study}
\label{sec:ablations}
Table~\ref{tab:closed-loop-ablations} ablates each core module across all $240$ portraits using PortraitAes-S (human-aligned structure, $\rho=+0.94$; Sec.~\ref{sec:human-evaluation}) and Q-Align (sharpness/IQA monitoring re-synthesis blur; all six metrics in Appendix~\ref{sec:supp-qual-ablations}, Table~\ref{tab:supp-ablation-full}).

\paragraph{Action-specific LoRAs vs.\ pretrained VDM.}
In the full loop (Table~\ref{tab:closed-loop-ablations}(a)), disabling our seven LoRAs on MiniMax H3 drops edit acceptance from $82.3\%$ to $59.4\%$, triples rejected rollouts ($436$ vs.\ $140$, $3.1\times$), and degrades optical and 3D camera control ($-62.1$\,pp acceptance on \texttt{cam\_dof}, $-43.1$\,pp on \texttt{cam\_translate}).
Even without LoRA, closed-loop selection with native H3 reaches $+12.34\%$ open-ended Structure, beating four baselines in Table~\ref{tab:main-results}, while action LoRAs improve all six open-ended and targeted metrics and fixed-capture background PSNR ($+3.07$\,dB).

\paragraph{Video simulator vs.\ extent-prompted image editors.}
Table~\ref{tab:closed-loop-ablations}(b) replaces our VDM simulator in the identical Planner--Selector loop with FLUX~3 Klein or Qwen-Image~2.1 at matching nominal extent fractions.
In blinded ranking across $42$ matched actions ($252$ pairs/executor), SimShoot achieves monotonic progression on $92.9\%$ ($39/42$) of actions with \emph{zero} reversals ($89.7\%$ strictly increasing pairs; $3/42$ subtle edits plateau by \texttt{F10}), whereas FLUX~3 Klein ($16.7\%$, $7/42$) and Qwen-Image~2.1 ($23.8\%$, $10/42$) suffer frequent reversals ($63$ and $53$) and early saturation.
Across all $240$ portraits, one 39-frame rollout supplies all extents ($3.29$ calls/portrait, $82.3\%$ acceptance), whereas image editors need $18.88$--$27.14$ calls/portrait ($66.1\%$ and $35.6\%$ acceptance) and compound re-synthesis drift: SimShoot leads five of six metrics on both tracks (Table~\ref{tab:supp-ablation-full}), with only ArtiMuse favoring re-editing sharpening artifacts (Sec.~\ref{sec:human-evaluation}).

\begin{figure}[t]
  \centering
  \captionsetup{skip=3pt}
  \includegraphics[width=0.99\columnwidth]{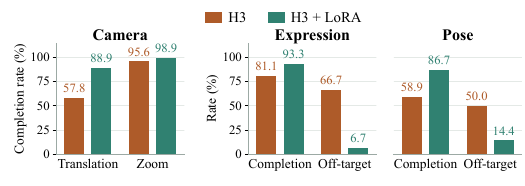}
  \caption{\textbf{Effect of LoRA on single-action terminal frames.} Three-judge completion and off-target rates ($30$ pairs/family, symmetrically imputing $1$ Gemini-blocked zoom/pose pair as \texttt{pass}/\texttt{present}; the $29$ complete pairs match: $95.4\%\to 98.9\%$ zoom, $57.5\%\to 86.2\%$ pose, $48.3\%\to 11.5\%$ pose off-target).}
  \label{fig:h3-action-control}
\end{figure}

\paragraph{Adaptive closed-loop replanning vs.\ fixed sequence.}
Executing a fixed list open-loop from the shared Round-1 rollout (Table~\ref{tab:closed-loop-ablations}(c)) forfeits dynamic revision: adaptive replanning alters accepted sequences on $77.1\%$ ($185/240$) of runs ($46.3\%$ [$111/240$] canonical variants) and action families on $45.0\%$ of open-ended runs.
Without replanning, $30.0\%$ ($36/120$) of open-ended fixed runs exhaust their list ($25$ after rejections; $303$ vs.\ $348$ open-ended edits, $594$ vs.\ $650$ overall), whereas adaptive replanning recovers or appends refinements ($0.0\%$ exhaustion) and leads all six open-ended and five of six targeted metrics (Table~\ref{tab:supp-ablation-full}).

\paragraph{Coarse-to-fine extent selection vs.\ endpoint-only.}
Across all $650$ accepted benchmark edits, $82.8\%$ pick an interior frame (\texttt{F1}--\texttt{F37}) rather than \texttt{F38}.
Replaying $240$ shared 39-frame rollouts across three menus (Table~\ref{tab:closed-loop-ablations}(d)), coarse 5-frame inspection drives the main gain over \texttt{\{F0, F38\}}, changing $71.7\%$ ($172/240$) of choices, lifting acceptance ($54.2\%\to 80.8\%$ open-ended, $80.0\%\to 93.3\%$ targeted), and significantly improving Structure and Q-Align, while zero-video-cost refinement calibrates subtle boundary cases ($67.4\%$ [$118/175$] fine picks online, $61.9\%$ [$26/42$] here, cutting \texttt{F38} picks from $18.8\%$ to $14.2\%$ with neutral mean shifts vs.\ coarse-5; Table~\ref{tab:supp-ablation-full}).

\begin{figure}[t]
  \centering
  \captionsetup{skip=3pt}
  \includegraphics[width=0.96\columnwidth]{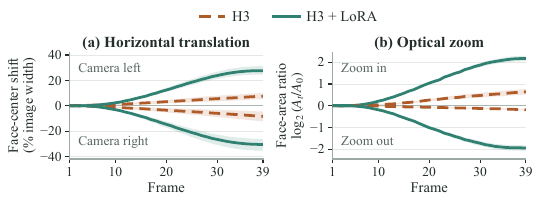}
  \caption{\textbf{Directional camera trajectories across $39$ frames.} Mean signed face-center shift (\% of image width) for lateral camera translation and $\log_2$ mean face-area ratio for focal-length zoom ($30$ matched pairs each; shaded bands denote $90\%$ paired bootstrap confidence intervals).}
  \label{fig:h3-motion-example}
\end{figure}

\paragraph{Single-action LoRA diagnostics.}
Finally, single-action terminal-frame tests (Figure~\ref{fig:h3-action-control}; $30$ pairs/category, $29$ three-judge pairs on zoom and pose) and 39-frame camera tracking (Figure~\ref{fig:h3-motion-example}) corroborate why LoRA strengthens the simulator: LoRA raises lateral translation completion from $57.8\%$ to $88.9\%$, cuts off-target drift on selected expression and pose subsets ($66.7\%\to 6.7\%$ and $50.0\%\to 14.4\%$), and expands 39-frame lateral face-center shifts from $\sim\!8\%$ to $28$--$31\%$ width, giving the Selector a wide span of edit strengths.

%% file: sec/6_experiment_tables.tex
\newcommand{\reshootMainResults}{%
\begin{table*}[t]
\centering
\footnotesize
\captionsetup{skip=4pt}
\caption{\textbf{Aesthetic comparison on open-ended and targeted reshooting ($240$ portraits).} Relative mean-score changes (\%) over inputs on the $768\times1120$ canvas (PortraitAes-S/C/E: Structure, Composition, Emotion; dark/light green: best/second-best in Tables~\ref{tab:main-results}--\ref{tab:background}).}
\label{tab:main-results}
\setlength{\tabcolsep}{2pt}
\renewcommand{\arraystretch}{0.95}
\begin{tabularx}{\textwidth}{@{}cl*{6}{Y}@{}}
\toprule
\textbf{Task} & \textbf{Method} & \mbox{\textbf{PortraitAes-S}} & \mbox{\textbf{PortraitAes-C}} & \mbox{\textbf{PortraitAes-E}} & \textbf{PhotoFramer} & \textbf{Q-Align} & \textbf{ArtiMuse} \\
\midrule
\multirow{6}{*}{\shortstack{\textit{Open-ended}\\\textit{reshooting}}} & Qwen-Image 2.1 & $+6.69$ & $+4.93$ & $+1.76$ & $+1.87$ & $+3.11$ & \resultshade{ResultSecond}{$+12.03$} \\
 & FLUX~3 Klein & $+4.25$ & $+4.55$ & $-1.37$ & $+2.72$ & \resultshade{ResultSecond}{$+7.60$} & $+2.85$ \\
 & Nano Banana 2.1 & \resultshade{ResultSecond}{$+14.65$} & \resultshade{ResultBest}{\resultemphasis{$+7.17$}} & \resultshade{ResultBest}{\resultemphasis{$+3.93$}} & $+3.59$ & $+0.38$ & \resultshade{ResultBest}{\resultemphasis{$+13.32$}} \\
 & ChronoEdit & $+2.16$ & $+0.99$ & $-0.92$ & $+2.17$ & $+4.27$ & $+1.99$ \\
 & AesFormer & $-2.36$ & $+4.42$ & $-0.95$ & \resultshade{ResultSecond}{$+5.03$} & $+7.11$ & $+3.27$ \\
\cmidrule(l){2-8}
 & \textbf{SimShoot (ours)} & \resultshade{ResultBest}{\resultemphasis{$+15.47$}} & \resultshade{ResultSecond}{$+5.05$} & \resultshade{ResultSecond}{$+2.79$} & \resultshade{ResultBest}{\resultemphasis{$+5.10$}} & \resultshade{ResultBest}{\resultemphasis{$+10.15$}} & $+5.35$ \\
\midrule
\multirow{6}{*}{\shortstack{\textit{Targeted}\\\textit{reshooting}}} & Qwen-Image 2.1 & $+0.57$ & \resultshade{ResultSecond}{$+2.00$} & \resultshade{ResultSecond}{$+0.06$} & $+2.56$ & $+1.77$ & \resultshade{ResultBest}{\resultemphasis{$+7.26$}} \\
 & FLUX~3 Klein & $-0.05$ & $+1.42$ & $-0.27$ & $+2.39$ & $+3.38$ & $+1.39$ \\
 & Nano Banana 2.1 & \resultshade{ResultSecond}{$+5.85$} & $-0.49$ & $-0.48$ & $-0.49$ & $-2.28$ & $-1.72$ \\
 & ChronoEdit & $-3.81$ & $-0.80$ & $-3.24$ & $+1.40$ & \resultshade{ResultSecond}{$+4.08$} & $-0.07$ \\
 & AesFormer & $-10.57$ & $+0.64$ & $-1.34$ & \resultshade{ResultBest}{\resultemphasis{$+4.29$}} & \resultshade{ResultBest}{\resultemphasis{$+4.27$}} & \resultshade{ResultSecond}{$+4.57$} \\
\cmidrule(l){2-8}
 & \textbf{SimShoot (ours)} & \resultshade{ResultBest}{\resultemphasis{$+10.16$}} & \resultshade{ResultBest}{\resultemphasis{$+2.08$}} & \resultshade{ResultBest}{\resultemphasis{$+0.54$}} & \resultshade{ResultSecond}{$+2.78$} & $+3.51$ & $+1.77$ \\
\bottomrule
\end{tabularx}
\end{table*}
}

\newcommand{\reshootAdherenceTable}{%
\begin{table}[t]
\centering
\footnotesize
\captionsetup{skip=3pt}
\caption{\textbf{Instruction following in targeted reshooting.} GPT-6.1 Sol, Gemini 3.8 Flash, and Qwen3.8 Max score per-portrait completion (\%) on the $118/120$ three-judge portraits.}
\label{tab:adherence}
\setlength{\tabcolsep}{2pt}
\renewcommand{\arraystretch}{0.94}
\begin{tabularx}{\columnwidth}{@{}l*{4}{Y}@{}}
\toprule
\textbf{Method} & \mbox{\textbf{GPT-6.1}} & \mbox{\textbf{Gemini 3.8}} & \mbox{\textbf{Qwen 3.8}} & \textbf{Avg.} \\
\midrule
Qwen-Image 2.1 & 71.61 & 57.77 & 71.47 & 66.95 \\
FLUX~3 Klein & \resultshade{ResultSecond}{75.85} & \resultshade{ResultSecond}{66.88} & \resultshade{ResultSecond}{83.47} & \resultshade{ResultSecond}{75.40} \\
Nano Banana 2.1 & 72.18 & 61.72 & 76.55 & 70.15 \\
ChronoEdit & 63.49 & 54.10 & 60.59 & 59.39 \\
AesFormer & 50.21 & 41.60 & 51.27 & 47.69 \\
\cmidrule{1-5}
\textbf{SimShoot (ours)} & \resultshade{ResultBest}{\resultemphasis{90.40}} & \resultshade{ResultBest}{\resultemphasis{75.21}} & \resultshade{ResultBest}{\resultemphasis{85.52}} & \resultshade{ResultBest}{\resultemphasis{83.71}} \\
\bottomrule
\end{tabularx}
\end{table}
}

\newcommand{\reshootBackgroundTable}{%
\begin{table}[t]
\centering
\footnotesize
\captionsetup{skip=4pt}
\caption{\textbf{Background preservation under fixed capture.} Evaluated across $60$ fixed-capture targeted portraits ($134$ pose and expression actions) outside the shared six-method foreground union.}
\label{tab:background}
\setlength{\tabcolsep}{3pt}
\renewcommand{\arraystretch}{0.94}
\begin{tabularx}{\columnwidth}{@{}l*{4}{Y}@{}}
\toprule
\textbf{Method} & PSNR $\uparrow$ & SSIM $\uparrow$ & LPIPS $\downarrow$ & $\lvert\Delta L^*\rvert$ $\downarrow$ \\
\midrule
Qwen-Image 2.1 & 19.73 & 0.5369 & 0.1343 & 7.01 \\
FLUX~3 Klein & \resultshade{ResultBest}{\resultemphasis{33.76}} & \resultshade{ResultBest}{\resultemphasis{0.9421}} & \resultshade{ResultBest}{\resultemphasis{0.0268}} & \resultshade{ResultBest}{\resultemphasis{1.58}} \\
Nano Banana 2.1 & 21.33 & 0.6129 & 0.1244 & 5.59 \\
ChronoEdit & 18.80 & 0.5663 & 0.2105 & 8.28 \\
AesFormer & 19.18 & 0.5634 & 0.3173 & 7.83 \\
\cmidrule{1-5}
\textbf{SimShoot (ours)} & \resultshade{ResultSecond}{27.30} & \resultshade{ResultSecond}{0.7916} & \resultshade{ResultSecond}{0.1083} & \resultshade{ResultSecond}{3.13} \\
\bottomrule
\end{tabularx}
\end{table}
}

\newcommand{\reshootHumanPreferenceTable}{%
\begin{table}[t]
\centering
\footnotesize
\captionsetup{skip=4pt}
\caption{\textbf{Crowd-sourced photographic preference ($12{,}000$ votes across $240$ portraits).} SimShoot pairwise win rate (\%) overall ($2{,}400$ votes/row) and by track ($1{,}200$/track); \emph{Overall $p$} is five-baseline Holm-adjusted. Dark and light green mark $\text{Holm }p<0.01$ and $0.01\le\text{Holm }p<0.05$ within each column.}
\label{tab:human-preference}
\setlength{\tabcolsep}{2.5pt}
\renewcommand{\arraystretch}{0.96}
\begin{tabularx}{\columnwidth}{@{}l*{4}{Y}@{}}
\toprule
\textbf{vs.\ Baseline} & \textbf{Overall} & \textbf{Open} & \textbf{Targeted} & \mbox{\textbf{Overall $p$}} \\
\midrule
Qwen-Image 2.1 & \resultshade{ResultBest}{\resultemphasis{54.92}} & \resultshade{ResultSecond}{55.08} & \resultshade{ResultSecond}{54.75} & \resultshade{ResultBest}{$<0.001$} \\
FLUX~3 Klein & \resultshade{ResultBest}{\resultemphasis{55.00}} & 53.42 & \resultshade{ResultBest}{\resultemphasis{56.58}} & \resultshade{ResultBest}{$<0.001$} \\
Nano Banana 2.1 & 49.67 & 48.33 & 51.00 & $0.789$ \\
ChronoEdit & \resultshade{ResultBest}{\resultemphasis{60.79}} & \resultshade{ResultBest}{\resultemphasis{56.75}} & \resultshade{ResultBest}{\resultemphasis{64.83}} & \resultshade{ResultBest}{$<0.001$} \\
AesFormer & \resultshade{ResultBest}{\resultemphasis{56.33}} & \resultshade{ResultSecond}{54.83} & \resultshade{ResultBest}{\resultemphasis{57.83}} & \resultshade{ResultBest}{$<0.001$} \\
\bottomrule
\end{tabularx}
\end{table}
}

\newcommand{\ablsub}[1]{\mbox{\scriptsize #1}}
\newcommand{\ablsep}{\arrayrulecolor{black!22}\midrule[0.35pt]\arrayrulecolor{black}}
\newcommand{\ablbest}[1]{\resultshade{ResultBest}{\resultemphasis{#1}}}
\newcommand{\reshootClosedLoopAblationTable}{%
\begin{table*}[t]
\centering
\footnotesize
\captionsetup{skip=3pt}
\caption{\textbf{Closed-loop component ablations across all $240$ benchmark portraits.} Each block isolates one module in the SimShoot reshooting loop while holding other modules fixed; variants within each block share an aligned evaluation protocol on identical inputs.}
\label{tab:closed-loop-ablations}
\setlength{\tabcolsep}{0pt}
\setlength{\aboverulesep}{1.4pt}
\setlength{\belowrulesep}{1.9pt}
\renewcommand{\arraystretch}{0.85}
\begin{tabularx}{\textwidth}{@{}c@{\hspace{8pt}}l@{\hspace{9pt}}>{\centering\arraybackslash}p{42pt}@{\hspace{4pt}}>{\centering\arraybackslash}p{42pt}@{\hspace{9pt}}>{\centering\arraybackslash}p{42pt}@{\hspace{4pt}}>{\centering\arraybackslash}p{42pt}@{\hspace{11pt}}Y@{\hspace{6pt}}Y@{}}
\toprule
\multirow{2}{*}[1pt]{\textbf{Ablated module}} & \multirow{2}{*}[1pt]{\textbf{Variant}} & \multicolumn{2}{c@{\hspace{9pt}}}{\textbf{PortraitAes-S} $\uparrow$} & \multicolumn{2}{c@{\hspace{11pt}}}{\textbf{Q-Align} $\uparrow$} & \multicolumn{2}{c@{}}{\multirow{2}{*}[1pt]{\textbf{Module-specific metrics}}} \\[-0.5pt]
\cmidrule(r{9pt}){3-4}\cmidrule(r{11pt}){5-6}
 & & \ablsub{Open-ended} & \ablsub{Targeted} & \ablsub{Open-ended} & \ablsub{Targeted} & \multicolumn{2}{c@{}}{} \\[-1pt]
\midrule
\multirow{3}{*}[-2pt]{\shortstack{\textit{(a) Simulator}\\\textit{adaptation}}}
 & & & & & & \ablsub{Background PSNR~$\uparrow$} & \ablsub{Edit acceptance~$\uparrow$} \\[-1.5pt]
\cmidrule(r{3pt}){7-7}\cmidrule(l{3pt}){8-8}
 & \textbf{SimShoot} & \ablbest{$+15.47\%$} & \ablbest{$+10.16\%$} & \ablbest{$+10.15\%$} & \ablbest{$+3.51\%$} & \ablbest{$27.08$\,dB} & \ablbest{$82.3\%$} \\
 & w/o H3 action LoRAs & $+12.34\%$ & $+8.58\%$ & $+6.24\%$ & $+2.69\%$ & $24.02$\,dB & $59.4\%$ \\
\ablsep
\multirow{4}{*}[-2pt]{\shortstack{\textit{(b) Simulator}\\\textit{replacement}}}
 & & & & & & \ablsub{Monotonic actions~$\uparrow$} & \ablsub{Gen.\ calls / portrait~$\downarrow$} \\[-1.5pt]
\cmidrule(r{3pt}){7-7}\cmidrule(l{3pt}){8-8}
 & \textbf{SimShoot} & \ablbest{$+15.47\%$} & \ablbest{$+10.16\%$} & \ablbest{$+10.15\%$} & \ablbest{$+3.51\%$} & \ablbest{$92.9\%$} & \ablbest{$3.29$} \\
 & Simulator $\to$ FLUX~3 Klein & $+2.98\%$ & $+1.63\%$ & $+5.69\%$ & $+1.13\%$ & $16.7\%$ & $18.88$ \\
 & Simulator $\to$ Qwen-Image 2.1 & $+7.97\%$ & $+1.94\%$ & $+4.83\%$ & $-2.41\%$ & $23.8\%$ & $27.14$ \\
\ablsep
\multirow{3}{*}[-2pt]{\shortstack{\textit{(c) Closed-loop}\\\textit{planning}}}
 & & & & & & \ablsub{Accepted edits~$\uparrow$} & \ablsub{Exhausted (open)~$\downarrow$} \\[-1.5pt]
\cmidrule(r{3pt}){7-7}\cmidrule(l{3pt}){8-8}
 & \textbf{SimShoot} & \ablbest{$+15.47\%$} & \ablbest{$+10.16\%$} & \ablbest{$+10.15\%$} & $+3.51\%$ & \ablbest{$650$} & \ablbest{$0.0\%$} \\
 & w/ fixed initial sequence & $+12.61\%$ & $+9.44\%$ & $+8.83\%$ & \ablbest{$+3.61\%$} & $594$ & $30.0\%$ \\
\ablsep
\multirow{4}{*}[-2pt]{\shortstack{\textit{(d) Extent}\\\textit{selection}\textsuperscript{\dag}}}
 & & & & & & \ablsub{Open-ended accept.~$\uparrow$} & \ablsub{Targeted accept.~$\uparrow$} \\[-1.5pt]
\cmidrule(r{3pt}){7-7}\cmidrule(l{3pt}){8-8}
 & \textbf{SimShoot} & \ablbest{$+4.32\%$} & $+6.65\%$ & $+1.70\%$ & \ablbest{$+1.49\%$} & $79.2\%$ & $89.2\%$ \\
 & w/ coarse-5 only & $+3.43\%$ & \ablbest{$+7.26\%$} & \ablbest{$+1.82\%$} & $+1.35\%$ & \ablbest{$80.8\%$} & \ablbest{$93.3\%$} \\
 & w/ endpoint-only selection & $+0.82\%$ & $+5.21\%$ & $+0.97\%$ & $+1.37\%$ & $54.2\%$ & $80.0\%$ \\
\bottomrule
\end{tabularx}
\\[1pt]
{\scriptsize\raggedright PortraitAes-S and Q-Align report relative mean-score gains (\%) over inputs ($120$/track; all six metrics in Table~\ref{tab:supp-ablation-full}; \textsuperscript{\dag}block~(d) replays $1$ shared rollout/portrait). Background PSNR uses the two-arm foreground union (vs.\ $27.30$\,dB under Table~\ref{tab:background}'s six-method union); edit acceptance includes fallbacks; monotonic actions ($42$) satisfy \texttt{F10}$\le$\texttt{F19}$\le$\texttt{F29}$\le$\texttt{F38} (non-constant); exhausted runs finish the fixed list ($25/36$ after rejections).\par}
\end{table*}
}

%% file: sec/7_conclusion.tex
\section{Conclusion and Limitations}
\label{sec:conclusion}
We presented \textbf{SimShoot}, a closed-loop framework for aesthetic portrait reshooting through simulated photographic actions and visual extent selection.
Decoupling the choice of a shooting adjustment from the choice of how far to execute it turns delicate continuous parameter tuning into direct visual comparison along a simulated transition.
Supported by single-action video supervision across BlenderSynth, RealSynth, and linear-light relighting routes, SimShoot achieves strong gains in portrait aesthetics, leads multi-VLM instruction-following judges with human-rated action completion comparable to the strongest editors, matches proprietary editing appeal in human preference while preserving scene fidelity, and outperforms iterative image-editor loops in extent monotonicity and multi-round stability.

\paragraph{Limitations and future work.}
First, rolling out multi-frame video transitions and querying a reasoning VLM across sequential rounds incurs higher inference latency than single-pass image editors; distilling few-step action simulators, adapting rollout length to each action, and pairing the simulator with lighter local VLMs can substantially accelerate interactive use.
Second, while sequential single-action steps with Selector gating keep visual drift well controlled over typical few-round reshoots, much longer multi-round horizons would scale runtime linearly and risk accumulating subtle texture or fidelity drift across passes.
Third, our current training pools and benchmark focus on single-person portraits, and large camera zoom-outs or wide viewpoint shifts synthesize previously out-of-frame regions from generative priors rather than recovering unobserved physical scene geometry; extending action simulation to multi-person group portraits is a promising direction for future work.

\paragraph{Ethics and broader impact.}
Crowd-sourced evaluations collected only minimal-risk, anonymous portrait ratings under Rapidata's Terms of Service and Privacy Policy (voluntary opt-in, withdrawal at any time, platform-managed compensation), with no personally identifiable information or persistent rater IDs.
Source portraits derive from PortraitCraft~\cite{lou1stPortraitCraftChallenge2026} and AesRecon~\cite{duAesFormerTransformEveryday2026} under their research terms.
Restricting edits to an identity-preserving photoshoot action space rather than face swapping or retouching curbs identity fabrication, though unauthorized modification and upstream demographic bias still warrant consent, provenance watermarking, and fairness auditing.

%% file: sec/X_suppl.tex
This appendix provides complete six-metric quantitative breakdowns and visual comparisons for the four closed-loop component ablations in Section~\ref{sec:ablations} (Appendix~\ref{sec:supp-qual-ablations}), step-by-step multi-round trajectory case studies (Appendix~\ref{sec:supp-case-studies}), and the rater interfaces used in our crowd-sourced human evaluation in Section~\ref{sec:human-evaluation} (Appendix~\ref{sec:supp-human-visuals}).
Synchronized 39-frame video rollouts and interactive multi-round walkthroughs are available on the Project Page (\href{https://shenouy03.github.io/simshoot}{\texttt{shenouy03.github.io/simshoot}}).

\section{Quantitative and Visual Ablations of Core Components}
\label{sec:supp-qual-ablations}

Table~\ref{tab:supp-ablation-full} reports relative mean-score gains (\%) across all six aesthetic evaluators from Table~\ref{tab:main-results} for every closed-loop ablation in Table~\ref{tab:closed-loop-ablations}.
Across multi-round ablations, SimShoot leads all six open-ended and targeted metrics in simulator adaptation (a), all six open-ended and five of six targeted metrics in closed-loop planning (c), and five of six metrics on both tracks in simulator replacement (b).
In block~(b), Qwen-Image~2.1's higher ArtiMuse gain co-occurs with negative targeted gains on PortraitAes-C, PortraitAes-E, and Q-Align, consistent with ArtiMuse rewarding contrast and sharpening artifacts from repeated re-synthesis (Section~\ref{sec:human-evaluation}).
In single-round replay (d), pooled over $240$ rollouts, both SimShoot and coarse-5 significantly beat endpoint-only on PortraitAes-S and Q-Align, while SimShoot vs.\ coarse-5 (and the other four metrics) spans zero.

\begin{table}[!htbp]
\centering
\footnotesize
\captionsetup{skip=3pt}
\caption{\textbf{Six-metric aesthetic results across all four component ablations ($240$ portraits; $120$/track).} Relative mean-score changes (\%) from input to output on the $768\times1120$ canvas across the six evaluators of Table~\ref{tab:main-results}. Shaded cells mark the highest gain per block and track.}
\label{tab:supp-ablation-full}
\setlength{\tabcolsep}{0.6pt}
\setlength{\aboverulesep}{1.5pt}
\setlength{\belowrulesep}{1.8pt}
\renewcommand{\arraystretch}{0.88}
\begin{tabularx}{\linewidth}{@{}c@{\hspace{4pt}}l@{\hspace{4pt}}l*{6}{Y}@{}}
\toprule
\textbf{Module} & \textbf{Track} & \textbf{Variant} & \mbox{\textbf{PortraitAes-S}} & \mbox{\textbf{PortraitAes-C}} & \mbox{\textbf{PortraitAes-E}} & \textbf{PhotoFramer} & \textbf{Q-Align} & \textbf{ArtiMuse} \\
\midrule
\multirow{4}{*}{\shortstack{\textit{(a) Simulator}\\\textit{adaptation}}}
 & \multirow{2}{*}{Open-ended} & \textbf{SimShoot} & \ablbest{$+15.47$} & \ablbest{$+5.05$} & \ablbest{$+2.79$} & \ablbest{$+5.10$} & \ablbest{$+10.15$} & \ablbest{$+5.35$} \\
 &  & w/o H3 action LoRAs & $+12.34$ & $+2.25$ & $+0.61$ & $+2.72$ & $+6.24$ & $+0.27$ \\
\cmidrule(l){2-9}
 & \multirow{2}{*}{Targeted} & \textbf{SimShoot} & \ablbest{$+10.16$} & \ablbest{$+2.08$} & \ablbest{$+0.54$} & \ablbest{$+2.78$} & \ablbest{$+3.51$} & \ablbest{$+1.77$} \\
 &  & w/o H3 action LoRAs & $+8.58$ & $+0.80$ & $+0.38$ & $+1.17$ & $+2.69$ & $-1.70$ \\
\ablsep
\multirow{6}{*}{\shortstack{\textit{(b) Simulator}\\\textit{replacement}}}
 & \multirow{3}{*}{Open-ended} & \textbf{SimShoot} & \ablbest{$+15.47$} & \ablbest{$+5.05$} & \ablbest{$+2.79$} & \ablbest{$+5.10$} & \ablbest{$+10.15$} & $+5.35$ \\
 &  & Simulator $\to$ FLUX~3 Klein & $+2.98$ & $+3.99$ & $+0.36$ & $+2.89$ & $+5.69$ & $+6.49$ \\
 &  & Simulator $\to$ Qwen-Image 2.1 & $+7.97$ & $+4.66$ & $+0.95$ & $+3.31$ & $+4.83$ & \ablbest{$+16.97$} \\
\cmidrule(l){2-9}
 & \multirow{3}{*}{Targeted} & \textbf{SimShoot} & \ablbest{$+10.16$} & \ablbest{$+2.08$} & \ablbest{$+0.54$} & \ablbest{$+2.78$} & \ablbest{$+3.51$} & $+1.77$ \\
 &  & Simulator $\to$ FLUX~3 Klein & $+1.63$ & $+1.53$ & $-0.82$ & $+1.20$ & $+1.13$ & $+4.81$ \\
 &  & Simulator $\to$ Qwen-Image 2.1 & $+1.94$ & $-1.37$ & $-3.29$ & $+0.26$ & $-2.41$ & \ablbest{$+6.55$} \\
\ablsep
\multirow{4}{*}{\shortstack{\textit{(c) Closed-loop}\\\textit{planning}}}
 & \multirow{2}{*}{Open-ended} & \textbf{SimShoot} & \ablbest{$+15.47$} & \ablbest{$+5.05$} & \ablbest{$+2.79$} & \ablbest{$+5.10$} & \ablbest{$+10.15$} & \ablbest{$+5.35$} \\
 &  & w/ fixed initial sequence & $+12.61$ & $+4.33$ & $+2.04$ & $+4.59$ & $+8.83$ & $+3.26$ \\
\cmidrule(l){2-9}
 & \multirow{2}{*}{Targeted} & \textbf{SimShoot} & \ablbest{$+10.16$} & \ablbest{$+2.08$} & \ablbest{$+0.54$} & \ablbest{$+2.78$} & $+3.51$ & \ablbest{$+1.77$} \\
 &  & w/ fixed initial sequence & $+9.44$ & $+1.35$ & $-0.06$ & $+2.71$ & \ablbest{$+3.61$} & $+1.65$ \\
\ablsep
\multirow{6}{*}{\shortstack{\textit{(d) Extent}\\\textit{selection}\textsuperscript{\dag}}}
 & \multirow{3}{*}{Open-ended} & \textbf{SimShoot} & \ablbest{$+4.32$} & \ablbest{$+1.34$} & $+0.82$ & $+1.74$ & $+1.70$ & $+2.86$ \\
 &  & w/ coarse-5 only & $+3.43$ & $+1.32$ & $+0.77$ & \ablbest{$+1.99$} & \ablbest{$+1.82$} & $+2.61$ \\
 &  & w/ endpoint-only selection & $+0.82$ & $+1.07$ & \ablbest{$+0.94$} & $+1.84$ & $+0.97$ & \ablbest{$+3.09$} \\
\cmidrule(l){2-9}
 & \multirow{3}{*}{Targeted} & \textbf{SimShoot} & $+6.65$ & \ablbest{$+1.10$} & \ablbest{$+0.46$} & \ablbest{$+0.83$} & \ablbest{$+1.49$} & $+0.25$ \\
 &  & w/ coarse-5 only & \ablbest{$+7.26$} & \ablbest{$+1.10$} & $+0.17$ & $+0.69$ & $+1.35$ & $+0.10$ \\
 &  & w/ endpoint-only selection & $+5.21$ & $+0.57$ & $+0.42$ & $+0.71$ & $+1.37$ & \ablbest{$+0.54$} \\
\bottomrule
\end{tabularx}
\\[1.5pt]
{\scriptsize\raggedright\textsuperscript{\dag}Block~(d) replays $1$ shared 39-frame rollout/portrait ($240$ total), comparing archived online SimShoot decisions against replayed coarse-5 (refinement disabled; $131/198$ exact frame matches on unrefined rollouts) and endpoint-only. Pooled ($N=240$), SimShoot ($+5.47\%$ Structure, $+1.60\%$ Q-Align) and coarse-5 ($+5.32\%$, $+1.59\%$) both beat endpoint-only ($+2.99\%$, $+1.17\%$; SimShoot vs.\ endpoint-only $\Delta=+2.48$\,pp $[+0.86,+4.46]$ and $+0.43$\,pp $[+0.11,+0.78]$), whereas SimShoot vs.\ coarse-5 spans zero overall ($+0.15$\,pp $[-1.18,+1.49]$, $+0.01$\,pp $[-0.22,+0.25]$) and on the $42$ refined rollouts ($-1.06$\,pp $[-5.35,+1.49]$, $+0.24$\,pp $[-0.44,+1.07]$).\par}
\end{table}

\begin{figure}[!htbp]
    \centering
    \captionsetup{skip=3pt}
    \input{sec/fig_supp_b1}
    \caption{\textbf{Action-specific LoRAs vs.\ pretrained H3 on matched Round-1 rollouts (\texttt{F0} and terminal \texttt{F38}).}
    Starting from identical \texttt{F0} inputs and prompts:
    (a)~Case~1 (camera orbit): native H3 leaves the low viewpoint unchanged whereas LoRA raises the camera cleanly;
    (b)~Case~2 (camera zoom): native H3 yields minimal zoom whereas LoRA reaches a close-up;
    (c)~Case~3 (body pose): native H3 leaves the right fingers outside the pocket whereas LoRA tucks them cleanly inside;
    (d)~Case~4 (lighting): native H3 warms unevenly while drifting the glass downward, whereas LoRA warms uniformly with fixed posture and glass.}
    \label{fig:supp-b1-lora}
\end{figure}
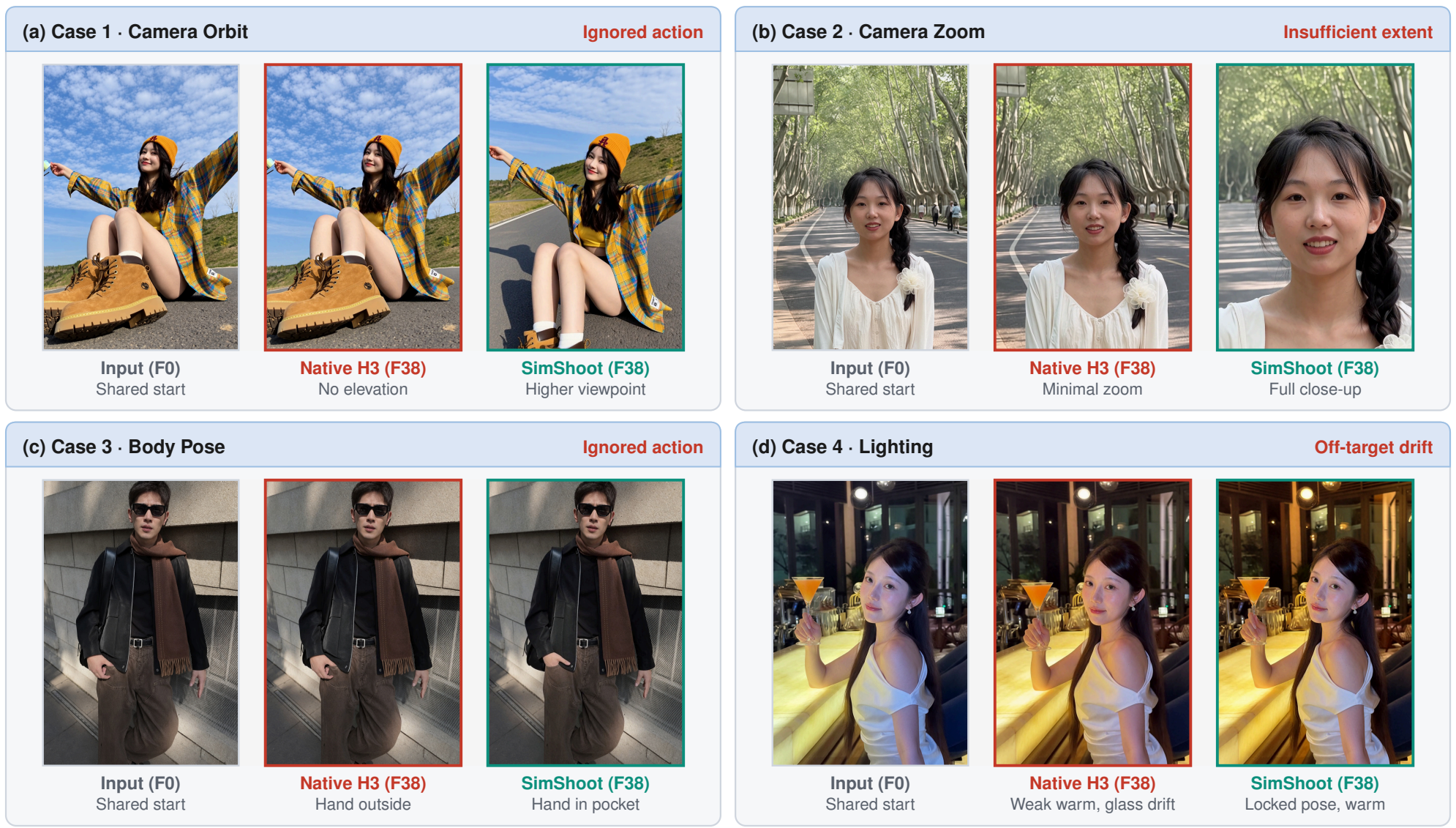

Below, we examine visual comparisons for each ablated module in turn.
Figure~\ref{fig:supp-b1-lora} compares matched single-action rollouts with and without action-specific LoRA adapters across camera orbit, focal-length zoom, body pose, and portrait lighting, highlighting three primary failure modes of the unadapted video backbone: ignoring geometric or postural commands, stalling at minimal extent, and introducing off-target drift.

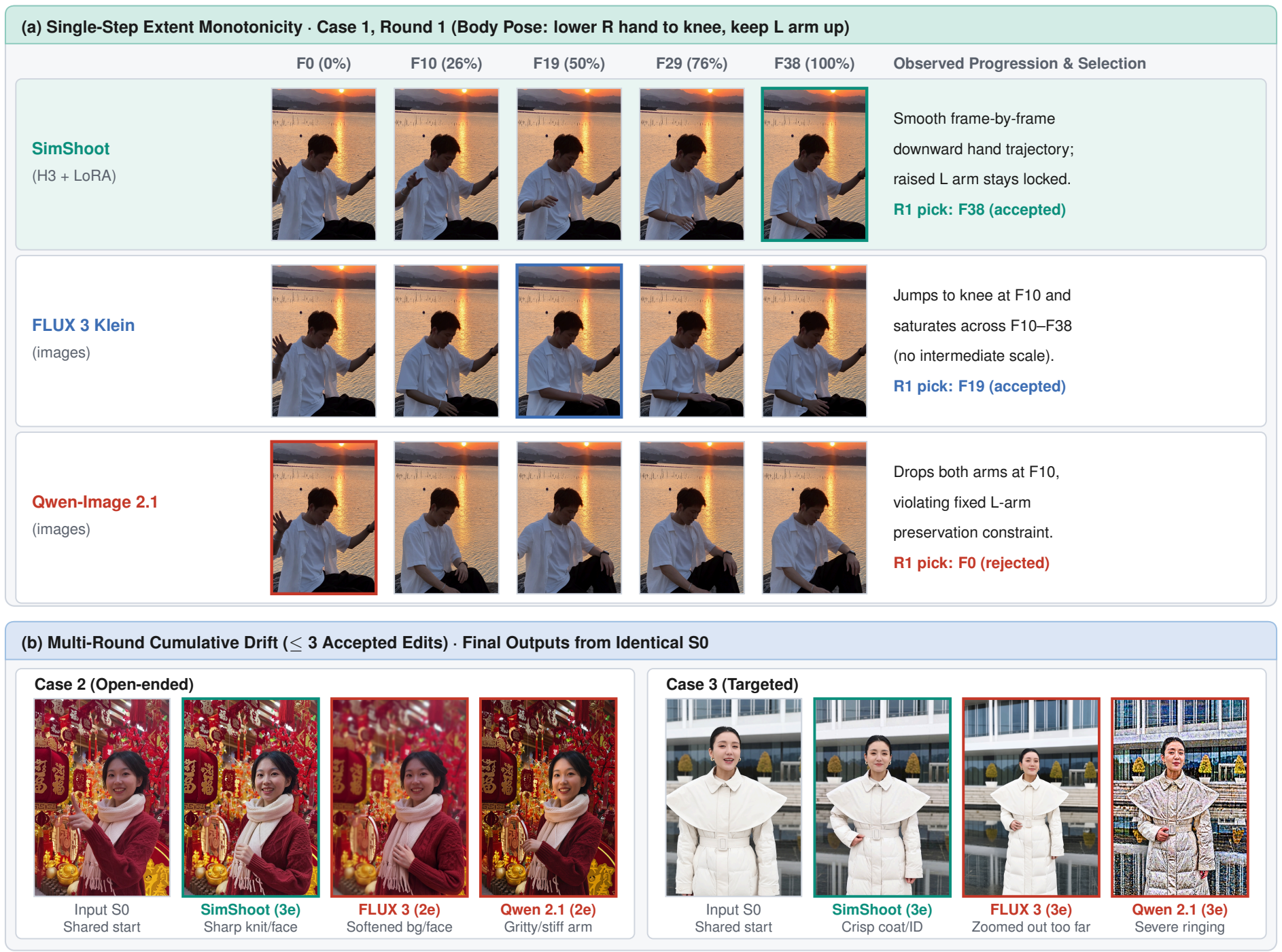
\begin{figure}[!htbp]
    \centering
    \captionsetup{skip=3pt}
    \input{sec/fig_supp_b4}
    \caption{\textbf{Video action simulation vs.\ extent-prompted image editing inside the same closed loop.}
    (a)~Single-step extent progression (Case~1, held-out pilot portrait outside the $240$, Round~1; lower the right hand to the knee while keeping the raised left arm fixed): SimShoot produces a smooth downward trajectory across \texttt{F0}--\texttt{F38}, whereas FLUX~3 Klein saturates at \texttt{F10} (\texttt{F19} selected) and Qwen-Image~2.1 drops both arms (\texttt{F0} rejected).
    (b)~Multi-round drift within $\le 3$ accepted edits on Case~2 (open-ended) and Case~3 (targeted): SimShoot preserves crisp detail, whereas FLUX~3 Klein softens the face/background (Case~2) or over-zooms out (Case~3), and Qwen-Image~2.1 adds gritty shadow blotches and halo ringing artifacts.}
    \label{fig:supp-b4-executor}
\end{figure}

Figure~\ref{fig:supp-b4-executor} compares our video simulator against extent-prompted image editors (FLUX~3 Klein and Qwen-Image~2.1) in the same closed loop, contrasting smooth single-step progression with non-monotonic jumps and multi-round re-synthesis drift.

\FloatBarrier

\begin{figure}[!htbp]
    \centering
    \captionsetup{skip=3pt}
    \input{sec/fig_supp_b3}
    \caption{\textbf{Adaptive closed-loop replanning vs.\ fixed initial sequence.}
    (a)~Dynamic plan adaptation (Case~1): after the shared Round-1 smile, the fixed plan applies only fill light and leaves the busy background mural sharp, whereas adaptive replanning inspects \texttt{S1} and appends zoom-in and aperture widening to isolate the subject.
    (b)~Recovery from rejection (Case~2): after Round~1 lowers the raised right arm with unintended mouth drift and is rejected (\texttt{F0}), the fixed plan fails to recover and leaves the arm raised, whereas adaptive replanning reformulates Round~2 with explicit lip preservation and completes the reshoot.}
    \label{fig:supp-b3-replanning}
\end{figure}
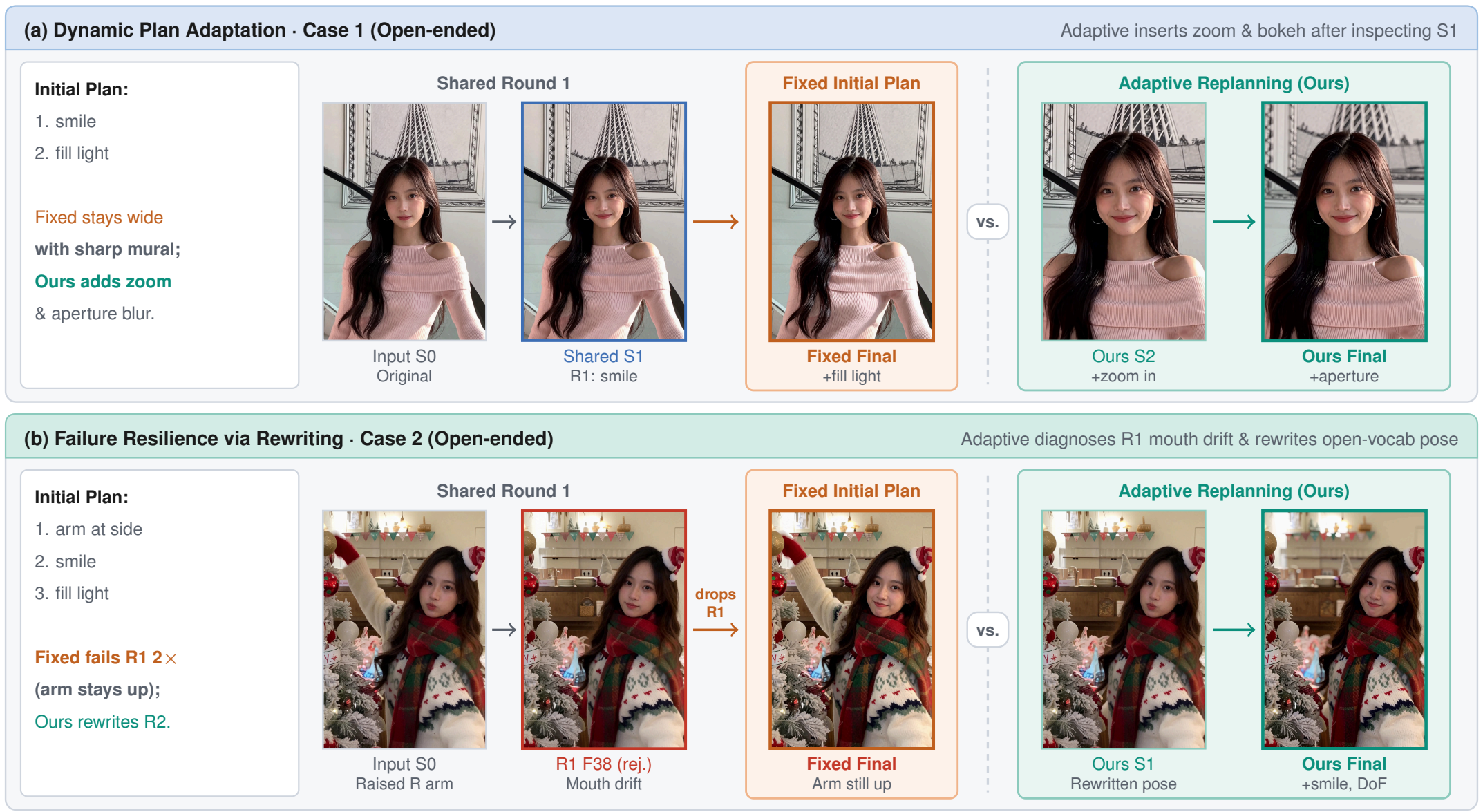

Figure~\ref{fig:supp-b3-replanning} compares closed-loop adaptive replanning against open-loop execution of a fixed sequence from Round~1, illustrating emergent compositional refinement from intermediate states and targeted rewriting after rejection.

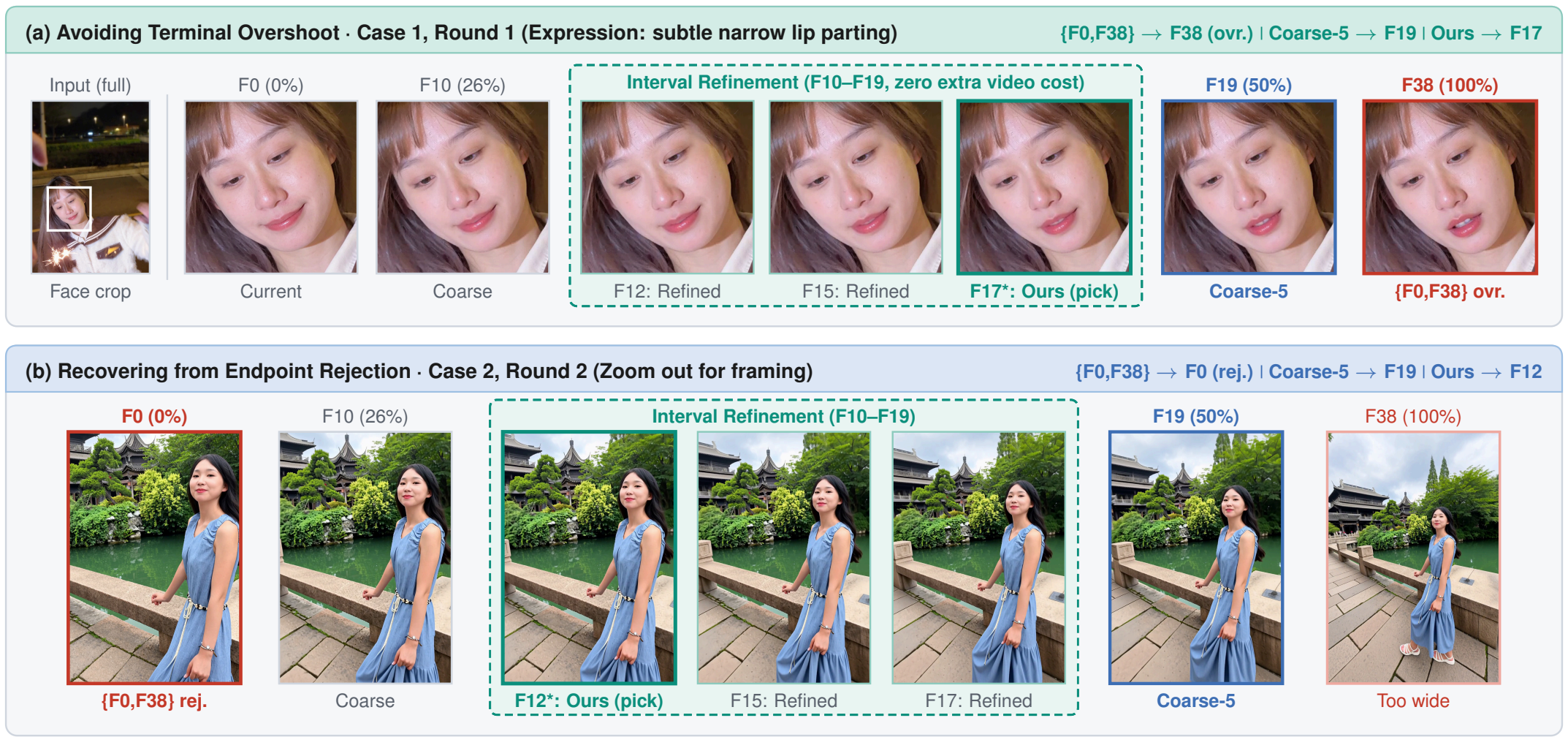
\begin{figure}[!htbp]
    \centering
    \captionsetup{skip=3pt}
    \input{sec/fig_supp_b2}
    \caption{\textbf{Two-stage coarse-to-fine frame selection vs.\ endpoint-only selection on identical 39-frame rollouts.}
    (a)~Avoiding terminal overshoot (Case~1, Round~1; face crops shown alongside full \texttt{F0}): \texttt{\{F0, F38\}} accepts a wider mouth opening with visible teeth at \texttt{F38}, \texttt{Coarse-5} selects \texttt{F19}, and refining \texttt{F10}--\texttt{F19} pinpoints subtle lip parting at \texttt{F17}.
    (b)~Recovering from endpoint rejection (Case~2, Round~2): \texttt{F38} zooms out too far to a distant full-body view and is rejected (\texttt{F0}) under \texttt{\{F0, F38\}}, whereas coarse-to-fine inspection refines \texttt{F10}--\texttt{F19} and selects balanced framing at \texttt{F12}.}
    \label{fig:supp-b2-extent}
\end{figure}

Figure~\ref{fig:supp-b2-extent} contrasts two-stage coarse-to-fine frame selection against endpoint-only (\texttt{\{F0, F38\}}) selection on identical 39-frame rollouts, showing how inspecting interior frames and refining promising local intervals avoids late-frame overshoot and recovers balanced intermediate framings at zero extra video generation cost.

\FloatBarrier

\section{Multi-Round Reshooting Trajectories and Case Studies}
\label{sec:supp-case-studies}

\begin{figure}[!htbp]
    \centering
    \captionsetup{skip=3pt}
    \includegraphics[width=0.99\linewidth]{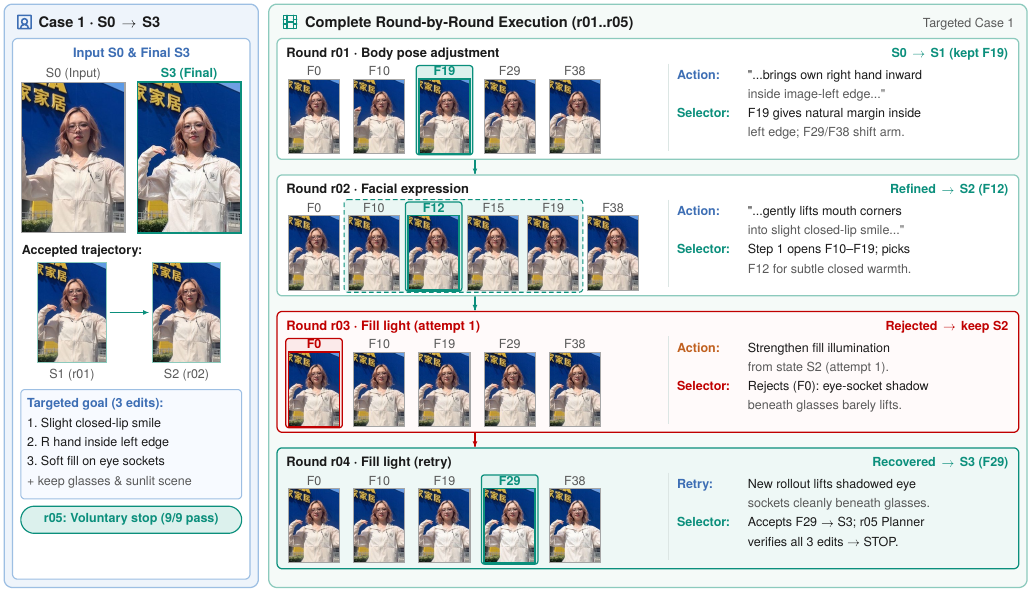}
    \caption{\textbf{Complete multi-round reshooting trajectory from input to voluntary stop (Case~1, \texttt{05900}, targeted).}
    Left: input \texttt{S0}, final portrait \texttt{S3}, intermediate accepted portraits \texttt{S1}--\texttt{S2}, and the three-part targeted request.
    Right: round-by-round execution (\texttt{r01}--\texttt{r05}) combining coarse selection (\texttt{r01} \texttt{F19}), interval refinement (\texttt{r02} \texttt{F12}), rejection (\texttt{r03} \texttt{F0}), retry (\texttt{r04} \texttt{F29}), and voluntary stop (\texttt{r05}).}
    \label{fig:supp-complete-run}
\end{figure}

Figure~\ref{fig:supp-complete-run} traces a complete targeted reshoot (Case~1, \texttt{05900}) from the input portrait \texttt{S0} to voluntary termination (\texttt{S3}) across a sunlit outdoor setting, where the user instruction requests three coordinated adjustments while preserving the subject's glasses, identity, and sunlit park backdrop.
The five-round trajectory illustrates how coarse interior selection (\texttt{r01} picking \texttt{F19} to bring the right hand cleanly inside the left frame margin as \texttt{S1}), local interval refinement (\texttt{r02} opening \texttt{F10}--\texttt{F19} to choose a subtle closed-lip smile at \texttt{F12} as \texttt{S2}), conservative rejection of an under-executed rollout (\texttt{r03} keeping \texttt{S2} via \texttt{F0} when shadow fill barely registers), targeted rollout retry (\texttt{r04} accepting \texttt{F29} as \texttt{S3} once shadowed eye sockets brighten cleanly beneath the glasses), and voluntary termination (\texttt{r05} once all three edits pass) work together in a single closed loop.

\begin{figure}[!htbp]
    \centering
    \captionsetup{skip=3pt}
    \includegraphics[width=0.99\linewidth]{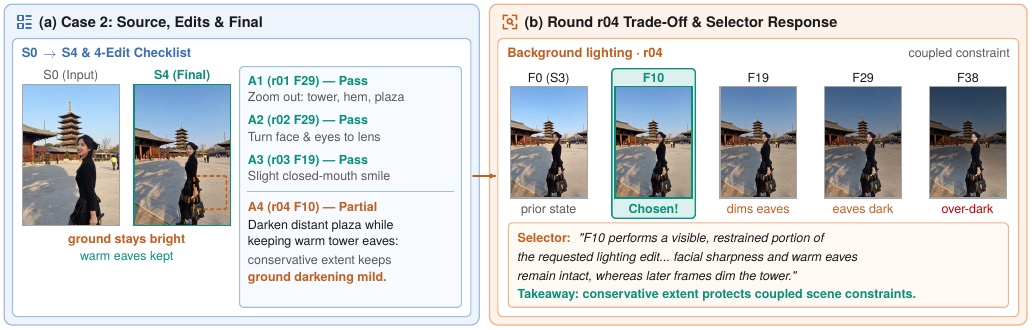}
    \caption{\textbf{Trade-off under coupled lighting preservation constraints (Case~2, \texttt{01867}, targeted).}
    (a)~Input \texttt{S0}, final portrait \texttt{S4}, and the four-edit checklist: framing (\texttt{A1}), head/gaze turn (\texttt{A2}), and closed-mouth smile (\texttt{A3}) are fully realized, whereas ground darkening (\texttt{A4}, orange box) stays mild.
    (b)~Round \texttt{r04}: \texttt{F19}--\texttt{F38} dim tower eaves alongside the ground, so the Selector picks \texttt{F10} to preserve constraints.}
    \label{fig:supp-incomplete-case}
\end{figure}

Figure~\ref{fig:supp-incomplete-case} examines a partially completed targeted request (Case~2, \texttt{01867}) involving four requested edits: zooming out to include the tower, skirt hem, and plaza (\texttt{A1}), turning the face and gaze toward the lens (\texttt{A2}), forming a slight closed-mouth smile (\texttt{A3}), and darkening the bright ground/plaza paving while preserving the warm sunlit glow on the tower eaves (\texttt{A4}).
Rounds \texttt{r01}--\texttt{r03} realize \texttt{A1}--\texttt{A3} cleanly (\texttt{3/3} VLM judge passes on each), advancing the portrait cleanly from \texttt{S0} to \texttt{S3}.
In Round~4 (\texttt{r04}), global background darkening progressively dims the bright paving across \texttt{F19}--\texttt{F38} but simultaneously darkens the warm wooden tower eaves in the upper background; inspecting the coarse sheet, the Selector chooses the conservative early frame \texttt{F10} (\texttt{S4}) to protect the eaves preservation constraint, leaving the ground darkening too mild to pass strict external action verification (\texttt{0/3} pass on \texttt{A4}).

\FloatBarrier

\section{Crowd-Evaluation Task Interfaces}
\label{sec:supp-human-visuals}

\begin{figure}[!htbp]
    \centering
    \captionsetup{skip=3pt}
    \begingroup
    \definecolor{suMut}{HTML}{595959}
    \resizebox{0.98\linewidth}{!}{%
    \begin{tikzpicture}[x=1bp,y=-1bp]
      \node[anchor=north west,inner sep=0pt,outer sep=0pt] at (0,0) {\includegraphics[width=495bp,height=228bp]{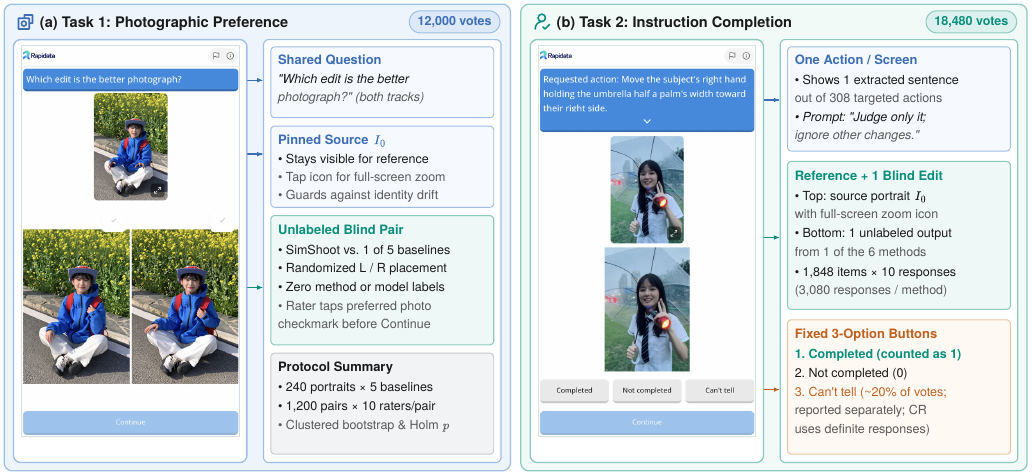}};
      \node[text=suMut,anchor=base,inner sep=0pt,font={\sffamily\fontsize{5.9bp}{6.6bp}\selectfont}] at (62.5,217.8) {Case 1 (blind screen)};
      \node[text=suMut,anchor=base,inner sep=0pt,font={\sffamily\fontsize{5.9bp}{6.6bp}\selectfont}] at (310.5,217.8) {Case 2 (blind screen)};
    \end{tikzpicture}%
    }
    \endgroup
    \caption{\textbf{Blinded crowd-evaluation interfaces on Rapidata.}
    (a)~Photographic preference (Task~1) pairs SimShoot with a baseline edit in randomized left--right order without method labels beneath the reference portrait $I_0$ (Case~1 shown).
    (b)~Targeted action completion (Task~2) displays one requested adjustment above $I_0$ and a single unlabeled edit with three choices (\emph{Completed}, \emph{Not completed}, \emph{Can't tell}; the in-screen footer note reflects the definite-response sensitivity metric reported alongside our prespecified all-response rate in Sec.~\ref{sec:human-evaluation}).}
    \label{fig:supp-rapidata-ui}
\end{figure}

Figure~\ref{fig:supp-rapidata-ui} shows the actual rater screens presented in our two blinded Rapidata studies (Sec.~\ref{sec:human-evaluation}).
Both tasks recruited English-speaking raters globally without method labels, pinned the original portrait $I_0$ on screen for direct reference, and collected ten independent responses per item ($12{,}000$ forced-choice votes across $1{,}200$ portrait pairs in Task~1, and $18{,}480$ single-action completion responses across $308$ targeted actions and six methods in Task~2).

%% file: sec/fig_supp_b1.tex
\begingroup%
\definecolor{sbW}{HTML}{FFFFFF}%
\definecolor{sbBg}{HTML}{F6F7F9}%
\definecolor{sbBdr}{HTML}{CFD5DE}%
\definecolor{sbInk}{HTML}{1A1A1A}%
\definecolor{sbMut}{HTML}{59606B}%
\definecolor{sbRule}{HTML}{E4E8EE}%
\definecolor{sbTeal}{HTML}{0B8F7C}%
\definecolor{sbTealBg}{HTML}{E9F5F2}%
\definecolor{sbTealHd}{HTML}{D3ECE6}%
\definecolor{sbTealBdr}{HTML}{8CCABD}%
\definecolor{sbRed}{HTML}{C43828}%
\definecolor{sbRedBg}{HTML}{FDEEED}%
\definecolor{sbRedBdr}{HTML}{EEA9A0}%
\definecolor{sbBlue}{HTML}{3D6EB4}%
\definecolor{sbBlueBg}{HTML}{EEF4FB}%
\definecolor{sbBlueHd}{HTML}{DCE6F5}%
\definecolor{sbBlueBdr}{HTML}{9DBFE3}%
\definecolor{sbOrg}{HTML}{C0601F}%
\definecolor{sbOrgBg}{HTML}{FDF2EA}%
\definecolor{sbOrgBdr}{HTML}{F0B48A}%
\resizebox{0.76\linewidth}{!}{%
\begin{tikzpicture}[x=1bp,y=-1bp]
  \path[fill=sbW] (0,0) rectangle (495,284);
  \path[fill=sbBg, draw=sbBdr, line width=0.55bp, rounded corners=3.5bp] (4.0,4.0) rectangle (245.0,140.0);
  \path[fill=sbBlueHd, draw=sbBlueBdr, line width=0.50bp] (4.0,19.0) -- (4.0,7.5) .. controls (4.0,5.2) and (5.2,4.0) .. (7.5,4.0) -- (241.5,4.0) .. controls (243.8,4.0) and (245.0,5.2) .. (245.0,7.5) -- (245.0,19.0) -- cycle;
  \node[text=sbInk,anchor=base west,inner sep=0pt,font={\sffamily\fontsize{6.4bp}{7.2bp}\selectfont\bfseries}] at (9.5,14.5) {(a) Case 1 \textperiodcentered{} Camera Orbit};
  \node[text=sbRed,anchor=base east,inner sep=0pt,font={\sffamily\fontsize{5.9bp}{6.8bp}\selectfont\bfseries}] at (239.5,14.5) {Ignored action};
  \node[anchor=north west,inner sep=0pt,outer sep=0pt] at (16.50,23.50) {\includegraphics[width=216.00bp,height=96.22bp]{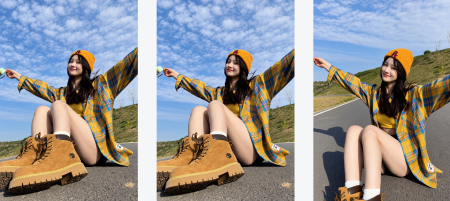}};
  \path[draw=sbBdr, line width=0.60bp] (16.50,23.50) rectangle (82.50,119.72);
  \path[draw=sbRed, line width=0.95bp] (91.50,23.50) rectangle (157.50,119.72);
  \path[draw=sbTeal, line width=0.95bp] (166.50,23.50) rectangle (232.50,119.72);
  \node[text=sbMut,anchor=base,inner sep=0pt,font={\sffamily\fontsize{5.9bp}{6.8bp}\selectfont\bfseries}] at (49.50,127.70) {Input (F0)};
  \node[text=sbMut,anchor=base,inner sep=0pt,font={\sffamily\fontsize{5.5bp}{6.2bp}\selectfont}] at (49.50,134.70) {Shared start};
  \node[text=sbRed,anchor=base,inner sep=0pt,font={\sffamily\fontsize{5.9bp}{6.8bp}\selectfont\bfseries}] at (124.50,127.70) {Native H3 (F38)};
  \node[text=sbMut,anchor=base,inner sep=0pt,font={\sffamily\fontsize{5.5bp}{6.2bp}\selectfont}] at (124.50,134.70) {No elevation};
  \node[text=sbTeal,anchor=base,inner sep=0pt,font={\sffamily\fontsize{5.9bp}{6.8bp}\selectfont\bfseries}] at (199.50,127.70) {SimShoot (F38)};
  \node[text=sbMut,anchor=base,inner sep=0pt,font={\sffamily\fontsize{5.5bp}{6.2bp}\selectfont}] at (199.50,134.70) {Higher viewpoint};
  \path[fill=sbBg, draw=sbBdr, line width=0.55bp, rounded corners=3.5bp] (250.0,4.0) rectangle (491.0,140.0);
  \path[fill=sbBlueHd, draw=sbBlueBdr, line width=0.50bp] (250.0,19.0) -- (250.0,7.5) .. controls (250.0,5.2) and (251.2,4.0) .. (253.5,4.0) -- (487.5,4.0) .. controls (489.8,4.0) and (491.0,5.2) .. (491.0,7.5) -- (491.0,19.0) -- cycle;
  \node[text=sbInk,anchor=base west,inner sep=0pt,font={\sffamily\fontsize{6.4bp}{7.2bp}\selectfont\bfseries}] at (255.5,14.5) {(b) Case 2 \textperiodcentered{} Camera Zoom};
  \node[text=sbRed,anchor=base east,inner sep=0pt,font={\sffamily\fontsize{5.9bp}{6.8bp}\selectfont\bfseries}] at (485.5,14.5) {Insufficient extent};
  \node[anchor=north west,inner sep=0pt,outer sep=0pt] at (262.50,23.50) {\includegraphics[width=216.00bp,height=96.22bp]{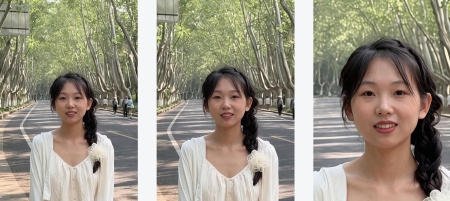}};
  \path[draw=sbBdr, line width=0.60bp] (262.50,23.50) rectangle (328.50,119.72);
  \path[draw=sbRed, line width=0.95bp] (337.50,23.50) rectangle (403.50,119.72);
  \path[draw=sbTeal, line width=0.95bp] (412.50,23.50) rectangle (478.50,119.72);
  \node[text=sbMut,anchor=base,inner sep=0pt,font={\sffamily\fontsize{5.9bp}{6.8bp}\selectfont\bfseries}] at (295.50,127.70) {Input (F0)};
  \node[text=sbMut,anchor=base,inner sep=0pt,font={\sffamily\fontsize{5.5bp}{6.2bp}\selectfont}] at (295.50,134.70) {Shared start};
  \node[text=sbRed,anchor=base,inner sep=0pt,font={\sffamily\fontsize{5.9bp}{6.8bp}\selectfont\bfseries}] at (370.50,127.70) {Native H3 (F38)};
  \node[text=sbMut,anchor=base,inner sep=0pt,font={\sffamily\fontsize{5.5bp}{6.2bp}\selectfont}] at (370.50,134.70) {Minimal zoom};
  \node[text=sbTeal,anchor=base,inner sep=0pt,font={\sffamily\fontsize{5.9bp}{6.8bp}\selectfont\bfseries}] at (445.50,127.70) {SimShoot (F38)};
  \node[text=sbMut,anchor=base,inner sep=0pt,font={\sffamily\fontsize{5.5bp}{6.2bp}\selectfont}] at (445.50,134.70) {Full close-up};
  \path[fill=sbBg, draw=sbBdr, line width=0.55bp, rounded corners=3.5bp] (4.0,144.0) rectangle (245.0,280.0);
  \path[fill=sbBlueHd, draw=sbBlueBdr, line width=0.50bp] (4.0,159.0) -- (4.0,147.5) .. controls (4.0,145.2) and (5.2,144.0) .. (7.5,144.0) -- (241.5,144.0) .. controls (243.8,144.0) and (245.0,145.2) .. (245.0,147.5) -- (245.0,159.0) -- cycle;
  \node[text=sbInk,anchor=base west,inner sep=0pt,font={\sffamily\fontsize{6.4bp}{7.2bp}\selectfont\bfseries}] at (9.5,154.5) {(c) Case 3 \textperiodcentered{} Body Pose};
  \node[text=sbRed,anchor=base east,inner sep=0pt,font={\sffamily\fontsize{5.9bp}{6.8bp}\selectfont\bfseries}] at (239.5,154.5) {Ignored action};
  \node[anchor=north west,inner sep=0pt,outer sep=0pt] at (16.50,163.50) {\includegraphics[width=216.00bp,height=96.22bp]{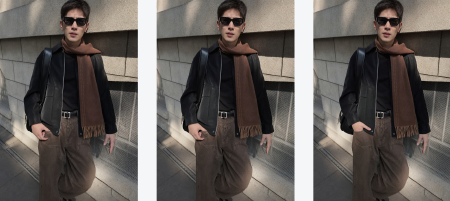}};
  \path[draw=sbBdr, line width=0.60bp] (16.50,163.50) rectangle (82.50,259.72);
  \path[draw=sbRed, line width=0.95bp] (91.50,163.50) rectangle (157.50,259.72);
  \path[draw=sbTeal, line width=0.95bp] (166.50,163.50) rectangle (232.50,259.72);
  \node[text=sbMut,anchor=base,inner sep=0pt,font={\sffamily\fontsize{5.9bp}{6.8bp}\selectfont\bfseries}] at (49.50,267.70) {Input (F0)};
  \node[text=sbMut,anchor=base,inner sep=0pt,font={\sffamily\fontsize{5.5bp}{6.2bp}\selectfont}] at (49.50,274.70) {Shared start};
  \node[text=sbRed,anchor=base,inner sep=0pt,font={\sffamily\fontsize{5.9bp}{6.8bp}\selectfont\bfseries}] at (124.50,267.70) {Native H3 (F38)};
  \node[text=sbMut,anchor=base,inner sep=0pt,font={\sffamily\fontsize{5.5bp}{6.2bp}\selectfont}] at (124.50,274.70) {Hand outside};
  \node[text=sbTeal,anchor=base,inner sep=0pt,font={\sffamily\fontsize{5.9bp}{6.8bp}\selectfont\bfseries}] at (199.50,267.70) {SimShoot (F38)};
  \node[text=sbMut,anchor=base,inner sep=0pt,font={\sffamily\fontsize{5.5bp}{6.2bp}\selectfont}] at (199.50,274.70) {Hand in pocket};
  \path[fill=sbBg, draw=sbBdr, line width=0.55bp, rounded corners=3.5bp] (250.0,144.0) rectangle (491.0,280.0);
  \path[fill=sbBlueHd, draw=sbBlueBdr, line width=0.50bp] (250.0,159.0) -- (250.0,147.5) .. controls (250.0,145.2) and (251.2,144.0) .. (253.5,144.0) -- (487.5,144.0) .. controls (489.8,144.0) and (491.0,145.2) .. (491.0,147.5) -- (491.0,159.0) -- cycle;
  \node[text=sbInk,anchor=base west,inner sep=0pt,font={\sffamily\fontsize{6.4bp}{7.2bp}\selectfont\bfseries}] at (255.5,154.5) {(d) Case 4 \textperiodcentered{} Lighting};
  \node[text=sbRed,anchor=base east,inner sep=0pt,font={\sffamily\fontsize{5.9bp}{6.8bp}\selectfont\bfseries}] at (485.5,154.5) {Off-target drift};
  \node[anchor=north west,inner sep=0pt,outer sep=0pt] at (262.50,163.50) {\includegraphics[width=216.00bp,height=96.22bp]{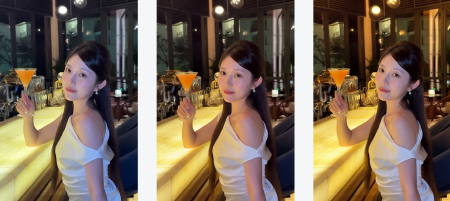}};
  \path[draw=sbBdr, line width=0.60bp] (262.50,163.50) rectangle (328.50,259.72);
  \path[draw=sbRed, line width=0.95bp] (337.50,163.50) rectangle (403.50,259.72);
  \path[draw=sbTeal, line width=0.95bp] (412.50,163.50) rectangle (478.50,259.72);
  \node[text=sbMut,anchor=base,inner sep=0pt,font={\sffamily\fontsize{5.9bp}{6.8bp}\selectfont\bfseries}] at (295.50,267.70) {Input (F0)};
  \node[text=sbMut,anchor=base,inner sep=0pt,font={\sffamily\fontsize{5.5bp}{6.2bp}\selectfont}] at (295.50,274.70) {Shared start};
  \node[text=sbRed,anchor=base,inner sep=0pt,font={\sffamily\fontsize{5.9bp}{6.8bp}\selectfont\bfseries}] at (370.50,267.70) {Native H3 (F38)};
  \node[text=sbMut,anchor=base,inner sep=0pt,font={\sffamily\fontsize{5.5bp}{6.2bp}\selectfont}] at (370.50,274.70) {Weak warm, glass drift};
  \node[text=sbTeal,anchor=base,inner sep=0pt,font={\sffamily\fontsize{5.9bp}{6.8bp}\selectfont\bfseries}] at (445.50,267.70) {SimShoot (F38)};
  \node[text=sbMut,anchor=base,inner sep=0pt,font={\sffamily\fontsize{5.5bp}{6.2bp}\selectfont}] at (445.50,274.70) {Locked pose, warm};
\end{tikzpicture}%
}%
\endgroup

%% file: sec/fig_supp_b4.tex
\begingroup%
\definecolor{sbW}{HTML}{FFFFFF}%
\definecolor{sbBg}{HTML}{F6F7F9}%
\definecolor{sbBdr}{HTML}{CFD5DE}%
\definecolor{sbInk}{HTML}{1A1A1A}%
\definecolor{sbMut}{HTML}{59606B}%
\definecolor{sbRule}{HTML}{E4E8EE}%
\definecolor{sbTeal}{HTML}{0B8F7C}%
\definecolor{sbTealBg}{HTML}{E9F5F2}%
\definecolor{sbTealHd}{HTML}{D3ECE6}%
\definecolor{sbTealBdr}{HTML}{8CCABD}%
\definecolor{sbRed}{HTML}{C43828}%
\definecolor{sbRedBg}{HTML}{FDEEED}%
\definecolor{sbRedBdr}{HTML}{EEA9A0}%
\definecolor{sbBlue}{HTML}{3D6EB4}%
\definecolor{sbBlueBg}{HTML}{EEF4FB}%
\definecolor{sbBlueHd}{HTML}{DCE6F5}%
\definecolor{sbBlueBdr}{HTML}{9DBFE3}%
\definecolor{sbOrg}{HTML}{C0601F}%
\definecolor{sbOrgBg}{HTML}{FDF2EA}%
\definecolor{sbOrgBdr}{HTML}{F0B48A}%
\resizebox{0.72\linewidth}{!}{%
\begin{tikzpicture}[x=1bp,y=-1bp]
  \path[fill=sbW] (0,0) rectangle (495,368);
  \path[fill=sbBg, draw=sbBdr, line width=0.55bp, rounded corners=3.5bp] (4.0,3.0) rectangle (491.0,233.0);
  \path[fill=sbTealHd, draw=sbTealBdr, line width=0.50bp] (4.0,17.5) -- (4.0,6.5) .. controls (4.0,4.2) and (5.2,3.0) .. (7.5,3.0) -- (487.5,3.0) .. controls (489.8,3.0) and (491.0,4.2) .. (491.0,6.5) -- (491.0,17.5) -- cycle;
  \node[text=sbInk,anchor=base west,inner sep=0pt,font={\sffamily\fontsize{6.4bp}{7.2bp}\selectfont\bfseries}] at (10.0,13.2) {(a) Single-Step Extent Monotonicity \textperiodcentered{} Case 1, Round 1 (Body Pose: lower R hand to knee, keep L arm up)};
  \node[text=sbMut,anchor=base,inner sep=0pt,font={\sffamily\fontsize{5.9bp}{6.7bp}\selectfont\bfseries}] at (126.0,27.0) {F0 (0\%)};
  \node[text=sbMut,anchor=base,inner sep=0pt,font={\sffamily\fontsize{5.9bp}{6.7bp}\selectfont\bfseries}] at (173.0,27.0) {F10 (26\%)};
  \node[text=sbMut,anchor=base,inner sep=0pt,font={\sffamily\fontsize{5.9bp}{6.7bp}\selectfont\bfseries}] at (220.0,27.0) {F19 (50\%)};
  \node[text=sbMut,anchor=base,inner sep=0pt,font={\sffamily\fontsize{5.9bp}{6.7bp}\selectfont\bfseries}] at (267.0,27.0) {F29 (76\%)};
  \node[text=sbMut,anchor=base,inner sep=0pt,font={\sffamily\fontsize{5.9bp}{6.7bp}\selectfont\bfseries}] at (314.0,27.0) {F38 (100\%)};
  \node[text=sbMut,anchor=base west,inner sep=0pt,font={\sffamily\fontsize{5.9bp}{6.7bp}\selectfont\bfseries}] at (344.0,27.0) {Observed Progression \& Selection};
  \path[fill=sbTealBg, draw=sbBdr, line width=0.50bp, rounded corners=2.2bp] (8.0,31.0) rectangle (487.0,96.5);
  \node[text=sbTeal,anchor=base west,inner sep=0pt,font={\sffamily\fontsize{6.4bp}{7.2bp}\selectfont\bfseries}] at (14.0,60.0) {SimShoot};
  \node[text=sbMut,anchor=base west,inner sep=0pt,font={\sffamily\fontsize{5.8bp}{6.6bp}\selectfont}] at (14.0,70.0) {(H3 + LoRA)};
  \node[anchor=north west,inner sep=0pt,outer sep=0pt] at (106.00,34.59) {\includegraphics[width=228.00bp,height=58.31bp]{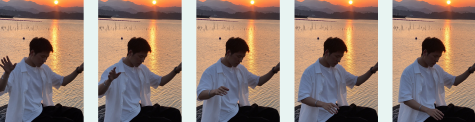}};
  \path[fill=sbTealBg] (146.0,34.4) rectangle (153.0,93.1);
  \path[fill=sbTealBg] (193.0,34.4) rectangle (200.0,93.1);
  \path[fill=sbTealBg] (240.0,34.4) rectangle (247.0,93.1);
  \path[fill=sbTealBg] (287.0,34.4) rectangle (294.0,93.1);
  \path[draw=sbBdr, line width=0.50bp] (106.00,34.59) rectangle (146.00,92.91);
  \path[draw=sbBdr, line width=0.50bp] (153.00,34.59) rectangle (193.00,92.91);
  \path[draw=sbBdr, line width=0.50bp] (200.00,34.59) rectangle (240.00,92.91);
  \path[draw=sbBdr, line width=0.50bp] (247.00,34.59) rectangle (287.00,92.91);
  \path[draw=sbTeal, line width=1.15bp] (294.00,34.59) rectangle (334.00,92.91);
  \node[text=sbInk,anchor=base west,inner sep=0pt,font={\sffamily\fontsize{5.9bp}{6.8bp}\selectfont}] at (344.0,48.2) {Smooth frame-by-frame};
  \node[text=sbInk,anchor=base west,inner sep=0pt,font={\sffamily\fontsize{5.9bp}{6.8bp}\selectfont}] at (344.0,59.8) {downward hand trajectory;};
  \node[text=sbInk,anchor=base west,inner sep=0pt,font={\sffamily\fontsize{5.9bp}{6.8bp}\selectfont}] at (344.0,71.4) {raised L arm stays locked.};
  \node[text=sbTeal,anchor=base west,inner sep=0pt,font={\sffamily\fontsize{5.9bp}{6.8bp}\selectfont\bfseries}] at (344.0,83.0) {R1 pick: F38 (accepted)};
  \path[fill=sbW, draw=sbBdr, line width=0.50bp, rounded corners=2.2bp] (8.0,98.7) rectangle (487.0,164.2);
  \node[text=sbBlue,anchor=base west,inner sep=0pt,font={\sffamily\fontsize{6.4bp}{7.2bp}\selectfont\bfseries}] at (14.0,127.7) {FLUX 3 Klein};
  \node[text=sbMut,anchor=base west,inner sep=0pt,font={\sffamily\fontsize{5.8bp}{6.6bp}\selectfont}] at (14.0,137.7) {(images)};
  \node[anchor=north west,inner sep=0pt,outer sep=0pt] at (106.00,102.30) {\includegraphics[width=228.00bp,height=58.31bp]{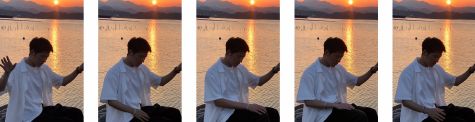}};
  \path[draw=sbBdr, line width=0.50bp] (106.00,102.30) rectangle (146.00,160.61);
  \path[draw=sbBdr, line width=0.50bp] (153.00,102.30) rectangle (193.00,160.61);
  \path[draw=sbBlue, line width=1.15bp] (200.00,102.30) rectangle (240.00,160.61);
  \path[draw=sbBdr, line width=0.50bp] (247.00,102.30) rectangle (287.00,160.61);
  \path[draw=sbBdr, line width=0.50bp] (294.00,102.30) rectangle (334.00,160.61);
  \node[text=sbInk,anchor=base west,inner sep=0pt,font={\sffamily\fontsize{5.9bp}{6.8bp}\selectfont}] at (344.0,115.9) {Jumps to knee at F10 and};
  \node[text=sbInk,anchor=base west,inner sep=0pt,font={\sffamily\fontsize{5.9bp}{6.8bp}\selectfont}] at (344.0,127.5) {saturates across F10--F38};
  \node[text=sbInk,anchor=base west,inner sep=0pt,font={\sffamily\fontsize{5.9bp}{6.8bp}\selectfont}] at (344.0,139.1) {(no intermediate scale).};
  \node[text=sbBlue,anchor=base west,inner sep=0pt,font={\sffamily\fontsize{5.9bp}{6.8bp}\selectfont\bfseries}] at (344.0,150.7) {R1 pick: F19 (accepted)};
  \path[fill=sbW, draw=sbBdr, line width=0.50bp, rounded corners=2.2bp] (8.0,166.4) rectangle (487.0,231.9);
  \node[text=sbRed,anchor=base west,inner sep=0pt,font={\sffamily\fontsize{6.4bp}{7.2bp}\selectfont\bfseries}] at (14.0,195.4) {Qwen-Image 2.1};
  \node[text=sbMut,anchor=base west,inner sep=0pt,font={\sffamily\fontsize{5.8bp}{6.6bp}\selectfont}] at (14.0,205.4) {(images)};
  \node[anchor=north west,inner sep=0pt,outer sep=0pt] at (106.00,170.00) {\includegraphics[width=228.00bp,height=58.31bp]{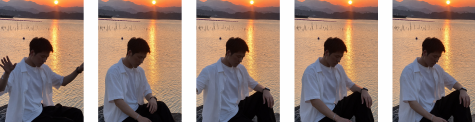}};
  \path[draw=sbRed, line width=1.15bp] (106.00,170.00) rectangle (146.00,228.31);
  \path[draw=sbBdr, line width=0.50bp] (153.00,170.00) rectangle (193.00,228.31);
  \path[draw=sbBdr, line width=0.50bp] (200.00,170.00) rectangle (240.00,228.31);
  \path[draw=sbBdr, line width=0.50bp] (247.00,170.00) rectangle (287.00,228.31);
  \path[draw=sbBdr, line width=0.50bp] (294.00,170.00) rectangle (334.00,228.31);
  \node[text=sbInk,anchor=base west,inner sep=0pt,font={\sffamily\fontsize{5.9bp}{6.8bp}\selectfont}] at (344.0,183.6) {Drops both arms at F10,};
  \node[text=sbInk,anchor=base west,inner sep=0pt,font={\sffamily\fontsize{5.9bp}{6.8bp}\selectfont}] at (344.0,195.2) {violating fixed L-arm};
  \node[text=sbInk,anchor=base west,inner sep=0pt,font={\sffamily\fontsize{5.9bp}{6.8bp}\selectfont}] at (344.0,206.8) {preservation constraint.};
  \node[text=sbRed,anchor=base west,inner sep=0pt,font={\sffamily\fontsize{5.9bp}{6.8bp}\selectfont\bfseries}] at (344.0,218.4) {R1 pick: F0 (rejected)};
  \path[fill=sbBg, draw=sbBdr, line width=0.55bp, rounded corners=3.5bp] (4.0,239.0) rectangle (491.0,365.0);
  \path[fill=sbBlueHd, draw=sbBlueBdr, line width=0.50bp] (4.0,253.5) -- (4.0,242.5) .. controls (4.0,240.2) and (5.2,239.0) .. (7.5,239.0) -- (487.5,239.0) .. controls (489.8,239.0) and (491.0,240.2) .. (491.0,242.5) -- (491.0,253.5) -- cycle;
  \node[text=sbInk,anchor=base west,inner sep=0pt,font={\sffamily\fontsize{6.4bp}{7.2bp}\selectfont\bfseries}] at (10.0,249.2) {(b) Multi-Round Cumulative Drift ({\fontsize{6pt}{6pt}\usefont{OMS}{cmsy}{m}{n}\symbol{20}} 3 Accepted Edits) \textperiodcentered{} Final Outputs from Identical S0};
  \path[fill=sbW, draw=sbBdr, line width=0.50bp, rounded corners=2.2bp] (8.0,257.0) rectangle (245.0,360.5);
  \node[text=sbInk,anchor=base west,inner sep=0pt,font={\sffamily\fontsize{6.1bp}{6.9bp}\selectfont\bfseries}] at (15.0,265.0) {Case 2 (Open-ended)};
  \node[anchor=north west,inner sep=0pt,outer sep=0pt] at (15.00,268.50) {\includegraphics[width=223.00bp,height=75.80bp]{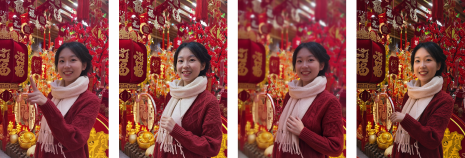}};
  \path[draw=sbBdr, line width=0.55bp] (15.00,268.50) rectangle (67.00,344.30);
  \node[text=sbMut,anchor=base,inner sep=0pt,font={\sffamily\fontsize{5.7bp}{6.4bp}\selectfont}] at (41.00,351.30) {Input S0};
  \node[text=sbMut,anchor=base,inner sep=0pt,font={\sffamily\fontsize{5.4bp}{6.1bp}\selectfont}] at (41.00,357.70) {Shared start};
  \path[draw=sbTeal, line width=1.00bp] (72.00,268.50) rectangle (124.00,344.30);
  \node[text=sbTeal,anchor=base,inner sep=0pt,font={\sffamily\fontsize{5.7bp}{6.4bp}\selectfont\bfseries}] at (98.00,351.30) {SimShoot (3e)};
  \node[text=sbMut,anchor=base,inner sep=0pt,font={\sffamily\fontsize{5.4bp}{6.1bp}\selectfont}] at (98.00,357.70) {Sharp knit/face};
  \path[draw=sbRed, line width=1.00bp] (129.00,268.50) rectangle (181.00,344.30);
  \node[text=sbRed,anchor=base,inner sep=0pt,font={\sffamily\fontsize{5.7bp}{6.4bp}\selectfont\bfseries}] at (155.00,351.30) {FLUX 3 (2e)};
  \node[text=sbMut,anchor=base,inner sep=0pt,font={\sffamily\fontsize{5.4bp}{6.1bp}\selectfont}] at (155.00,357.70) {Softened bg/face};
  \path[draw=sbRed, line width=1.00bp] (186.00,268.50) rectangle (238.00,344.30);
  \node[text=sbRed,anchor=base,inner sep=0pt,font={\sffamily\fontsize{5.7bp}{6.4bp}\selectfont\bfseries}] at (212.00,351.30) {Qwen 2.1 (2e)};
  \node[text=sbMut,anchor=base,inner sep=0pt,font={\sffamily\fontsize{5.4bp}{6.1bp}\selectfont}] at (212.00,357.70) {Gritty/stiff arm};
  \path[fill=sbW, draw=sbBdr, line width=0.50bp, rounded corners=2.2bp] (250.0,257.0) rectangle (487.0,360.5);
  \node[text=sbInk,anchor=base west,inner sep=0pt,font={\sffamily\fontsize{6.1bp}{6.9bp}\selectfont\bfseries}] at (257.0,265.0) {Case 3 (Targeted)};
  \node[anchor=north west,inner sep=0pt,outer sep=0pt] at (257.00,268.50) {\includegraphics[width=223.00bp,height=75.80bp]{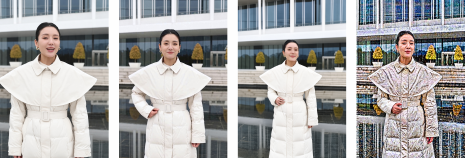}};
  \path[draw=sbBdr, line width=0.55bp] (257.00,268.50) rectangle (309.00,344.30);
  \node[text=sbMut,anchor=base,inner sep=0pt,font={\sffamily\fontsize{5.7bp}{6.4bp}\selectfont}] at (283.00,351.30) {Input S0};
  \node[text=sbMut,anchor=base,inner sep=0pt,font={\sffamily\fontsize{5.4bp}{6.1bp}\selectfont}] at (283.00,357.70) {Shared start};
  \path[draw=sbTeal, line width=1.00bp] (314.00,268.50) rectangle (366.00,344.30);
  \node[text=sbTeal,anchor=base,inner sep=0pt,font={\sffamily\fontsize{5.7bp}{6.4bp}\selectfont\bfseries}] at (340.00,351.30) {SimShoot (3e)};
  \node[text=sbMut,anchor=base,inner sep=0pt,font={\sffamily\fontsize{5.4bp}{6.1bp}\selectfont}] at (340.00,357.70) {Crisp coat/ID};
  \path[draw=sbRed, line width=1.00bp] (371.00,268.50) rectangle (423.00,344.30);
  \node[text=sbRed,anchor=base,inner sep=0pt,font={\sffamily\fontsize{5.7bp}{6.4bp}\selectfont\bfseries}] at (397.00,351.30) {FLUX 3 (3e)};
  \node[text=sbMut,anchor=base,inner sep=0pt,font={\sffamily\fontsize{5.4bp}{6.1bp}\selectfont}] at (397.00,357.70) {Zoomed out too far};
  \path[draw=sbRed, line width=1.00bp] (428.00,268.50) rectangle (480.00,344.30);
  \node[text=sbRed,anchor=base,inner sep=0pt,font={\sffamily\fontsize{5.7bp}{6.4bp}\selectfont\bfseries}] at (454.00,351.30) {Qwen 2.1 (3e)};
  \node[text=sbMut,anchor=base,inner sep=0pt,font={\sffamily\fontsize{5.4bp}{6.1bp}\selectfont}] at (454.00,357.70) {Severe ringing};
\end{tikzpicture}%
}%
\endgroup

%% file: sec/fig_supp_b3.tex
\begingroup%
\definecolor{sbW}{HTML}{FFFFFF}%
\definecolor{sbBg}{HTML}{F6F7F9}%
\definecolor{sbBdr}{HTML}{CFD5DE}%
\definecolor{sbInk}{HTML}{1A1A1A}%
\definecolor{sbMut}{HTML}{59606B}%
\definecolor{sbRule}{HTML}{E4E8EE}%
\definecolor{sbTeal}{HTML}{0B8F7C}%
\definecolor{sbTealBg}{HTML}{E9F5F2}%
\definecolor{sbTealHd}{HTML}{D3ECE6}%
\definecolor{sbTealBdr}{HTML}{8CCABD}%
\definecolor{sbRed}{HTML}{C43828}%
\definecolor{sbRedBg}{HTML}{FDEEED}%
\definecolor{sbRedBdr}{HTML}{EEA9A0}%
\definecolor{sbBlue}{HTML}{3D6EB4}%
\definecolor{sbBlueBg}{HTML}{EEF4FB}%
\definecolor{sbBlueHd}{HTML}{DCE6F5}%
\definecolor{sbBlueBdr}{HTML}{9DBFE3}%
\definecolor{sbOrg}{HTML}{C0601F}%
\definecolor{sbOrgBg}{HTML}{FDF2EA}%
\definecolor{sbOrgBdr}{HTML}{F0B48A}%
\resizebox{0.82\linewidth}{!}{%
\begin{tikzpicture}[x=1bp,y=-1bp]
  \path[fill=sbW] (0,0) rectangle (495,272);
  \path[fill=sbBg, draw=sbBdr, line width=0.55bp, rounded corners=3.5bp] (4.0,3.0) rectangle (491.0,134.0);
  \path[fill=sbBlueHd, draw=sbBlueBdr, line width=0.50bp] (4.0,17.5) -- (4.0,6.5) .. controls (4.0,4.2) and (5.2,3.0) .. (7.5,3.0) -- (487.5,3.0) .. controls (489.8,3.0) and (491.0,4.2) .. (491.0,6.5) -- (491.0,17.5) -- cycle;
  \node[text=sbInk,anchor=base west,inner sep=0pt,font={\sffamily\fontsize{6.4bp}{7.2bp}\selectfont\bfseries}] at (10.0,13.2) {(a) Dynamic Plan Adaptation \textperiodcentered{} Case 1 (Open-ended)};
  \node[text=sbMut,anchor=base east,inner sep=0pt,font={\sffamily\fontsize{5.9bp}{6.8bp}\selectfont}] at (485.0,13.2) {Adaptive inserts zoom \& bokeh after inspecting S1};
  \path[fill=sbW, draw=sbBdr, line width=0.50bp, rounded corners=2.2bp] (9.0,21.5) rectangle (101.0,129.5);
  \node[text=sbInk,anchor=base west,inner sep=0pt,font={\sffamily\fontsize{5.8bp}{6.6bp}\selectfont\bfseries}] at (13.5,32.5) {Initial Plan:};
  \node[text=sbMut,anchor=base west,inner sep=0pt,font={\sffamily\fontsize{5.8bp}{6.6bp}\selectfont}] at (13.5,43.1) {1. smile};
  \node[text=sbMut,anchor=base west,inner sep=0pt,font={\sffamily\fontsize{5.8bp}{6.6bp}\selectfont}] at (13.5,53.7) {2. fill light};
  \node[text=sbOrg,anchor=base west,inner sep=0pt,font={\sffamily\fontsize{5.8bp}{6.6bp}\selectfont}] at (13.5,74.9) {Fixed stays wide};
  \node[text=sbMut,anchor=base west,inner sep=0pt,font={\sffamily\fontsize{5.8bp}{6.6bp}\selectfont\bfseries}] at (13.5,85.5) {with sharp mural;};
  \node[text=sbTeal,anchor=base west,inner sep=0pt,font={\sffamily\fontsize{5.8bp}{6.6bp}\selectfont\bfseries}] at (13.5,96.1) {Ours adds zoom};
  \node[text=sbMut,anchor=base west,inner sep=0pt,font={\sffamily\fontsize{5.8bp}{6.6bp}\selectfont}] at (13.5,106.7) {\& aperture blur.};
  \node[text=sbMut,anchor=base,inner sep=0pt,font={\sffamily\fontsize{5.8bp}{6.6bp}\selectfont\bfseries}] at (169.0,30.5) {Shared Round 1};
  \path[fill=sbOrgBg, rounded corners=2.5bp] (249.0,21.5) rectangle (319.0,130.2);
  \path[fill=sbTealBg, rounded corners=2.5bp] (339.0,21.5) rectangle (482.0,130.2);
  \node[text=sbOrg,anchor=base,inner sep=0pt,font={\sffamily\fontsize{5.8bp}{6.6bp}\selectfont\bfseries}] at (284.0,30.5) {Fixed Initial Plan};
  \node[text=sbTeal,anchor=base,inner sep=0pt,font={\sffamily\fontsize{5.8bp}{6.6bp}\selectfont\bfseries}] at (410.5,30.5) {Adaptive Replanning (Ours)};
  \node[anchor=north west,inner sep=0pt,outer sep=0pt] at (109.00,35.00) {\includegraphics[width=365.00bp,height=78.72bp]{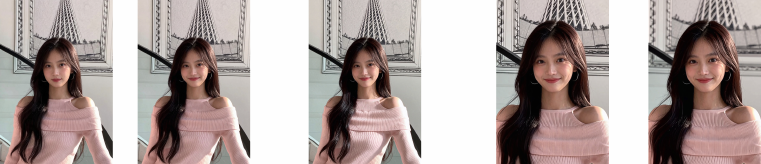}};
  \path[fill=sbBg] (163.0,34.8) rectangle (175.0,114.0);
  \path[fill=sbBg] (229.0,34.8) rectangle (249.0,114.0);
  \path[fill=sbOrgBg] (249.0,34.8) rectangle (257.0,114.0);
  \path[fill=sbOrgBg] (311.0,34.8) rectangle (319.0,114.0);
  \path[fill=sbBg] (319.0,34.8) rectangle (339.0,114.0);
  \path[fill=sbTealBg] (339.0,34.8) rectangle (347.0,114.0);
  \path[fill=sbTealBg] (401.0,34.8) rectangle (420.0,114.0);
  \path[fill=sbTealBg] (474.0,34.8) rectangle (482.0,114.0);
  \path[draw=sbOrgBdr, line width=0.65bp, rounded corners=2.5bp] (249.0,21.5) rectangle (319.0,130.2);
  \path[draw=sbTealBdr, line width=0.65bp, rounded corners=2.5bp] (339.0,21.5) rectangle (482.0,130.2);
  \draw[draw=sbBdr, line width=0.75bp, dash pattern=on 2.2bp off 1.8bp] (329.0,23.5) -- (329.0,128.2);
  \path[fill=sbW, draw=sbBdr, line width=0.55bp, rounded corners=3.5bp] (322.2,68.6) rectangle (335.8,80.1);
  \node[text=sbMut,anchor=base,inner sep=0pt,font={\sffamily\fontsize{5.5bp}{6.2bp}\selectfont\bfseries}] at (329.0,76.3) {vs.};
  \draw[->, draw=sbMut, line width=0.75bp] (165.0,74.4) -- (173.0,74.4);
  \draw[->, draw=sbOrg, line width=0.80bp] (231.5,74.4) -- (246.5,74.4);
  \draw[->, draw=sbTeal, line width=0.80bp] (403.5,74.4) -- (417.5,74.4);
  \path[draw=sbBdr, line width=0.60bp] (109.00,35.00) rectangle (163.00,113.72);
  \node[text=sbMut,anchor=base,inner sep=0pt,font={\sffamily\fontsize{5.7bp}{6.4bp}\selectfont}] at (136.00,120.80) {Input S0};
  \node[text=sbMut,anchor=base,inner sep=0pt,font={\sffamily\fontsize{5.3bp}{6.0bp}\selectfont}] at (136.00,127.20) {Original};
  \path[draw=sbBlue, line width=0.90bp] (175.00,35.00) rectangle (229.00,113.72);
  \node[text=sbBlue,anchor=base,inner sep=0pt,font={\sffamily\fontsize{5.7bp}{6.4bp}\selectfont}] at (202.00,120.80) {Shared S1};
  \node[text=sbMut,anchor=base,inner sep=0pt,font={\sffamily\fontsize{5.3bp}{6.0bp}\selectfont}] at (202.00,127.20) {R1: smile};
  \path[draw=sbOrg, line width=1.05bp] (257.00,35.00) rectangle (311.00,113.72);
  \node[text=sbOrg,anchor=base,inner sep=0pt,font={\sffamily\fontsize{5.7bp}{6.4bp}\selectfont\bfseries}] at (284.00,120.80) {Fixed Final};
  \node[text=sbMut,anchor=base,inner sep=0pt,font={\sffamily\fontsize{5.3bp}{6.0bp}\selectfont}] at (284.00,127.20) {+fill light};
  \path[draw=sbTealBdr, line width=0.60bp] (347.00,35.00) rectangle (401.00,113.72);
  \node[text=sbTeal,anchor=base,inner sep=0pt,font={\sffamily\fontsize{5.7bp}{6.4bp}\selectfont}] at (374.00,120.80) {Ours S2};
  \node[text=sbMut,anchor=base,inner sep=0pt,font={\sffamily\fontsize{5.3bp}{6.0bp}\selectfont}] at (374.00,127.20) {+zoom in};
  \path[draw=sbTeal, line width=1.05bp] (420.00,35.00) rectangle (474.00,113.72);
  \node[text=sbTeal,anchor=base,inner sep=0pt,font={\sffamily\fontsize{5.7bp}{6.4bp}\selectfont\bfseries}] at (447.00,120.80) {Ours Final};
  \node[text=sbMut,anchor=base,inner sep=0pt,font={\sffamily\fontsize{5.3bp}{6.0bp}\selectfont}] at (447.00,127.20) {+aperture};
  \path[fill=sbBg, draw=sbBdr, line width=0.55bp, rounded corners=3.5bp] (4.0,138.0) rectangle (491.0,269.0);
  \path[fill=sbTealHd, draw=sbTealBdr, line width=0.50bp] (4.0,152.5) -- (4.0,141.5) .. controls (4.0,139.2) and (5.2,138.0) .. (7.5,138.0) -- (487.5,138.0) .. controls (489.8,138.0) and (491.0,139.2) .. (491.0,141.5) -- (491.0,152.5) -- cycle;
  \node[text=sbInk,anchor=base west,inner sep=0pt,font={\sffamily\fontsize{6.4bp}{7.2bp}\selectfont\bfseries}] at (10.0,148.2) {(b) Failure Resilience via Rewriting \textperiodcentered{} Case 2 (Open-ended)};
  \node[text=sbMut,anchor=base east,inner sep=0pt,font={\sffamily\fontsize{5.9bp}{6.8bp}\selectfont}] at (485.0,148.2) {Adaptive diagnoses R1 mouth drift \& rewrites open-vocab pose};
  \path[fill=sbW, draw=sbBdr, line width=0.50bp, rounded corners=2.2bp] (9.0,156.5) rectangle (101.0,264.5);
  \node[text=sbInk,anchor=base west,inner sep=0pt,font={\sffamily\fontsize{5.8bp}{6.6bp}\selectfont\bfseries}] at (13.5,167.5) {Initial Plan:};
  \node[text=sbMut,anchor=base west,inner sep=0pt,font={\sffamily\fontsize{5.8bp}{6.6bp}\selectfont}] at (13.5,178.1) {1. arm at side};
  \node[text=sbMut,anchor=base west,inner sep=0pt,font={\sffamily\fontsize{5.8bp}{6.6bp}\selectfont}] at (13.5,188.7) {2. smile};
  \node[text=sbMut,anchor=base west,inner sep=0pt,font={\sffamily\fontsize{5.8bp}{6.6bp}\selectfont}] at (13.5,199.3) {3. fill light};
  \node[text=sbOrg,anchor=base west,inner sep=0pt,font={\sffamily\fontsize{5.8bp}{6.6bp}\selectfont\bfseries}] at (13.5,220.5) {Fixed fails R1 2{\fontsize{6pt}{6pt}\usefont{OMS}{cmsy}{m}{n}\symbol{2}}};
  \node[text=sbMut,anchor=base west,inner sep=0pt,font={\sffamily\fontsize{5.8bp}{6.6bp}\selectfont\bfseries}] at (13.5,231.1) {(arm stays up);};
  \node[text=sbTeal,anchor=base west,inner sep=0pt,font={\sffamily\fontsize{5.8bp}{6.6bp}\selectfont}] at (13.5,241.7) {Ours rewrites R2.};
  \node[text=sbMut,anchor=base,inner sep=0pt,font={\sffamily\fontsize{5.8bp}{6.6bp}\selectfont\bfseries}] at (169.0,165.5) {Shared Round 1};
  \path[fill=sbOrgBg, rounded corners=2.5bp] (249.0,156.5) rectangle (319.0,265.2);
  \path[fill=sbTealBg, rounded corners=2.5bp] (339.0,156.5) rectangle (482.0,265.2);
  \node[text=sbOrg,anchor=base,inner sep=0pt,font={\sffamily\fontsize{5.8bp}{6.6bp}\selectfont\bfseries}] at (284.0,165.5) {Fixed Initial Plan};
  \node[text=sbTeal,anchor=base,inner sep=0pt,font={\sffamily\fontsize{5.8bp}{6.6bp}\selectfont\bfseries}] at (410.5,165.5) {Adaptive Replanning (Ours)};
  \node[anchor=north west,inner sep=0pt,outer sep=0pt] at (109.00,170.00) {\includegraphics[width=365.00bp,height=78.72bp]{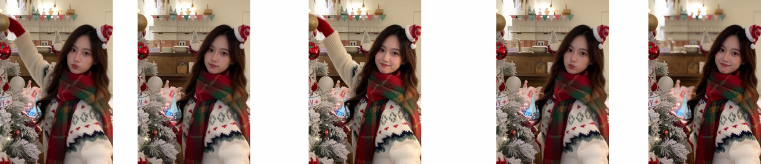}};
  \path[fill=sbBg] (163.0,169.8) rectangle (175.0,249.0);
  \path[fill=sbBg] (229.0,169.8) rectangle (249.0,249.0);
  \path[fill=sbOrgBg] (249.0,169.8) rectangle (257.0,249.0);
  \path[fill=sbOrgBg] (311.0,169.8) rectangle (319.0,249.0);
  \path[fill=sbBg] (319.0,169.8) rectangle (339.0,249.0);
  \path[fill=sbTealBg] (339.0,169.8) rectangle (347.0,249.0);
  \path[fill=sbTealBg] (401.0,169.8) rectangle (420.0,249.0);
  \path[fill=sbTealBg] (474.0,169.8) rectangle (482.0,249.0);
  \path[draw=sbOrgBdr, line width=0.65bp, rounded corners=2.5bp] (249.0,156.5) rectangle (319.0,265.2);
  \path[draw=sbTealBdr, line width=0.65bp, rounded corners=2.5bp] (339.0,156.5) rectangle (482.0,265.2);
  \draw[draw=sbBdr, line width=0.75bp, dash pattern=on 2.2bp off 1.8bp] (329.0,158.5) -- (329.0,263.2);
  \path[fill=sbW, draw=sbBdr, line width=0.55bp, rounded corners=3.5bp] (322.2,203.6) rectangle (335.8,215.1);
  \node[text=sbMut,anchor=base,inner sep=0pt,font={\sffamily\fontsize{5.5bp}{6.2bp}\selectfont\bfseries}] at (329.0,211.3) {vs.};
  \draw[->, draw=sbMut, line width=0.75bp] (165.0,209.4) -- (173.0,209.4);
  \draw[->, draw=sbOrg, line width=0.80bp] (231.5,209.4) -- (246.5,209.4);
  \draw[->, draw=sbTeal, line width=0.80bp] (403.5,209.4) -- (417.5,209.4);
  \node[text=sbOrg,anchor=base,inner sep=0pt,font={\sffamily\fontsize{4.9bp}{5.5bp}\selectfont\bfseries}] at (239.0,199.2) {drops};
  \node[text=sbOrg,anchor=base,inner sep=0pt,font={\sffamily\fontsize{4.9bp}{5.5bp}\selectfont\bfseries}] at (239.0,205.2) {R1};
  \path[draw=sbBdr, line width=0.60bp] (109.00,170.00) rectangle (163.00,248.72);
  \node[text=sbMut,anchor=base,inner sep=0pt,font={\sffamily\fontsize{5.7bp}{6.4bp}\selectfont}] at (136.00,255.80) {Input S0};
  \node[text=sbMut,anchor=base,inner sep=0pt,font={\sffamily\fontsize{5.3bp}{6.0bp}\selectfont}] at (136.00,262.20) {Raised R arm};
  \path[draw=sbRed, line width=0.90bp] (175.00,170.00) rectangle (229.00,248.72);
  \node[text=sbRed,anchor=base,inner sep=0pt,font={\sffamily\fontsize{5.7bp}{6.4bp}\selectfont}] at (202.00,255.80) {R1 F38 (rej.)};
  \node[text=sbMut,anchor=base,inner sep=0pt,font={\sffamily\fontsize{5.3bp}{6.0bp}\selectfont}] at (202.00,262.20) {Mouth drift};
  \path[draw=sbOrg, line width=1.05bp] (257.00,170.00) rectangle (311.00,248.72);
  \node[text=sbRed,anchor=base,inner sep=0pt,font={\sffamily\fontsize{5.7bp}{6.4bp}\selectfont\bfseries}] at (284.00,255.80) {Fixed Final};
  \node[text=sbMut,anchor=base,inner sep=0pt,font={\sffamily\fontsize{5.3bp}{6.0bp}\selectfont}] at (284.00,262.20) {Arm still up};
  \path[draw=sbTealBdr, line width=0.60bp] (347.00,170.00) rectangle (401.00,248.72);
  \node[text=sbTeal,anchor=base,inner sep=0pt,font={\sffamily\fontsize{5.7bp}{6.4bp}\selectfont}] at (374.00,255.80) {Ours S1};
  \node[text=sbMut,anchor=base,inner sep=0pt,font={\sffamily\fontsize{5.3bp}{6.0bp}\selectfont}] at (374.00,262.20) {Rewritten pose};
  \path[draw=sbTeal, line width=1.05bp] (420.00,170.00) rectangle (474.00,248.72);
  \node[text=sbTeal,anchor=base,inner sep=0pt,font={\sffamily\fontsize{5.7bp}{6.4bp}\selectfont\bfseries}] at (447.00,255.80) {Ours Final};
  \node[text=sbMut,anchor=base,inner sep=0pt,font={\sffamily\fontsize{5.3bp}{6.0bp}\selectfont}] at (447.00,262.20) {+smile, DoF};
\end{tikzpicture}%
}%
\endgroup

%% file: sec/fig_supp_b2.tex
\begingroup%
\definecolor{sbW}{HTML}{FFFFFF}%
\definecolor{sbBg}{HTML}{F6F7F9}%
\definecolor{sbBdr}{HTML}{CFD5DE}%
\definecolor{sbInk}{HTML}{1A1A1A}%
\definecolor{sbMut}{HTML}{59606B}%
\definecolor{sbRule}{HTML}{E4E8EE}%
\definecolor{sbTeal}{HTML}{0B8F7C}%
\definecolor{sbTealBg}{HTML}{E9F5F2}%
\definecolor{sbTealHd}{HTML}{D3ECE6}%
\definecolor{sbTealBdr}{HTML}{8CCABD}%
\definecolor{sbRed}{HTML}{C43828}%
\definecolor{sbRedBg}{HTML}{FDEEED}%
\definecolor{sbRedBdr}{HTML}{EEA9A0}%
\definecolor{sbBlue}{HTML}{3D6EB4}%
\definecolor{sbBlueBg}{HTML}{EEF4FB}%
\definecolor{sbBlueHd}{HTML}{DCE6F5}%
\definecolor{sbBlueBdr}{HTML}{9DBFE3}%
\definecolor{sbOrg}{HTML}{C0601F}%
\definecolor{sbOrgBg}{HTML}{FDF2EA}%
\definecolor{sbOrgBdr}{HTML}{F0B48A}%
\resizebox{0.82\linewidth}{!}{%
\begin{tikzpicture}[x=1bp,y=-1bp]
  \path[fill=sbW] (0,0) rectangle (495,234);
  \path[fill=sbBg, draw=sbBdr, line width=0.55bp, rounded corners=3.5bp] (4.0,3.0) rectangle (491.0,103.0);
  \path[fill=sbTealHd, draw=sbTealBdr, line width=0.50bp] (4.0,17.5) -- (4.0,6.5) .. controls (4.0,4.2) and (5.2,3.0) .. (7.5,3.0) -- (487.5,3.0) .. controls (489.8,3.0) and (491.0,4.2) .. (491.0,6.5) -- (491.0,17.5) -- cycle;
  \node[text=sbInk,anchor=base west,inner sep=0pt,font={\sffamily\fontsize{6.4bp}{7.2bp}\selectfont\bfseries}] at (10.0,13.2) {(a) Avoiding Terminal Overshoot \textperiodcentered{} Case 1, Round 1 (Expression: subtle narrow lip parting)};
  \node[text=sbTeal,anchor=base east,inner sep=0pt,font={\sffamily\fontsize{5.9bp}{6.8bp}\selectfont\bfseries}] at (485.0,13.2) {\{F0,F38\} {\fontsize{6pt}{6pt}\usefont{OMS}{cmsy}{m}{n}\symbol{33}} F38 (ovr.) {\normalfont\textbar} Coarse-5 {\fontsize{6pt}{6pt}\usefont{OMS}{cmsy}{m}{n}\symbol{33}} F19 {\normalfont\textbar} Ours {\fontsize{6pt}{6pt}\usefont{OMS}{cmsy}{m}{n}\symbol{33}} F17};
  \path[fill=sbTealBg, draw=sbTeal, line width=0.65bp, dash pattern=on 2.5bp off 1.5bp, rounded corners=2.2bp] (180.5,21.3) rectangle (359.5,96.7);
  \node[text=sbTeal,anchor=base,inner sep=0pt,font={\sffamily\fontsize{5.8bp}{6.6bp}\selectfont\bfseries}] at (270.0,28.7) {Interval Refinement (F10--F19, zero extra video cost)};
  \node[anchor=north west,inner sep=0pt,outer sep=0pt] at (12.00,32.50) {\includegraphics[width=471.00bp,height=54.00bp]{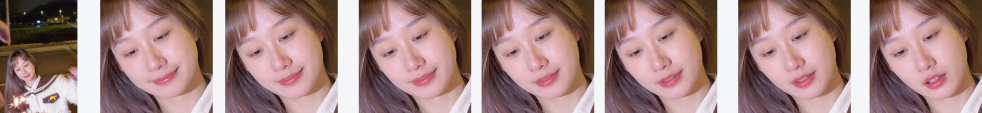}};
  \path[fill=sbBg] (49.0,32.3) rectangle (60.0,86.7);
  \path[fill=sbBg] (114.0,32.3) rectangle (120.0,86.7);
  \path[fill=sbBg] (174.0,32.3) rectangle (180.5,86.7);
  \path[fill=sbTealBg] (180.5,32.3) rectangle (184.0,86.7);
  \path[fill=sbTealBg] (238.0,32.3) rectangle (243.0,86.7);
  \path[fill=sbTealBg] (297.0,32.3) rectangle (302.0,86.7);
  \path[fill=sbTealBg] (356.0,32.3) rectangle (359.5,86.7);
  \path[fill=sbBg] (359.5,32.3) rectangle (366.0,86.7);
  \path[fill=sbBg] (420.0,32.3) rectangle (429.0,86.7);
  \path[draw=sbTeal, line width=0.65bp, dash pattern=on 2.5bp off 1.5bp, rounded corners=2.2bp] (180.5,21.3) rectangle (359.5,96.7);
  \path[draw=sbW, line width=0.75bp] (17.09,59.59) rectangle (30.50,73.02);
  \draw[draw=sbBdr, line width=0.55bp] (54.5,30.5) -- (54.5,88.5);
  \path[draw=sbBdr, line width=0.55bp] (12.00,32.50) rectangle (49.00,86.50);
  \node[text=sbMut,anchor=base,inner sep=0pt,font={\sffamily\fontsize{5.8bp}{6.6bp}\selectfont}] at (30.50,29.50) {Input (full)};
  \node[text=sbMut,anchor=base,inner sep=0pt,font={\sffamily\fontsize{5.8bp}{6.6bp}\selectfont}] at (30.50,94.00) {Face crop};
  \path[draw=sbBdr, line width=0.60bp] (60.00,32.50) rectangle (114.00,86.50);
  \node[text=sbMut,anchor=base,inner sep=0pt,font={\sffamily\fontsize{5.8bp}{6.6bp}\selectfont}] at (87.00,29.50) {F0 (0\%)};
  \node[text=sbMut,anchor=base,inner sep=0pt,font={\sffamily\fontsize{5.8bp}{6.6bp}\selectfont}] at (87.00,94.00) {Current};
  \path[draw=sbBdr, line width=0.60bp] (120.00,32.50) rectangle (174.00,86.50);
  \node[text=sbMut,anchor=base,inner sep=0pt,font={\sffamily\fontsize{5.8bp}{6.6bp}\selectfont}] at (147.00,29.50) {F10 (26\%)};
  \node[text=sbMut,anchor=base,inner sep=0pt,font={\sffamily\fontsize{5.8bp}{6.6bp}\selectfont}] at (147.00,94.00) {Coarse};
  \path[draw=sbTealBdr, line width=0.55bp] (184.00,32.50) rectangle (238.00,86.50);
  \node[text=sbMut,anchor=base,inner sep=0pt,font={\sffamily\fontsize{5.8bp}{6.6bp}\selectfont}] at (211.00,94.00) {F12: Refined};
  \path[draw=sbTealBdr, line width=0.55bp] (243.00,32.50) rectangle (297.00,86.50);
  \node[text=sbMut,anchor=base,inner sep=0pt,font={\sffamily\fontsize{5.8bp}{6.6bp}\selectfont}] at (270.00,94.00) {F15: Refined};
  \path[draw=sbTeal, line width=1.15bp] (302.00,32.50) rectangle (356.00,86.50);
  \node[text=sbTeal,anchor=base,inner sep=0pt,font={\sffamily\fontsize{5.8bp}{6.6bp}\selectfont\bfseries}] at (329.00,94.00) {F17*: Ours (pick)};
  \path[draw=sbBlue, line width=1.00bp] (366.00,32.50) rectangle (420.00,86.50);
  \node[text=sbBlue,anchor=base,inner sep=0pt,font={\sffamily\fontsize{5.8bp}{6.6bp}\selectfont\bfseries}] at (393.00,29.50) {F19 (50\%)};
  \node[text=sbBlue,anchor=base,inner sep=0pt,font={\sffamily\fontsize{5.8bp}{6.6bp}\selectfont\bfseries}] at (393.00,94.00) {Coarse-5};
  \path[draw=sbRed, line width=1.05bp] (429.00,32.50) rectangle (483.00,86.50);
  \node[text=sbRed,anchor=base,inner sep=0pt,font={\sffamily\fontsize{5.8bp}{6.6bp}\selectfont\bfseries}] at (456.00,29.50) {F38 (100\%)};
  \node[text=sbRed,anchor=base,inner sep=0pt,font={\sffamily\fontsize{5.8bp}{6.6bp}\selectfont\bfseries}] at (456.00,94.00) {\{F0,F38\} ovr.};
  \path[fill=sbBg, draw=sbBdr, line width=0.55bp, rounded corners=3.5bp] (4.0,109.0) rectangle (491.0,231.0);
  \path[fill=sbBlueHd, draw=sbBlueBdr, line width=0.50bp] (4.0,123.5) -- (4.0,112.5) .. controls (4.0,110.2) and (5.2,109.0) .. (7.5,109.0) -- (487.5,109.0) .. controls (489.8,109.0) and (491.0,110.2) .. (491.0,112.5) -- (491.0,123.5) -- cycle;
  \node[text=sbInk,anchor=base west,inner sep=0pt,font={\sffamily\fontsize{6.4bp}{7.2bp}\selectfont\bfseries}] at (10.0,119.2) {(b) Recovering from Endpoint Rejection \textperiodcentered{} Case 2, Round 2 (Zoom out for framing)};
  \node[text=sbBlue,anchor=base east,inner sep=0pt,font={\sffamily\fontsize{5.9bp}{6.8bp}\selectfont\bfseries}] at (485.0,119.2) {\{F0,F38\} {\fontsize{6pt}{6pt}\usefont{OMS}{cmsy}{m}{n}\symbol{33}} F0 (rej.) {\normalfont\textbar} Coarse-5 {\fontsize{6pt}{6pt}\usefont{OMS}{cmsy}{m}{n}\symbol{33}} F19 {\normalfont\textbar} Ours {\fontsize{6pt}{6pt}\usefont{OMS}{cmsy}{m}{n}\symbol{33}} F12};
  \path[fill=sbTealBg, draw=sbTeal, line width=0.65bp, dash pattern=on 2.5bp off 1.5bp, rounded corners=2.2bp] (155.5,126.0) rectangle (339.5,224.5);
  \node[text=sbTeal,anchor=base,inner sep=0pt,font={\sffamily\fontsize{5.8bp}{6.6bp}\selectfont\bfseries}] at (247.5,133.2) {Interval Refinement (F10--F19)};
  \node[anchor=north west,inner sep=0pt,outer sep=0pt] at (23.50,136.00) {\includegraphics[width=448.00bp,height=78.72bp]{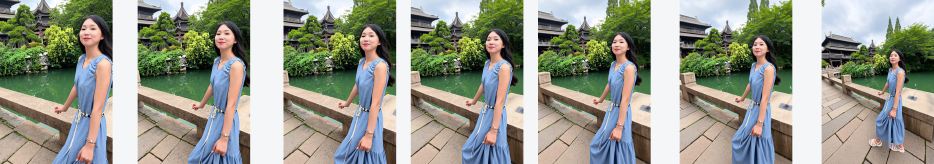}};
  \path[fill=sbBg] (77.5,135.8) rectangle (89.5,214.9);
  \path[fill=sbBg] (143.5,135.8) rectangle (155.5,214.9);
  \path[fill=sbTealBg] (155.5,135.8) rectangle (159.5,214.9);
  \path[fill=sbTealBg] (213.5,135.8) rectangle (220.5,214.9);
  \path[fill=sbTealBg] (274.5,135.8) rectangle (281.5,214.9);
  \path[fill=sbTealBg] (335.5,135.8) rectangle (339.5,214.9);
  \path[fill=sbBg] (339.5,135.8) rectangle (349.5,214.9);
  \path[fill=sbBg] (403.5,135.8) rectangle (417.5,214.9);
  \path[draw=sbTeal, line width=0.65bp, dash pattern=on 2.5bp off 1.5bp, rounded corners=2.2bp] (155.5,126.0) rectangle (339.5,224.5);
  \path[draw=sbRed, line width=1.05bp] (23.50,136.00) rectangle (77.50,214.72);
  \node[text=sbRed,anchor=base,inner sep=0pt,font={\sffamily\fontsize{5.8bp}{6.6bp}\selectfont\bfseries}] at (50.50,133.20) {F0 (0\%)};
  \node[text=sbRed,anchor=base,inner sep=0pt,font={\sffamily\fontsize{5.8bp}{6.6bp}\selectfont\bfseries}] at (50.50,222.00) {\{F0,F38\} rej.};
  \path[draw=sbBdr, line width=0.60bp] (89.50,136.00) rectangle (143.50,214.72);
  \node[text=sbMut,anchor=base,inner sep=0pt,font={\sffamily\fontsize{5.8bp}{6.6bp}\selectfont}] at (116.50,133.20) {F10 (26\%)};
  \node[text=sbMut,anchor=base,inner sep=0pt,font={\sffamily\fontsize{5.8bp}{6.6bp}\selectfont}] at (116.50,222.00) {Coarse};
  \path[draw=sbTeal, line width=1.15bp] (159.50,136.00) rectangle (213.50,214.72);
  \node[text=sbTeal,anchor=base,inner sep=0pt,font={\sffamily\fontsize{5.8bp}{6.6bp}\selectfont\bfseries}] at (186.50,222.00) {F12*: Ours (pick)};
  \path[draw=sbTealBdr, line width=0.55bp] (220.50,136.00) rectangle (274.50,214.72);
  \node[text=sbMut,anchor=base,inner sep=0pt,font={\sffamily\fontsize{5.8bp}{6.6bp}\selectfont}] at (247.50,222.00) {F15: Refined};
  \path[draw=sbTealBdr, line width=0.55bp] (281.50,136.00) rectangle (335.50,214.72);
  \node[text=sbMut,anchor=base,inner sep=0pt,font={\sffamily\fontsize{5.8bp}{6.6bp}\selectfont}] at (308.50,222.00) {F17: Refined};
  \path[draw=sbBlue, line width=1.00bp] (349.50,136.00) rectangle (403.50,214.72);
  \node[text=sbBlue,anchor=base,inner sep=0pt,font={\sffamily\fontsize{5.8bp}{6.6bp}\selectfont\bfseries}] at (376.50,133.20) {F19 (50\%)};
  \node[text=sbBlue,anchor=base,inner sep=0pt,font={\sffamily\fontsize{5.8bp}{6.6bp}\selectfont\bfseries}] at (376.50,222.00) {Coarse-5};
  \path[draw=sbRedBdr, line width=0.80bp] (417.50,136.00) rectangle (471.50,214.72);
  \node[text=sbRed,anchor=base,inner sep=0pt,font={\sffamily\fontsize{5.8bp}{6.6bp}\selectfont}] at (444.50,133.20) {F38 (100\%)};
  \node[text=sbRed,anchor=base,inner sep=0pt,font={\sffamily\fontsize{5.8bp}{6.6bp}\selectfont}] at (444.50,222.00) {Too wide};
\end{tikzpicture}%
}%
\endgroup